\documentclass[accepted]{uai2025} 
                        
\usepackage[american]{babel}

\usepackage{natbib} 
\usepackage{mathtools} 
\usepackage{booktabs} 
\usepackage{tikz} 
\usepackage{multirow}
\usepackage{amssymb}

\title{Probabilistic Embeddings for Frozen Vision-Language Models: Uncertainty
Quantification with Gaussian Process Latent Variable Models}

\author[1]{\href{mailto:<aishwarya.venkataramanan@uni-jena.de>?Subject=Your UAI 2025 paper}{Aishwarya Venkataramanan}{}}
\author[1]{\href{mailto:<paul.bodesheim@uni-jena.de>?Subject=Your UAI 2025 paper}{Paul Bodesheim}{}}
\author[1]{\href{mailto:<joachim.denzler@uni-jena.de>?Subject=Your UAI 2025 paper}{Joachim Denzler}{}}
\affil[1]{%
    Computer Vision Group, Friedrich Schiller University Jena, Germany
}
  
\begin{document}
\maketitle

\begin{abstract}
Vision-Language Models (VLMs) learn joint representations  by mapping images and text into a shared latent space. However, recent research highlights that deterministic embeddings from standard VLMs often struggle to capture the uncertainties arising from the ambiguities in visual and textual descriptions and the multiple possible correspondences between images and texts. Existing approaches tackle this by learning probabilistic embeddings during VLM training, which demands large datasets and does not leverage the powerful representations already learned by large-scale VLMs like CLIP.
In this paper, we propose GroVE, a post-hoc approach to obtaining probabilistic embeddings from frozen VLMs. 
GroVE builds on Gaussian Process Latent Variable Model (GPLVM) to learn a shared low-dimensional latent space where image and text inputs are mapped to a unified representation, optimized through single-modal embedding reconstruction and cross-modal alignment objectives. Once trained, the Gaussian Process model generates uncertainty-aware probabilistic embeddings.
Evaluation shows that GroVE achieves state-of-the-art uncertainty calibration across multiple downstream tasks, including cross-modal retrieval, visual question answering, and active learning.
\end{abstract}

\section{Introduction}\label{sec:intro}
\label{sec:intro}
Deep learning has seen remarkable success over the last decade, yet its practical applicability, especially in safety-critical areas is limited by unreliable, overconfident predictions~\citep{abdar2021review}.
This has motivated the development of methods to quantify uncertainty in model predictions, including stochastic~\citep{blundell2015weight, gal2016dropout}, deterministic~\citep{van2020uncertainty, mukhoti2023deep, venkataramanan2023gaussian}, evidential~\citep{sensoy2018evidential}, and post-hoc approaches~\citep{corbiere2021confidence}, with the aim to produce calibrated confidence values that better reflect the model's actual performance~\citep{guo2017calibration}. While these methods have shown strong performance in tasks involving data from a single modality, they often struggle in multi-modal settings, such as vision language models (VLMs), where inputs come from different domains, such as images and text~\citep{jung2022uncertainty}.
The challenge arises because these single-modal approaches fail to capture the uncertainties that emerge from interactions between the different modalities.

\begin{figure}
    \centering
    \includegraphics[width=0.95\linewidth]{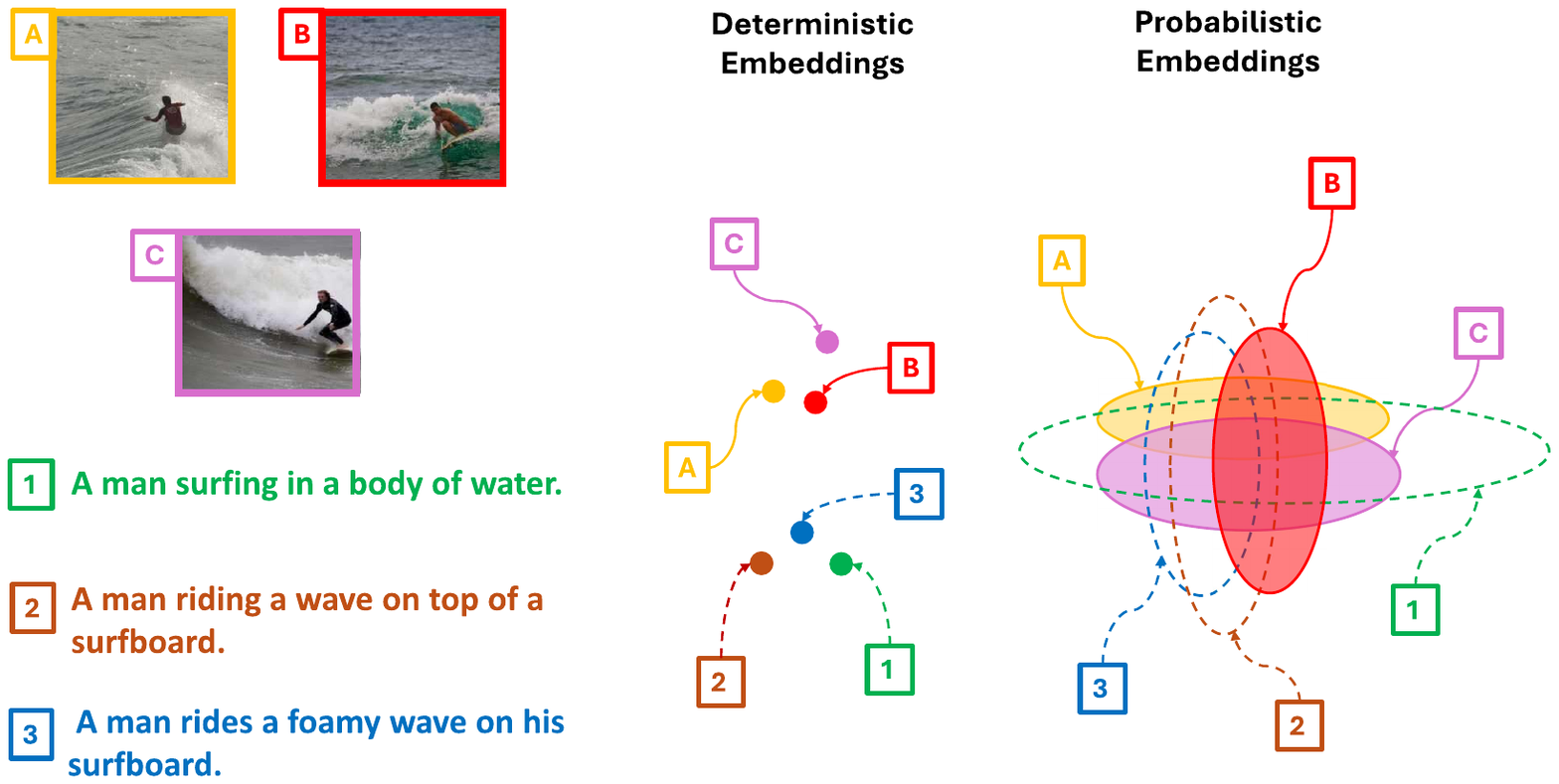}
    \caption{Illustration of uncertainty arising from multiple correspondences between image and text descriptions. Deterministic embeddings represent the instances as fixed points. In contrast, probabilistic embeddings capture uncertainty by modeling text and images as distributions, allowing for multiple reasonable matches.}
    \label{fig:det_vs_probabilsitic}
\end{figure}

VLMs typically encode images and their corresponding text descriptions into vector representations within a joint embedding space. While combining modalities enriches semantics and boosts performance on various tasks~\citep{zhang2024vision}, it also introduces additional uncertainties. Beyond the inherent uncertainty of each modality, there is an uncertainty due to the ambiguous relationships between images and text. This is illustrated in Figure~\ref{fig:det_vs_probabilsitic}, where each image can correspond to multiple text descriptions, and each text description can be associated with multiple images.
Deterministic embeddings from VLMs often fail to capture these uncertainties, motivating the development of probabilistic embeddings~\citep{ji2023map, chun2021probabilistic, chun2023improved}. Probabilistic embeddings represent a distribution, thereby capturing a range of possible representations for ambiguous or uncertain data. Typically, the embeddings are modeled as Gaussian distributions, and deep neural networks are trained to maximize their likelihood, learning the distribution parameters.
However, these methods require training the VLMs from scratch, which requires large-scale datasets, and does not effectively leverage the strong multi-modal representations already provided by the pre-trained large-scale VLMs~\citep{radford2021learning, li2022blip, singh2022flava}.


In this work, we introduce GroVE, a method to generate probabilistic embeddings for VLMs in a post-hoc manner
that builds on Gaussian Process Latent Variable Model (GPLVM)~\citep{lawrence2003gaussian}. GroVE stands for \underline{G}aussian Process for P\underline{ro}babilistic \underline{V}LM \underline{E}mbeddings.
A GPLVM models the relationship between a low-dimensional latent space and a high-dimensional observational space using Gaussian Processes (GPs). Traditionally, the latent space is used for dimensionality reduction~\citep{lawrence2003gaussian,lalchand2022generalised}, and less commonly for more task-specific applications, such as classification~\citep{eleftheriadis2014discriminative} and cross-modal retrieval~\citep{song2017multimodal}.
In our approach, we adopt an extension of the GPLVM framework to a multi-modal context, and show that it provides a principled approach for obtaining probabilistic image and text embeddings from the deterministic embeddings of the large-scale frozen VLMs. To achieve this, we learn a joint low-dimensional latent space, where each pair of image and text embeddings derived from a VLM is represented as a single unified point.
The mapping between the latent space and the observed VLM embeddings is established through two GPs: one for image embeddings and one for text embeddings. Our training objective consists of an embedding reconstruction loss to learn this mapping, and a cross-modal alignment that regularizes the latent space to preserve the semantic structure of the data.
Once the latent space is learned, the trained GP models are used to obtain probabilistic embeddings for images and texts. 

We evaluate GroVE for uncertainty calibration in cross-modal retrieval using CLIP~\citep{radford2021learning} and BLIP~\citep{li2022blip} on the following standard benchmarks: common objects datasets MS-COCO~\citep{lin2014microsoft} and Flickr30k\citep{young2014image}, as well as fine-grained datasets CUB-200-2011~\citep{wah2011caltech} and Oxford Flowers 102~\citep{nilsback2008automated}. We further demonstrate the applicability of our approach in an active learning setting. We also evaluate its ability to provide calibrated uncertainty estimates in visual question answering (VQA) using the VQA 2.0 dataset~\citep{goyal2017making}.
Our results show that GroVE effectively learns probabilistic embeddings that provide calibrated uncertainty estimates. 

The contributions are summarized as follows:  
i) We propose GroVE, which extends GPLVM and provides a principled approach to obtain probabilistic VLM embeddings for both images and text. 
ii) We show that GroVE produces calibrated uncertainty estimates for cross-modal retrieval and VQA, and demonstrate its practical utility in active learning. 
iii) We design GroVE to work in a post-hoc manner on frozen VLMs, avoiding the need for retraining large-scale models from scratch. 
Code is available: \url{https://github.com/cvjena/GroVE-Probabilistic_VLM_embeddings.git}

\section{Related Work} \label{sec:related_work}

\textbf{Vision Language Models.} 
Early vision-language approaches used textual data to embed images in a semantic space, capturing semantic relationships and improving zero-shot capabilities~\cite{frome2013devise, barz2019hierarchy, venkataramanan2023integrating, li2017learning, zhang2015zero}.
The advent of transformers~\cite{vaswani2017attention} revolutionized the landscape of vision-language modeling. Models like VisualBERT~\cite{li2019visualbert}, ViLBERT~\cite{lu2019vilbert} and LXMERT~\cite{tan2019lxmert} extended the BERT architecture~\cite{kenton2019bert} 
to model complex relationships between image regions and text tokens. 
CLIP~\cite{radford2021learning} is a prominent VLM trained on 400 million web-sourced image-caption pairs using a contrastive learning objective~\cite{gutmann2010noise,oord2018representation} to align images with their textual descriptions in a shared embedding space while separating unrelated pairs. CLIP demonstrates strong zero-shot performance across diverse tasks, including image classification~\cite{li2023rs, qian2024online}, object detection~\cite{lin2023gridclip, wu2023cora}, cross-modal retrieval~\cite{xia2023clip, li2024ckdh}, and visual question answering~\cite{xing2024clipvqa, parelli2023clip}.
BLIP~\cite{li2022blip} leverages noisy data through bootstrapping, combined with contrastive learning to achieve state-of-the-art performance. 
However, these methods rely on deterministic embeddings that do not capture modality uncertainty. In contrast, our approach converts deterministic embeddings into probabilistic representations,
with proper uncertainty estimates from GP models.

\textbf{Uncertainty Quantification in VLMs.} 
Input data ambiguities in VLMs are often addressed by replacing traditional deterministic embeddings with probabilistic embeddings~\cite{li2022differentiable}. PCME~\citep{chun2021probabilistic} models image and text embeddings as Gaussians with learned means and variances, optimizing the joint embedding space with a soft cross-modal contrastive loss. PCME++~\citep{chun2023improved} introduces Closed-Form Sampled Distance (CSD) to compute Gaussian embeddings of images and text for faster  uncertainty estimation compared to PCME. MAP~\citep{ji2023map} introduces a Probability Distribution Encoder to model multi-modal representations as probabilistic distributions. However, all these methods require training from scratch, and do not effectively leverage the strong multi-modal representations already learned by the pre-trained large-scale VLMs. 
ProbVLM~\citep{upadhyay2023probvlm} is a post-hoc approach that trains neural networks to estimate the parameters of  Generalized Gaussian distribution for image and text embeddings.
Although being straightforward, the prediction of distribution parameters lacks proper probabilistic modeling of statistical processes underlying the sampling of data.
Furthermore, neural networks are prone to uncalibrated predictions when presented with out-of-distribution (OOD) data or limited training samples~\citep{guo2017calibration}. In contrast, our approach leverages GPs, a Bayesian method that inherently incorporates probabilistic reasoning with reliable and theoretically sound uncertainty quantification as well as distance-awareness through the covariance function, which has proven effective in calibrated uncertainty estimation~\citep{liu2020simple, jung2022uncertainty}.

\textbf{Post-hoc approaches for uncertainty quantification.}
Some of the widely used post-hoc calibration techniques for data from a single modality are temperature scaling~\citep{guo2017calibration} and Platt scaling~\citep{platt1999probabilistic}, which adjust the model's predicted probabilities after training to better align predicted confidence scores with actual performance. Test-Time Data Augmentation (TTDA)~\citep{ayhan2018test, wang2019aleatoric} quantifies uncertainty by applying various transformations to input data during inference, generating multiple predictions, and measuring the variability among them to assess the uncertainty.
A line of work~\citep{corbiere2021confidence, yu2021slurp, hornauer2023out,  shi2019probabilistic} focuses on training auxiliary models to quantify uncertainty in the primary model, allowing for uncertainty estimation without impacting the performance of the primary model. Unlike these single-modal approaches, our method captures uncertainty from the relationship between visual and textual modalities, which is crucial for obtaining accurate uncertainty estimates in VLMs~\citep{jung2022uncertainty}.

\section{Method}
\label{sec:method}

\begin{figure}[t]
    \centering
    \vspace{-25pt}
    \includegraphics[angle=-90, width=\linewidth]{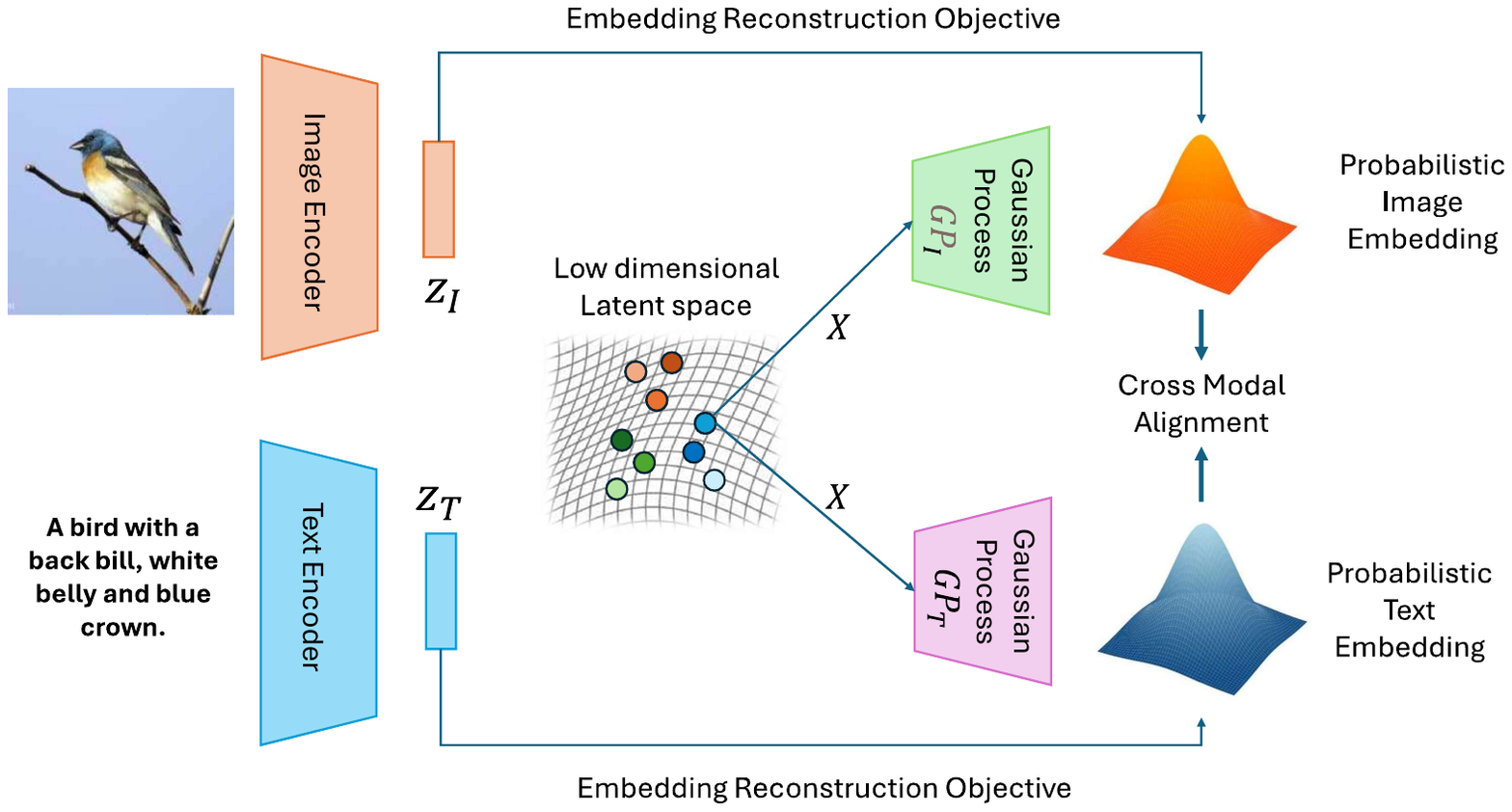}
    \vspace{-25pt}
    \caption{\textbf{Method overview of GroVE.} Given deterministic image and text embeddings from a frozen VLM, GroVE learns a joint low dimensional latent space, where each image-text pair is represented by a single point. Two GP models learn to reconstruct the image and text embeddings from the latent space points through single-modal reconstruction and cross-modal alignment objectives. The GP models act as probabilistic mappings that model the uncertainty in both the image and text modalities.}
    \label{fig:method}
\end{figure}

GroVE builds on the GPLVM framework to learn a shared latent space for image and text inputs using GPs. It optimizes this space through single-modal reconstruction and cross-modal alignment loss, generating probabilistic embeddings from deterministic VLM embeddings to capture uncertainty. Figure~\ref{fig:method} illustrates the overall pipeline of GroVE.

\subsection{Problem Description} \label{sec:formulation}

Let $\mathcal{D} = \{(I_n, T_n)\}_{n=1}^{N}\subset{\mathcal{I}\times\mathcal{T}}$ 
represent a dataset of $N$ paired samples, where $I_n\in\mathcal{I}$ is an image sampled from the image space $\mathcal{I}$, and $T_n\in\mathcal{T}$ is the corresponding text description sampled from the text space $\mathcal{T}$. 
The VLM maps an image $I$ and a text $T$ into a shared embedding space $\mathcal{Z}\subseteq\mathbb{R}^D$. To achieve this, the VLM consists of an image encoder $f_\mathcal{I}^{\theta_\mathcal{I}}:\mathcal{I} \rightarrow \mathcal{Z}$ with  parameters $\theta_\mathcal{I}$, and a text encoder $f_\mathcal{T}^{\theta_\mathcal{T}}:\mathcal{T} \rightarrow \mathcal{Z}$ with parameters $\theta_\mathcal{T}$. We assume that the VLM has already been trained on a large-scale dataset, and the parameters of 
$f_\mathcal{I}^{\theta_\mathcal{I}}$ and $f_\mathcal{T}^{\theta_\mathcal{T}}$ are fixed as $\theta_\mathcal{I}^{*}$ and $\theta_\mathcal{T}^{*}$, respectively. The encoders have been trained such that, for a given image-text pair $(I,T)$, the resulting embeddings $\mathbf{z}_I=f_\mathcal{I}^{\theta_\mathcal{I}^{*}}(I)$ and $\mathbf{z}_T=f_\mathcal{T}^{\theta_\mathcal{T}^{*}}(T)$ are positioned close to one another in $\mathcal{Z}$, so that semantically related visual and textual information is aligned.

While deterministic VLMs provide fixed embeddings, they lack the ability to represent the uncertainty associated with these embeddings. To address this, we propose GroVE, a method that leverages GPLVM to obtain probabilistic embeddings in a post-hoc manner to model the uncertainties.

\subsection{GroVE Model}

We obtain the image and text embeddings from the frozen VLM on $\mathcal{D}$:
\begin{equation}
    \left\{\Bigl(\mathbf{z}_{I_n},\mathbf{z}_{T_n}\Bigr) = \Bigl(f_\mathcal{I}^{\theta_\mathcal{I}^{*}}(I_n),f_\mathcal{T}^{\theta_\mathcal{T}^{*}}(T_n)\Bigr)\right\}_{n=1}^N   ,
\end{equation}
where $\mathbf{z}_{I_n}, \mathbf{z}_{T_n}\in\mathbb{R}^D$ are $D$-dimensional image and text embeddings, respectively.

To derive probabilistic embeddings using GPLVM, we assume that $\mathbf{z}_{I_n}$ and $\mathbf{z}_{T_n}$ are generated from a shared low-dimensional latent space $\mathcal{X}\subseteq\mathbb{R}^Q$ with $Q \ll D$, where each image-text pair $(\mathbf{z}_{I_n},\mathbf{z}_{T_n})$ is associated with a common latent point $\mathbf{x}_n\in\mathbb{R}^Q$. 
We define two GPLVM models $\mathcal{GP_I}$ and $\mathcal{GP_T}$, one for each modality (images from $\mathcal{I}$ and text from $\mathcal{T}$), to learn the mappings $G_\mathcal{I}: \mathcal{X}\to\mathcal{Z_I}$ and $G_\mathcal{T}: \mathcal{X}\to\mathcal{Z_T}$ from $\mathbf{x}_n$ to the high-dimensional embeddings $\mathbf{z}_{I_n}$ and $\mathbf{z}_{T_n}$, respectively. During GP model training, the latent points $\mathbf{x}_n$ are optimized to maximize the likelihood of the observed embeddings $\mathbf{z}_{I_n}$ and $\mathbf{z}_{T_n}$

\textbf{GP model definitions.}
For describing the GPLVM models, we define the matrix $\mathbf{X}\in\mathbb{R}^{N{\times}Q}$ as the collection of the $N$ latent inputs $\mathbf{x}_n$.
Image embeddings $\mathbf{z}_{I_n}$ and text embeddings $\mathbf{z}_{T_n}$ are supposed to be computed from latent functions $G_\mathcal{I}$ and $G_\mathcal{T}$: 
\begin{equation}
    \mathbf{z}_{I_n} = G_\mathcal{I}(\mathbf{x}_n) + \boldsymbol{\epsilon}_\mathcal{I} ; \quad
    \mathbf{z}_{T_n} = G_\mathcal{T}(\mathbf{x}_n) + \boldsymbol{\epsilon}_\mathcal{T}   ,
\end{equation}
with noise terms $\boldsymbol{\epsilon}_\mathcal{I}$ and $\boldsymbol{\epsilon}_\mathcal{T}$ and a GP prior such that for each dimension $d$ of the embeddings $\mathbf{z}_{I_n}$ and $\mathbf{z}_{T_n}$, the latent function values $\mathbf{g}_I^d, \mathbf{g}_T^d\in\mathbb{R}^N$ of the $N$ samples follow a multivariate Gaussian distribution: 
\begin{equation}
\begin{aligned}
    \mathbf{g}_\mathcal{I}^d &\sim \mathcal{N}(m_\mathcal{I}(\mathbf{X}), k_\mathcal{I}(\mathbf{X},\mathbf{X})),\\
    \mathbf{g}_\mathcal{T}^d &\sim \mathcal{N}(m_\mathcal{T}(\mathbf{X}), k_\mathcal{T}(\mathbf{X},\mathbf{X})).
\end{aligned}
\end{equation}

These distributions are parameterized by a mean function $m(\cdot)$ and a covariance function $k(\cdot,\cdot)$, which defines the covariance matrix between pairs of points in $\mathbf{X}$.
For both GPLVM models, we use a constant mean function $m(\mathbf{X})=\mathbf{m}$ and a radial basis function (RBF) kernel $k(\mathbf{x}_i, \mathbf{x}_j)=\exp{\left( -\frac{\|\mathbf{x}_i-\mathbf{x}_j\|^2}{2\ell^2} \right)}$ with length-scale hyperparameter $\ell$. 
However, optimal values for $\mathbf{m}$ and $\ell$ are learnt separately for each modality $\mathcal{I}$ and $\mathcal{T}$.
The likelihood functions are  defined as:
\begin{equation}
\begin{aligned}       p\bigl(\mathbf{z}_\mathcal{I}^d|\mathbf{g}_\mathcal{I}^d\bigr) &= 
\prod_{n=1}^{N}p(\mathbf{z}_{I_n}^{d}|\mathbf{g}_{I_n}^d) = 
\mathcal{N}\bigl(\mathbf{g}_\mathcal{I}^d, \sigma_\mathcal{I}^2\mathbf{I}\bigr),\\
p\bigl(\mathbf{z}_\mathcal{T}^d|\mathbf{g}_\mathcal{T}^d\bigr) &= 
\prod_{n=1}^{N}p(\mathbf{z}_{T_n}^{d}|\mathbf{g}_{T_n}^d) = \mathcal{N}\bigl(\mathbf{g}_\mathcal{T}^d, \sigma_\mathcal{T}^2\mathbf{I}\bigr),
\end{aligned}
\end{equation}
where $\sigma_\mathcal{I}^2,\sigma_\mathcal{T}^2$ are the parameters of the Gaussian noise model, which are learned along with the model parameters during the training.

\textbf{Embedding Reconstruction Objective.}
Given the prior and the likelihood, our goal is to estimate the posterior distribution.
While the exact inference is possible, it is computationally expensive, with cost $\mathcal{O}(N^3)$. In this work, we adopt a sparse GP with inducing points and variational inference~\cite{titsias2009variational}. 
We introduce $M$ inducing points in $\mathcal{X}$ for each modality $\mathcal{I}$ and $\mathcal{T}$, 
where $M \ll N$. 
Each inducing point corresponds to an inducing variable, represented as the latent function values $\mathbf{u}_\mathcal{I}^d \in \mathbb{R}^{M}$ and $\mathbf{u}_\mathcal{T}^d \in \mathbb{R}^{M}$, which capture the latent function values at these locations. The key idea is to approximate the true posterior distribution over the latent function values at the observed data points by conditioning on the inducing variables. 
This reduces the computational complexity of the model to $\mathcal{O}(NM^2)$.

To achieve this, we introduce a variational distribution over the inducing variables as:
\begin{equation} \label{eq:var_posterior}
    q(\mathbf{u}_\mathcal{I}^d) = \mathcal{N}(\mathbf{u}_\mathcal{I}^d | \boldsymbol{\mu}_\mathcal{I}^d, \mathbf{S}_\mathcal{I}^d); \;
    q(\mathbf{u}_\mathcal{T}^d) = \mathcal{N}(\mathbf{u}_\mathcal{T}^d | \boldsymbol{\mu}_\mathcal{T}^d, \mathbf{S}_\mathcal{T}^d) ,
\end{equation}
where $\boldsymbol{\mu}_\mathcal{I}^d$ and $\boldsymbol{\mu}_\mathcal{T}^d$, $\mathbf{S}_\mathcal{I}^d$ and $\mathbf{S}_\mathcal{T}^d$ are variational parameters that are optimized during training. These variational parameters, the inducing points, along with the model parameters $\mathbf{m}_\mathcal{I}$, $\mathbf{m}_\mathcal{T}$, $l_\mathcal{I}$, $l_\mathcal{T}$, $\sigma_\mathcal{I}^2$ and $\sigma_\mathcal{T}^2$ 
are learned by maximizing the lower bound on the marginal likelihood of the data i.e. the evidence lower bound (ELBO), given by
\begin{equation}
\begin{aligned}
    \mathcal{L}_{ELBO}^d =  \mathbb{E}_{q(\mathbf{g}_\mathcal{I}^d)}[\log p(\mathbf{z}_\mathcal{I}^d|\mathbf{g}_\mathcal{I}^d)] - D_{KL}(q(\mathbf{u}_\mathcal{I}^d)||p(\mathbf{u}_\mathcal{I}^d)) \\ + \mathbb{E}_{q(\mathbf{g}_\mathcal{T}^d)}[\log p(\mathbf{z}_\mathcal{T}^d|\mathbf{g}_\mathcal{T}^d)] - D_{KL}(q(\mathbf{u}_\mathcal{T}^d)||p(\mathbf{u}_\mathcal{T}^d)),
\end{aligned}
\end{equation}
where $D_{KL}$ is the Kullback-Leibler (KL) divergence, and is measured between the variational distributions and their corresponding priors obtained by the GP prior evaluated at the inducing points. 
The embedding reconstruction objective is given by:
\begin{equation}
    \mathcal{L}_{emb} = -\sum_{d=1}^{D} \mathcal{L}_{ELBO}^d 
\end{equation}

\textbf{Cross-modal Alignment Objective.} 
In addition to this reconstruction objective, we introduce a regularization term, so that the predicted distributions of the corresponding image and text embeddings from the GPs match. Aligning these distributions encourages the latent space to learn a shared underlying structure between the modalities, so that semantically related data points are represented by similar latent variables.
To enforce this, we define a KL divergence loss function between the distributions of the image and text embeddings from the GP models, 
which take the forms $\mathcal{N}(\hat{\boldsymbol{\mu}}_\mathcal{I}, \hat{\boldsymbol{\Sigma}}_\mathcal{I})$ and $\mathcal{N}(\hat{\boldsymbol{\mu}}_\mathcal{T}, \hat{\boldsymbol{\Sigma}}_\mathcal{T})$ respectively (refer Sec.~\ref{sec:inference} for inference using GP). 
The resulting objective $\mathcal{L}_{KL}$ is the mean of the KL divergence in both directions (image-to-text and text-to-image):
\begin{multline} \label{eq:kl}
    \mathcal{L}_{KL} = \frac{1}{2}[ D_{KL}(\mathcal{N}(\hat{\boldsymbol{\mu}}_\mathcal{I}, \hat{\boldsymbol{\Sigma}}_\mathcal{I}) \| \mathcal{N}(\hat{\boldsymbol{\mu}}_\mathcal{T}, \hat{\boldsymbol{\Sigma}}_\mathcal{T})) + \\ 
    D_{KL}(\mathcal{N}(\hat{\boldsymbol{\mu}}_\mathcal{T}, \hat{\boldsymbol{\Sigma}}_\mathcal{T}) \| \mathcal{N}(\hat{\boldsymbol{\mu}}_\mathcal{I}, \hat{\boldsymbol{\Sigma}}_\mathcal{I})) ]. 
\end{multline}

\textbf{Final Objective.} The overall objective function is the weighted sum of the embedding reconstruction loss and the cross-modal alignment loss:
\begin{equation}
    \mathcal{L}_{total} = \lambda_1\mathcal{L}_{emb} + \lambda_2\mathcal{L}_{KL} 
\end{equation}
where $\lambda_1$ and $\lambda_2$ are trade-off parameters.

\subsection{Probabilistic Embeddings}\label{sec:inference}
Once the latent space representation $\mathbf{X}$ is learned, we use $\mathcal{GP_I}$ and $\mathcal{GP_T}$ to predict the probabilistic image and text embeddings. Given a new embedding $\mathbf{z}_*$ (image or text) obtained from the VLM, we first infer its latent representation $\mathbf{x}_*$ by randomly initializing $\mathbf{x}_*$ and iteratively optimizing it with the ELBO. This approximates the posterior distribution $p(\mathbf{x}_*|\mathbf{z}_*, \mathbf{z}_\mathcal{M})$, where $\mathcal{M}$ denotes the modality (either $\mathcal{I}$ or $\mathcal{T}$).
From $\mathbf{x}_*$, the probabilistic embedding can be inferred using the respective GP.

\textbf{Inference using GP.} The predictive distribution, which defines the predicted probabilistic embedding is given by:
\begin{equation}
    p(\mathbf{g}_*^d) = \int p(\mathbf{g}_*^d|\mathbf{u}^d_\mathcal{M})q(\mathbf{u}^d_\mathcal{M})d\mathbf{u}^d_\mathcal{M}
\end{equation}
Evaluating the integral results in a Gaussian distribution~\cite{hensman2015scalable}:
\begin{equation}
    p(\mathbf{g}_*^d) = \mathcal{N}(\mathbf{g}_*^d | \boldsymbol{\hat{\mu}}_*^d, \boldsymbol{\tilde{\Sigma}}_*^d)
\end{equation}
where the mean $\boldsymbol{\hat{\mu}}_*^d$ and covariance $\boldsymbol{\hat{\Sigma}}_*^d$ of the embedding is:
\begin{equation} \label{eq:gp_mean}
    \boldsymbol{\hat{\mu}}_*^d = \mathbf{m}_\mathcal{M} + \mathbf{A}(\boldsymbol{\mu}^d_\mathcal{M} - \mathbf{m}_{\mathbf{v}_\mathcal{M}})
\end{equation}
\begin{equation} \label{eq:gp_var}
    \boldsymbol{\hat{\Sigma}}_*^d = k(\mathbf{x}_*,\mathbf{x}_*) - \mathbf{A}(\mathbf{S}^d_\mathcal{M}-k(\mathbf{v}_\mathcal{M},\mathbf{v}_\mathcal{M}))\mathbf{A}^T,
\end{equation}
where $\mathbf{v}_\mathcal{M}$ refers to the inducing points of the respective modality, 
$\mathbf{A}=k(\mathbf{x}_*,\mathbf{v}_\mathcal{M})k(\mathbf{v}_\mathcal{M},\mathbf{v}_\mathcal{M})^{-1}$, with dimensions $k(\mathbf{v}_\mathcal{M},\mathbf{v}_\mathcal{M}) \in \mathbb{R}^{M \times M}$ and $k(\mathbf{x}_*,\mathbf{v}_\mathcal{M}) \in \mathbb{R}^M$ and $\mathbf{m}_{\mathbf{v}_\mathcal{M}}$ is the prior mean evaluated at $\mathbf{v}_\mathcal{M}$.

\begin{table*}[h]
    \begin{adjustbox}{width=2\columnwidth, center}
    \centering
\begin{tabular}{@{}p{-1cm}lcccccccccccc@{}}
    \toprule
&\multirow{2}{*}{Method} & \multicolumn{3}{c}{Flickr} & \multicolumn{3}{c}{COCO} & \multicolumn{3}{c}{CUB} & \multicolumn{3}{c}{Flowers} \\ \cmidrule(lr){3-5} \cmidrule(lr){6-8} \cmidrule(lr){9-11} \cmidrule(lr){12-14}
 && $S \downarrow$ & $R^2 \uparrow$ & $-SR^2 \uparrow$ & $S \downarrow$ & $R^2 \uparrow$ & $-SR^2 \uparrow$ & $S \downarrow$ & $R^2 \uparrow$ & $-SR^2 \uparrow$ & $S \downarrow$ & $R^2 \uparrow$ & $-SR^2 \uparrow$ \\ \midrule
        \multirow{7}{*}{\rotatebox{90}{\textbf{Image to Text}}} 
        & Deterministic & \underline{-0.80$\pm$0.00} & \underline{0.66$\pm$0.00} & \underline{0.52$\pm$0.00} & 
        \underline{-0.80$\pm$0.00} & 0.64$\pm$0.00 & \underline{0.51$\pm$0.00} & 
        -0.10$\pm$0.00 & 0.05$\pm$0.00 & 0.00$\pm$0.00 &
        -0.10$\pm$0.00 & 0.00$\pm$0.00 & 0.00$\pm$0.00  \\
        &TTDA & 0.12$\pm$0.03 & 0.32$\pm$0.07 & -0.03$\pm$0.01 & 
        -0.36$\pm$0.05  & 0.38$\pm$0.08 & 0.17$\pm$0.05& 
        -0.60$\pm$0.00 & 0.36$\pm$0.07 & 0.21 $\pm$0.04 &
        \underline{-0.78$\pm$0.04} & 0.37$\pm$0.07 & 0.28$\pm$0.06  \\
        &PFE & -0.34$\pm$0.06 & 0.45$\pm$0.04 & 0.13$\pm$0.03 &
        0.63$\pm$0.05 & \underline{0.72$\pm$0.07} & -0.46$\pm$0.05 & 
        -0.13$\pm$0.04 & 0.28$\pm$0.03 & 0.02$\pm$0.01 &
        -0.11$\pm$0.05 & 0.29$\pm$0.04 & 0.04$\pm$0.01  \\    
        &PCME & 0.61$\pm$0.06 & 0.18$\pm$0.02 & -0.11$\pm$0.02 & 
        -0.63$\pm$0.00 & 0.50$\pm$0.03 & 0.31$\pm$0.02 & 
        -0.19$\pm$0.05 & 0.13$\pm$0.03 & 0.03$\pm$0.01&
        0.12$\pm$0.07 & 0.04$\pm$0.03 & 0.00$\pm$0.01  \\          
        &PCME++ & -0.08$\pm$0.04 & 0.33$\pm$0.04 & 0.04$\pm$0.02 & 
        -0.30$\pm$0.07 & 0.37$\pm$0.04 & 0.10$\pm$0.03 & 
        \textbf{-0.62$\pm$0.05} & \underline{0.67$\pm$0.05} & \underline{0.38$\pm$0.05} &
        -0.61$\pm$0.11 & \underline{0.55$\pm$0.04} & 0.32$\pm$0.08  \\  
        &ProbVLM & -0.79$\pm$0.05 & 0.52$\pm$0.04 & 0.38$\pm$0.04 & 
        -0.72$\pm$0.04 & 0.21$\pm$0.02 & 0.14$\pm$0.02 & 
        -0.33$\pm$0.05 & 0.46$\pm$0.04 & 0.15$\pm$0.02 &
        \underline{-0.78$\pm$0.03} & 0.47$\pm$0.03 & \underline{0.36$\pm$0.03}  \\  
        &GroVE & \textbf{-0.87$\pm$0.06} & \textbf{0.85$\pm$0.04} & \textbf{0.77$\pm$0.05} &
        \textbf{-0.90$\pm$0.03} & \textbf{0.88$\pm$0.04} & \textbf{0.79$\pm$0.02} & \underline{-0.61$\pm$0.07} & \textbf{0.75$\pm$0.04} & \textbf{0.46$\pm$0.06} &
        \textbf{-0.88$\pm$0..04} & \textbf{0.81$\pm$0.01} & \textbf{0.70$\pm$0.03}   \\  
        \midrule
        \multirow{7}{*}{\rotatebox{90}{\textbf{Text to Image}}} 
        & Deterministic & \underline{-0.90$\pm$0.00} & \textbf{0.80$\pm$0.00} & \textbf{0.73$\pm$0.00} & 
        \underline{-0.80$\pm$0.00} & 0.76$\pm$0.00 & \underline{0.61$\pm$0.00} & 
        0.60$\pm$0.00 & 0.12$\pm$0.00 & -0.06$\pm$0.00 &
        -0.30$\pm$0.00 & 0.17$\pm$0.00 & 0.05$\pm$0.00  \\  
        &TTDA & 0.08$\pm$0.06 & 0.02$\pm$0.06 & 0.01$\pm$0.01 &
        -0.61$\pm$0.06 & 0.20$\pm$0.05 & 0.12$\pm$0.04 &
        -0.53$\pm$0.05 & \textbf{0.64$\pm$0.03} & 0.32$\pm$0.02 &
        0.04$\pm$0.01 & 0.04$\pm$0.00 & 0.00$\pm$0.02  \\  
        &PFE & -0.68$\pm$0.07 & 0.56$\pm$0.05 & 0.38$\pm$0.06 & 
        0.33$\pm$0.04 & 0.52$\pm$0.02 &-0.16$\pm$0.02 & 
        -0.32$\pm$0.07 &0.34$\pm$0.02 & 0.09$\pm$0.02 &
        0.21$\pm$0.04 & 0.43$\pm$0.02 & -0.10$\pm$0.02  \\  
        &PCME & 0.18$\pm$0.08 & 0.42$\pm$0.02 & -0.07$\pm$0.04 & 
        0.86$\pm$0.04 & \textbf{0.84$\pm$0.03} & -0.74$\pm$0.05 & 
        0.57$\pm$0.04 & 0.05$\pm$0.00 & -0.03$\pm$0.00 &
        0.72$\pm$0.05 & 0.45$\pm$0.03 & -0.29$\pm$0.03  \\  
        &PCME++ & -0.13$\pm$0.04 & 0.06$\pm$0.03 & 0.01$\pm$0.00 & 
        0.02$\pm$0.07 & 0.38$\pm$0.02 & 0.01$\pm$0.03 & 
        -0.28$\pm$0.05 & 0.02$\pm$0.01 & 0.01$\pm$0.00 &
        0.12$\pm$0.07 & \underline{0.47$\pm$0.02} & -0.06$\pm$0.04  \\  
        &ProbVLM & -0.54$\pm$0.03 & 0.68$\pm$0.07 & 0.34$\pm$0.03 & 
        0.09$\pm$0.02 & 0.11$\pm$0.04 & -0.01$\pm$0.00 & 
        \textbf{-0.92$\pm$0.03} & 0.52$\pm$0.05 & \underline{0.48$\pm$0.04} &
        \underline{-0.60$\pm$0.03} & 0.16$\pm$0.06 & \underline{0.10$\pm$0.03}  \\  
        &GroVE & \textbf{-0.92$\pm$0.04} & \underline{0.74$\pm$0.04} & \underline{0.66$\pm$0.04} & 
        \textbf{-0.81$\pm$0.02} & \underline{0.81$\pm$0.01} & \textbf{0.65$\pm$0.02} & 
        \underline{-0.78$\pm$0.07} & \underline{0.60$\pm$0.02} & \textbf{0.49$\pm$0.05} &
        \textbf{-0.62$\pm$0.08} & \textbf{0.66$\pm$0.04} & \textbf{0.43$\pm$0.06} \\  
    
     \bottomrule
        
    \end{tabular}
    \end{adjustbox}
    \caption{ \textbf{Uncertainty calibration for cross-modal retrieval using CLIP.} GroVE  demonstrates superior performance in uncertainty calibration in majority cases compared to baseline models. The best scores are highlighted in bold and the second-best scores are underlined.
    }
    \label{tab:clip_unc_results}
\end{table*}

\textbf{Uncertainty Quantification.} When an embedding $\mathbf{z}_*$ belongs to an ambiguous input, the uncertainty associated with the posterior distribution $p(\mathbf{x}_*|\mathbf{z}_*, \mathbf{z}_\mathcal{M})$ increases. 
This uncertainty is propagated to the predictive distribution, which can be written as:
$p(\mathbf{g}_*) = \int p(\mathbf{g}_*|\mathbf{x}_*)p(\mathbf{x}_*|\mathbf{z}_*, \mathbf{z}_\mathcal{M})d\mathbf{x}_*$. 
Here, $p(\mathbf{g}_*|\mathbf{x}_*)$ is a multivariate Gaussian distribution that describes the function values at the fixed point $\mathbf{x}_*$. Thus, a large uncertainty in $\mathbf{x}_*$ increases the variance of the predictive distribution $p(\mathbf{g}_*|\mathbf{z}_*, \mathbf{z}_\mathcal{M})$. The uncertainty is captured by $\boldsymbol{\hat{\Sigma}}_*$, which accounts for variance contributions from both the latent space uncertainty and inherent noise in the model’s predictions. The final uncertainty is obtained by averaging the uncertainty values across all dimensions in $\boldsymbol{\hat{\Sigma}}_*$.

\begin{table*}[h]
    \begin{adjustbox}{width=2\columnwidth, center}
    \centering
\begin{tabular}{@{}p{-1cm}lcccccccccccc@{}}
    \toprule
&\multirow{2}{*}{Method} & \multicolumn{3}{c}{Flickr} & \multicolumn{3}{c}{COCO} & \multicolumn{3}{c}{CUB} & \multicolumn{3}{c}{Flowers} \\ \cmidrule(lr){3-5} \cmidrule(lr){6-8} \cmidrule(lr){9-11} \cmidrule(lr){12-14} 
 & & $S \downarrow$ & $R^2 \uparrow$ & $-SR^2 \uparrow$ & $S \downarrow$ & $R^2 \uparrow$ & $-SR^2 \uparrow$ & $S \downarrow$ & $R^2 \uparrow$ & $-SR^2 \uparrow$ & $S \downarrow$ & $R^2 \uparrow$ & $-SR^2 \uparrow$ \\ \midrule
 \multirow{7}{*}{\rotatebox{90}{\textbf{Image to Text}}}
        & Deterministic & 
        \underline{-0.70$\pm$0.00} & \textbf{0.78$\pm$0.00} & \textbf{0.55$\pm$0.00} &
        \underline{-0.80$\pm$0.00} & \textbf{0.84$\pm$0.00} & \underline{0.67$\pm$0.00} &
        0.50$\pm$0.00 & 0.13$\pm$0.00 & -0.07$\pm$0.00 &
        -0.20$\pm$0.00 & 0.05$\pm$0.00 & 0.01$\pm$0.00 \\
        &TTDA & 
        -0.68$\pm$0.04 & 0.27$\pm$0.03 & 0.19$\pm$0.02 &
        -0.72$\pm$0.05 & 0.48$\pm$0.04 & 0.32$\pm$0.04 &
        -0.70$\pm$0.05 & 0.50$\pm$0.02 & 0.33$\pm$0.03 &
        -0.63$\pm$0.04 & 0.24$\pm$0.03 & 0.13$\pm$0.02 \\
        &PFE & 
        0.12$\pm$0.06 & 0.37$\pm$0.02 & -0.04$\pm$0.02 & 
        0.04$\pm$0.00 & 0.32$\pm$0.05 & 0.00$\pm$0.00 & 
        0.56$\pm$0.06 & 0.53$\pm$0.04 & 0.32$\pm$0.03 &
        0.13$\pm$0.04 & 0.02$\pm$0.03 & -0.01$\pm$0.03 \\        
        &PCME & 
        -0.31$\pm$0.06 & 0.17$\pm$0.04 & 0.05$\pm$0.03 & 
        -0.62$\pm$0.03 & 0.24$\pm$0.02 & 0.14$\pm$0.02 & 
        -0.64$\pm$0.03 & \textbf{0.63}$\pm$0.03 & 0.38$\pm$0.03 &
        0.08$\pm$0.03 & 0.25$\pm$0.04 & -0.03$\pm$0.02 \\          
        &PCME++ & 
        -0.68$\pm$0.03 & 0.26$\pm$0.03 & 0.18$\pm$0.03 & 
        -0.69$\pm$0.04 & 0.50$\pm$0.04 & 0.34$\pm$0.03 & 
        \underline{-0.71$\pm$0.04} & 0.57$\pm$0.03 & 0.40$\pm$0.03 &
        \underline{-0.69$\pm$0.06} & 0.53$\pm$0.02 & 0.37$\pm$0.02 \\  
        &ProbVLM & 
        0.03$\pm$0.07 & 0.48$\pm$0.02 & 0.02$\pm$0.02 & 
        -0.61$\pm$0.03 & 0.50$\pm$0.04 & 0.30$\pm$0.03 & 
        -0.68$\pm$0.06 & \underline{0.60$\pm$0.03} & \underline{0.42$\pm$0.04} &
        -0.67$\pm$0.00 & \underline{0.65$\pm$0.02} & \underline{0.46$\pm$0.01} \\  
        &GroVE & 
        \textbf{-0.72$\pm$0.03} & \underline{0.74$\pm$0.02} & \underline{0.51$\pm$0.03} & 
        \textbf{-0.93$\pm$0.05} & \underline{0.76$\pm$0.03} & \textbf{0.68$\pm$0.03} & \textbf{-0.89$\pm$0.04} & \underline{0.60$\pm$0.04} & \textbf{0.54$\pm$0.02} &
        \textbf{-0.72$\pm$0.07} & \textbf{0.72$\pm$0.06} & \textbf{0.50$\pm$0.05}  \\  
    
     \midrule
        \multirow{7}{*}{\rotatebox{90}{\textbf{Text to Image}}} 
        & Deterministic & 
        \underline{-0.90$\pm$0.00} & \underline{0.88$\pm$0.00} & \underline{0.79$\pm$0.00} &
        \textbf{-0.90$\pm$0.00} & \textbf{0.88$\pm$0.00} & \textbf{0.80$\pm$0.00} &
        0.40$\pm$0.00 & 0.06$\pm$0.00 & 0.02$\pm$0.00 &
        -0.10$\pm$0.00 & 0.00$\pm$0.00 & 0.00$\pm$0.00 \\  
        &TTDA & 
        -0.37$\pm$0.03 & 0.35$\pm$0.04 & 0.14$\pm$0.03 &
        0.41$\pm$0.06 &  0.00$\pm$0.01 & 0.00$\pm$0.03 & 
        -0.68$\pm$0.05 & 0.48$\pm$0.06 & 0.34$\pm$0.05 &
        0.09$\pm$0.03 & \underline{0.43$\pm$0.02} & -0.04$\pm$0.04 \\  
        &PFE & 
        -0.58$\pm$0.04 & 0.50$\pm$0.03 & 0.30$\pm$0.04 & 
        0.11$\pm$0.05 & 0.15$\pm$0.04 & -0.02$\pm$0.02 & 
        \underline{-0.78$\pm$0.03} & \underline{0.58$\pm$0.02} & \underline{0.47$\pm$0.02} &
        -0.23$\pm$0.06 & 0.01$\pm$0.03 & 0.00$\pm$0.03 \\  
        &PCME & 
        -0.12$\pm$0.04 & 0.50$\pm$0.02 & 0.05$\pm$0.01 & 
        0.62$\pm$0.03 & 0.42$\pm$0.06 & -0.25$\pm$0.03 & 
        -0.68$\pm$0.04 & \underline{0.58$\pm$0.03} & 0.41$\pm$0.02 &
        \underline{-0.31$\pm$0.03} & 0.26$\pm$0.03 & \underline{0.08$\pm$0.04} \\  
        &PCME++& 
        -0.72$\pm$0.06 & 0.30$\pm$0.04 & 0.21$\pm$0.04 & 
        -0.48$\pm$0.03 & 0.31$\pm$0.02 & 0.15$\pm$0.03 &
        -0.12$\pm$0.08 & 0.00$\pm$0.04 & 0.00$\pm$0.02 &
        -0.20$\pm$0.07 & 0.06$\pm$0.06 & 0.01$\pm$0.03 \\  
        &ProbVLM & 
        -0.56$\pm$0.04 & 0.50$\pm$0.03 & 0.31$\pm$0.03 & 
        -0.12$\pm$0.05 & \underline{0.48$\pm$0.04} & 0.05$\pm$0.04 & 
        -0.43$\pm$0.03 & 0.50$\pm$0.02 & 0.18$\pm$0.03 &
        0.38$\pm$0.03 & 0.02$\pm$0.04 & -0.01$\pm$0.04 \\  
        &GroVE & 
        \textbf{-0.92$\pm$0.04} & \textbf{0.90$\pm$0.03} & \textbf{0.81$\pm$0.04} & \underline{-0.62$\pm$0.03} & 0.36$\pm$0.06 & \underline{0.22$\pm$0.04} & \textbf{-0.89$\pm$0.02} & \textbf{0.75$\pm$0.03} & \textbf{0.74$\pm$0.03} &
        \textbf{-0.73$\pm$0.03} & \textbf{0.62$\pm$0.04} & \textbf{0.44$\pm$0.02} \\  
    
     \bottomrule
        
    \end{tabular}
    \end{adjustbox}
    \caption{ \textbf{Uncertainty calibration for cross-modal retrieval using BLIP.} GroVE  demonstrates superior performance in uncertainty calibration in majority cases compared to baseline models. The best scores are highlighted in bold and the second-best scores are underlined.
    }
    \label{tab:blip_unc_results}
\end{table*}

\section{Experiments} \label{sec:expts}
\subsection{Experimental Setup}
\textbf{Baselines and Datasets.} We evaluate GroVE against six baseline methods: Deterministic, TTDA~\citep{ayhan2018test}, PFE~\citep{shi2019probabilistic}, PCME~\citep{chun2021probabilistic}, PCME++~\citep{chun2023improved}, and ProbVLM~\citep{upadhyay2023probvlm}, using two VLMs—CLIP~\citep{radford2021learning} and BLIP~\citep{li2022blip}—with a focus on uncertainty calibration for downstream tasks. In the deterministic approach, uncertainty is quantified by the cosine distance between the image and text embeddings derived from the VLM. 
While PFE, PCME and PCME++ are methods to learn probabilistic embeddings for pre-training VLMs, we follow \cite{upadhyay2023probvlm}, and adapt them to work in a post-hoc manner. 
The similarity ranking between probabilistic image and text embeddings is determined by the Wasserstein distance, with embeddings ranked based on the increasing distance, while for the deterministic embeddings, the cosine similarity is used.
The implementation details for these methods are provided in Appendix~\ref{sec:baselines}. 
The methods are evaluated on MS-COCO~\citep{lin2014microsoft}, Flickr30k~\citep{young2014image}, CUB-200-2011~\citep{wah2011caltech} and Oxford Flowers 102~\citep{nilsback2008automated} for cross-modal retrieval, and VQA2.0~\citep{goyal2017making} for visual question answering. The captions for the CUB and Flowers datasets were obtained from \cite{reed2016learning}.

\textbf{Evaluation Metrics.} The cross-modal retrieval is evaluated using the Recall@1 metric. For evaluating uncertainty calibration, we adopt the metrics used in \cite{upadhyay2023probvlm}, which computes the Spearman rank correlation (S) between different uncertainty levels and Recall@1, $R^2$ value for the regression between the uncertainty levels and the Recall@1 performance, and their product $-SR^2$.  For an ideal model, the Recall@1 score should decrease with increasing uncertainty, resulting in a $S$ value of -1. A higher $R^2$ score indicates that with increasing uncertainty levels, the model's performance declines linearly. A higher $-SR^2$ score implies better uncertainty calibration, reflecting both a strong negative correlation and a monotonic decrease in performance with increasing uncertainty. 
VQA is evaluated using the soft voting accuracy of 10 human-annotated answers~\citep{goyal2017making}.
Calibration is evaluated by the Expected Calibration Error (ECE) score between the model's confidence and the soft voting accuracy. Model confidence is computed by first predicting an uncertainty score $u(a)$ for each candidate answer $a$, and then applying a softmax function over these uncertainty scores. The model confidence is given by $\text{conf}(a)= 1 - \text{softmax}(u(a))$.

\textbf{Implementation details.}
The experiments on CLIP were conducted using the ViT-B/32 model as the image encoder, with $D=512$. For BLIP, we adopt the ViT-B architecture as the image encoder. We trained the GPs with a latent space dimension of $Q=5$ for MS-COCO, Flickr30k and VQA2.0, and $Q=10$ for CUB and Flowers alongside trade-off parameters $\lambda_1=0.01$ and $\lambda_2=400$ and 250 inducing points, determined through grid search.
The models were implemented with GPyTorch~\cite{gardner2018gpytorch}, and 
trained for 200 epochs using the Adam optimizer with a learning rate of $1e^{-5}$ and a batch size of 64. The detailed implementation, including data processing and hyper-parameter tuning is provided in Appendix~\ref{sec:app_dataset} and \ref{sec:app_hyper} respectively.
  
\subsection{Uncertainty Calibration in Cross-modal Retrieval} \label{sec:unc_calibration}
\textbf{Quantitative Results.} The uncertainty calibration results for CLIP and BLIP is provided in Table~\ref{tab:clip_unc_results} and Table~\ref{tab:blip_unc_results} respectively. We observe that GroVE demonstrates superior performance across all four datasets in both image-to-text and text-to-image retrieval tasks, outperforming other methods in most cases.  
A high $-SR^2$ value for GroVE indicates that the model maintains strong performance when uncertainty is low, and the decline in performance is well-aligned with increasing uncertainty scores, indicating effective uncertainty calibration. 
Interestingly, the Deterministic baseline also performs competitively on the Flickr30k and MS-COCO datasets. This is because the VLMs were trained on datasets with common real-world objects, well-represented in these datasets, allowing the deterministic approach to benefit from familiar image-text pair contexts.
However, on fine-grained datasets like CUB and Flowers, which are less represented in the training data, it exhibits a noticeable drop in performance. In these cases, the probabilistic methods outperform the deterministic approach, with GroVE consistently leading across both common object and fine-grained datasets. 

\begin{figure}[tb]
    \centering
    \includegraphics[width=0.75\linewidth]{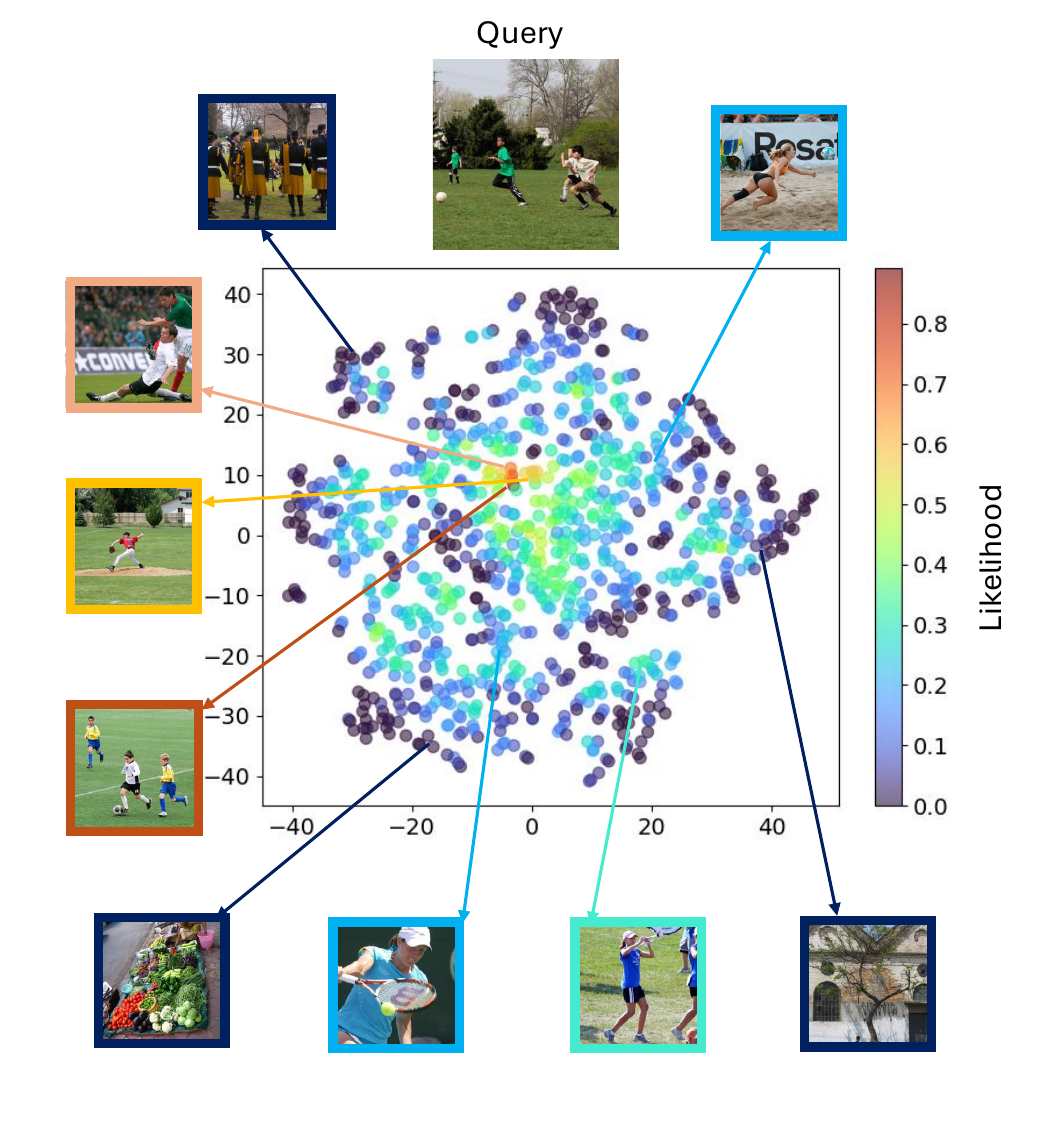}
    \vspace{-15pt}
    \caption{Given a probabilistic query image embedding from COCO, the plot shows a t-SNE visualization of Flickr30k embeddings, colored by their likelihood of belonging to the query distribution. Sample images are shown in colored boxes, where images with high likelihoods share similar semantic and visual content to the query.}
    \label{fig:likelihood}
\end{figure}

\textbf{Qualitative Results. }Given a query image from MS-COCO, we obtain its probabilistic embedding using GroVE. Using the distribution of this embedding, we compute the likelihood of each image in the Flickr30k dataset. Figure~\ref{fig:likelihood} shows a t-SNE plot of the mean embeddings on Flickr30k, colored by likelihood scores. The query image depicts children playing on a field. We observe that the images with the highest likelihood scores, share similar semantic content, such as scenes of people playing in fields. In contrast, images with lower likelihood values (close to 0.0) show little to no semantic or visual similarity to the query. 

Additional results containing the retrieval performance, zero-shot performance, calibration plots and qualitative analysis is provided in Appendix.~\ref{sec:app_cross-modal} and \ref{sec:app_zs}. 

\subsection{Active Learning}

\begin{figure}[h]
\hspace{-15pt}
    \centering
    \begin{minipage}{0.2\columnwidth} 
        \centering
        \rotatebox{90}{\includegraphics[height=0.7\textwidth]{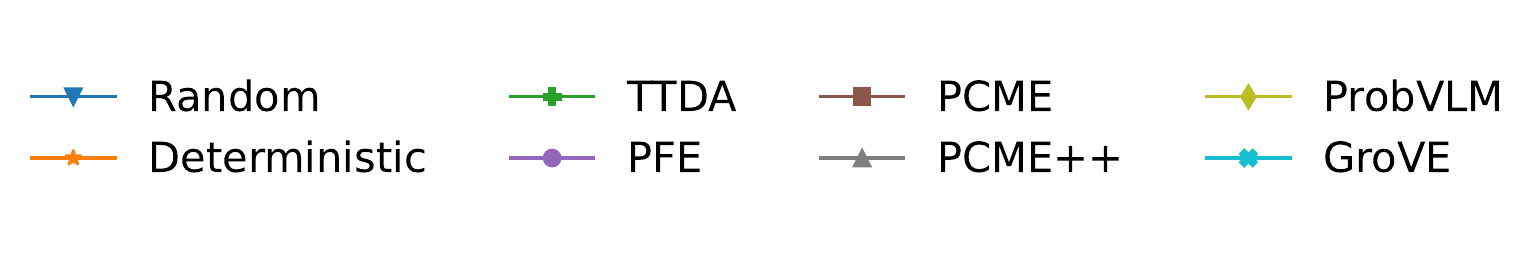}} 
    \end{minipage}%
    \begin{minipage}{0.75\columnwidth} 
        \centering
        \includegraphics[width=\columnwidth]{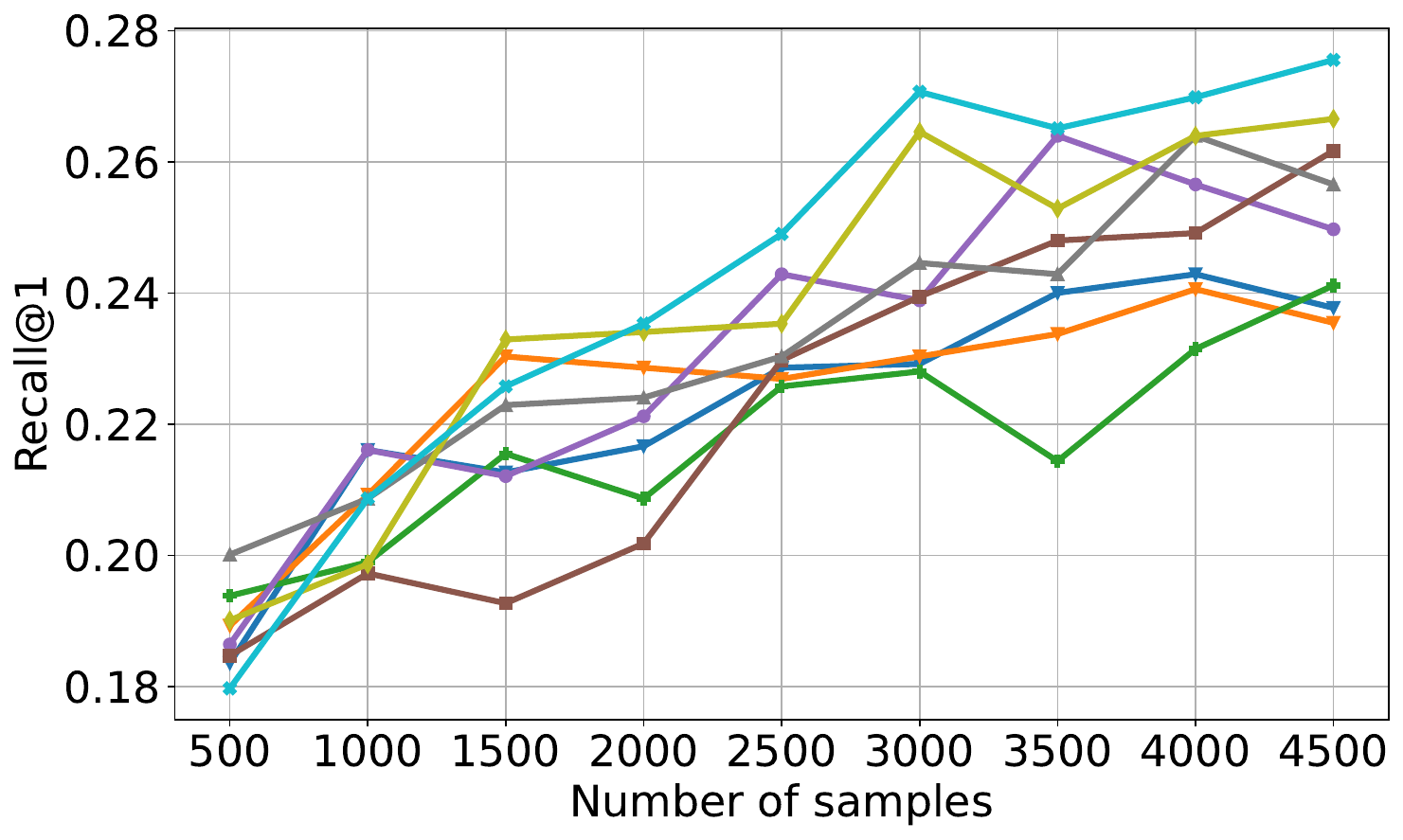} \\
        \vspace{0.5cm} 
        \includegraphics[width=\columnwidth]{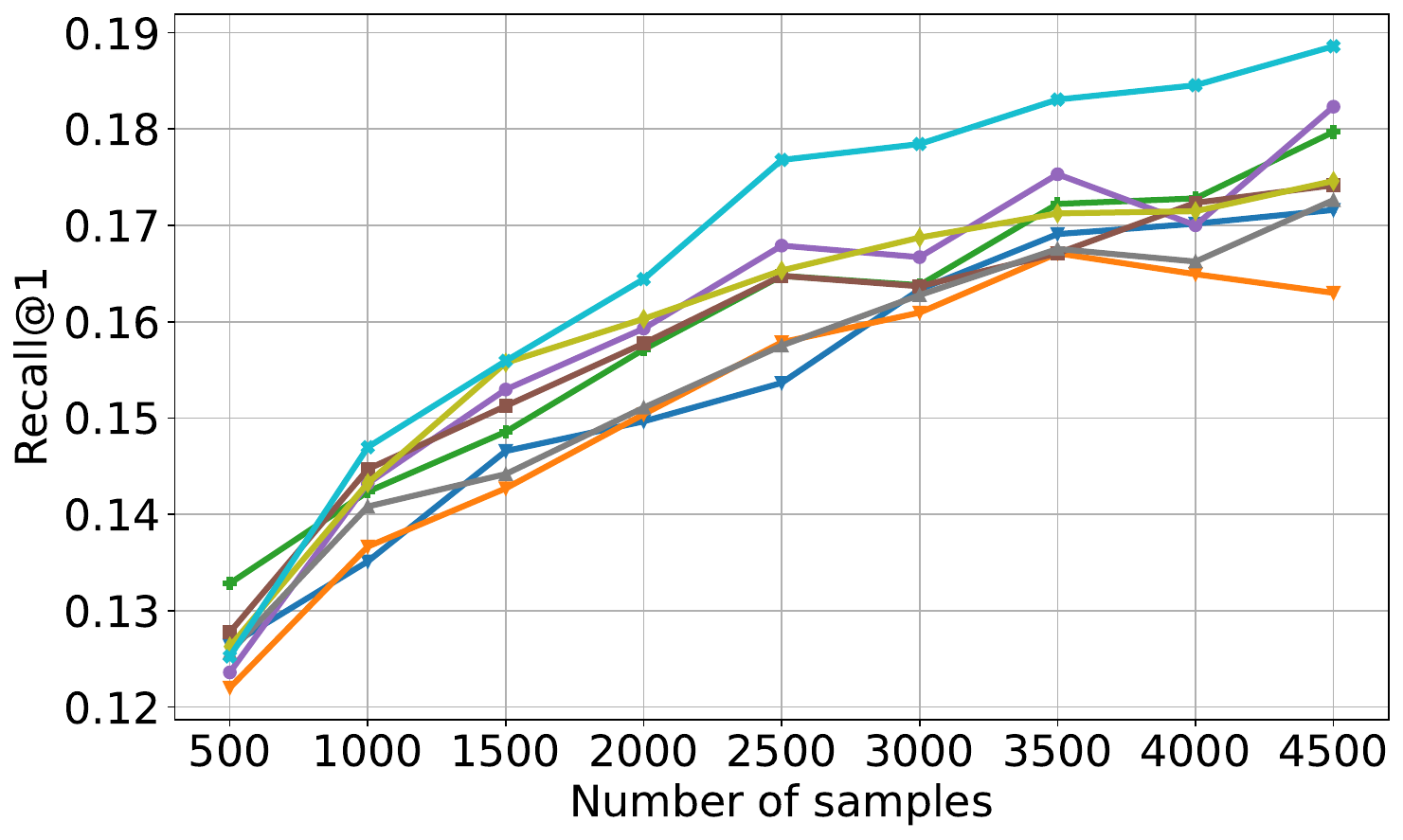}
    \end{minipage}
        \caption{\textbf{Active Learning.} The results highlight GroVE's ability to effectively leverage uncertainty estimates to guide sample selection, outperforming the baselines on both image-to-text (left) and text-to-image (right) retrieval.}
    \label{fig:active_learning}
\end{figure}




The objective of this experiment is to fine-tune the CLIP model on the CUB dataset with limited labeled data. We estimate the uncertainty of image and text embeddings to identify the most uncertain samples from the unlabeled CUB dataset, which are labeled for fine-tuning. For methods using auxiliary models, we derive uncertainty estimates from models trained on COCO. We sample the top 500 uncertain samples at each step for fine-tuning with contrastive loss. A random sampling baseline is also included.
Figure~\ref{fig:active_learning} provides the Recall@1 scores achieved in relation to the number of samples used for fine-tuning the CLIP model. GroVE achieves consistently better performance compared to others, demonstrating that its uncertainty estimates effectively identify the informative samples for active learning.

\subsection{Uncertainty in Few-shot Setting}
In this experiment, we explore a practical scenario where labeled training data is scarce. To simulate this, we create a few-shot dataset by randomly selecting three images and their corresponding text descriptions from 150 classes of the CUB dataset as done by \cite{verma2021towards}. The probabilistic adapters were trained on this dataset using embeddings obtained from CLIP, and the uncertainty calibration was evaluated for cross-modal retrieval. Table~\ref{tab:few_shot} shows the $-SR^2$ scores obtained for the baselines and GroVE with different numbers of inducing points as well as exact GP models, where training and inference is performed without approximations \citep{williams2006gaussian}. The results show that while the calibration performance improves as the number of inducing points increases, GroVE consistently outperforms the baselines in terms of calibration quality. The best performance was achieved with exact GP models and no approximation.
A comparison of the retrieval performance and the inference time is provided in Appendix~\ref{sec:app_few_shot}.

\begin{table}[tb]
    \begin{adjustbox}{width=0.65\columnwidth, center}
    \centering
    \begin{tabular}{ccc}\\
    \toprule
    {Method}&{Image to Text} & {Text to Image} \\ \midrule 
     TTDA &  0.03$\pm$0.06 & 0.01$\pm$0.04\\
     PFE  & -0.29$\pm$0.02 & -0.22$\pm$0.03\\
     PCME &  0.04$\pm$0.03 & 0.14$\pm$0.02\\
     PCME++ &  0.27$\pm$0.03 & 0.01$\pm$0.04\\
     ProbVLM  & -0.12$\pm$0.04 & -0.43$\pm$0.04\\
     \midrule
     GroVE (M=50) & 0.24$\pm$0.03 & 0.22$\pm$0.03\\
     GroVE (M=150) & 0.36$\pm$0.04 & 0.31$\pm$0.02\\
     GroVE (M=250) &  0.35$\pm$0.03 &  0.31$\pm$0.03\\
     GroVE (exact GP)  & \textbf{0.39$\pm$0.03} & \textbf{0.36$\pm$0.02} \\
     \bottomrule
    \end{tabular}
    \end{adjustbox}
     \caption{\textbf{Few-shot uncertainty calibration.}  GroVE outperforms other baselines, achieving superior uncertainty calibration in few-shot settings in terms of -$SR^2$ ($\uparrow$).}
    \label{tab:few_shot}
\end{table}

\subsection{Uncertainty Calibration for VQA}
Table~\ref{tab:vqa_expts} shows the accuracy and ECE scores obtained for VQA2.0 using BLIP as the VLM. All the baselines achieve similar accuracy scores, with deterministic achieving the best accuracy. When evaluated for confidence calibration, GroVE achieves the lowest ECE score.

\begin{table}[tb]
    \begin{adjustbox}{width=0.55\columnwidth, center}
    \centering
    \begin{tabular}{ccc}\\
    \toprule
    Method & Accuracy $\uparrow$ & ECE $\downarrow$ \\ \midrule
     Determinsitic & \textbf{78.20} & 0.56  \\
     TTDA  & \underline{77.67$\pm$2.23} & \underline{0.48$\pm$0.06} \\
     PFE  & 76.34$\pm$1.98 & 0.65$\pm$0.02\\
     PCME & 77.25$\pm$1.76 & 0.64$\pm$0.01\\
     PCME++ & 77.53$\pm$1.71 & 0.64$\pm$0.02 \\
     ProbVLM & 76.66$\pm$1.13 & 0.69$\pm$0.01\\
     GroVE & 77.48$\pm$2.15 & \textbf{0.24$\pm$0.04} \\
     \bottomrule
    \end{tabular}
    \end{adjustbox}
     \caption{\textbf{Results for VQA.} While all methods achieve similar accuracy (with the deterministic model performing best), GroVE reaches the best calibration performance in terms of ECE ($\downarrow$). }
    \label{tab:vqa_expts}
\end{table}

\subsection{Ablation Analysis} \label{sec:ablation}


\begin{table}[tb]
    \begin{adjustbox}{width=0.85\columnwidth, center}
    \centering
    \begin{tabular}{ccccc}\\
    \toprule
    \multirow{2}{*}{Kernel}& \multicolumn{2}{c}{Image to Text} & \multicolumn{2}{c}{Text to Image} \\ \cmidrule(lr){2-3} \cmidrule(lr){4-5}
      &  COCO & CUB & COCO & CUB\\ \midrule
     RBF & \textbf{0.79$\pm$0.02} & 0.46$\pm$0.06 & \textbf{0.65$\pm$0.02} & \textbf{0.49$\pm$0.05} \\
     Mat\'ern ($\nu=1.5$) & 0.27$\pm$0.03 & \textbf{0.47$\pm$0.05} & 0.41$\pm$0.04 & 0.22$\pm$0.04\\
     Mat\'ern ($\nu=2.5$) & 0.52$\pm$0.05 & 0.38$\pm$0.04 & 0.43$\pm$0.04 & 0.12$\pm$0.05\\
     Cosine Similarity & 0.46$\pm$0.04 & 0.39$\pm$0.03 & 0.35$\pm$0.03 & 0.30$\pm$0.02\\
     \bottomrule
    \end{tabular}
    \end{adjustbox}
     \caption{\textbf{Ablation on choice of GP kernel.}  GroVE achieves the best performance on MS-COCO and CUB-200-2011 with the RBF kernel.}
    \label{tab:ablation_kernels}
\end{table}

\begin{figure}
    \centering
    \includegraphics[width=0.48\linewidth]{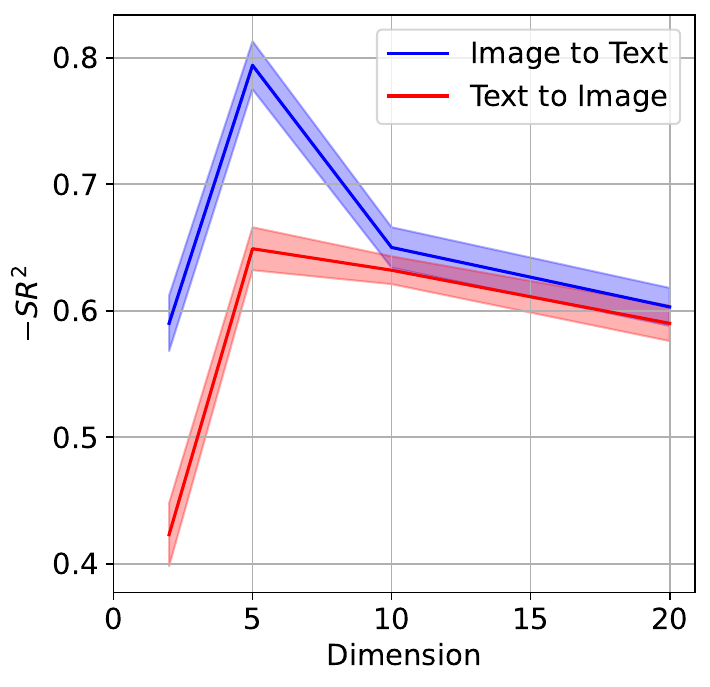} 
    \includegraphics[width=0.48\linewidth]{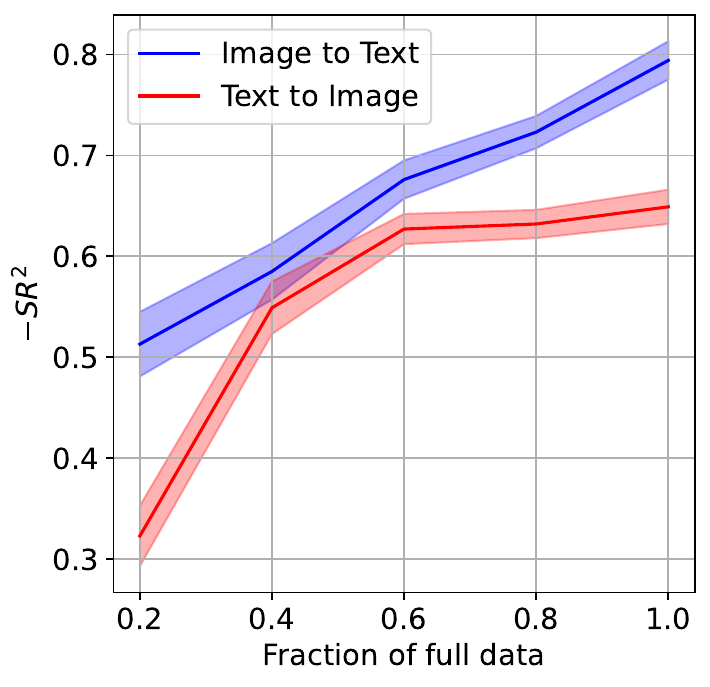}
    \caption{Ablation using MS-COCO: (i) \textbf{latent space dimension} (left). Low latent space dimensions results in loss of information, while higher dimensions results in performance degradation due to over-fitting. (ii) \textbf{dataset size for training} (right). GroVE achieves good performance with just 60\% of the total training dataset.}
    \label{fig:lat_dim_data_sub}
\end{figure}

\textbf{GP Kernel.} 
We evaluate the performance of the RBF, Mat\'ern ($\nu = 1.5$ and $2.5$, where $\nu$ is the smoothness parameter) and the cosine similarity kernel on GroVE's performance on the MS-COCO and CUB data. From Table~\ref{tab:ablation_kernels}, the RBF kernel achieves superior performance compared to the other kernels across both datasets, with improvements up to 53\%. 
The kernels are defined in Appendix~\ref{sec:def_kernel}.

\textbf{Latent Space Dimension. } We investigate the influence of the latent space dimension $Q$ on GroVE's performance using the MS-COCO dataset. Figure~\ref{fig:lat_dim_data_sub} (left) presents the $-SR^2$ scores for various values of $Q$. Low values of $Q$ lead to information loss, which compromises the model's ability to capture complex patterns in the data. Conversely, high values of $Q$ result in overfitting and make the model more challenging to optimize, resulting in a performance decline. The optimal performance was observed when $Q=5$.

\textbf{Dataset Size. } We study the impact of the dataset size on GroVE's performance by training it on various fractions of the COCO training dataset. As shown in Figure~\ref{fig:lat_dim_data_sub} (right), the model achieves good performance when trained on 60\% of the full dataset. While the uncertainty calibration performance for text-to-image retrieval plateaus beyond this point, the performance for image-to-text retrieval continues to improve almost linearly as more data is utilized.

\begin{figure}
    \centering
    \includegraphics[height=4.5cm, width=0.52\linewidth]{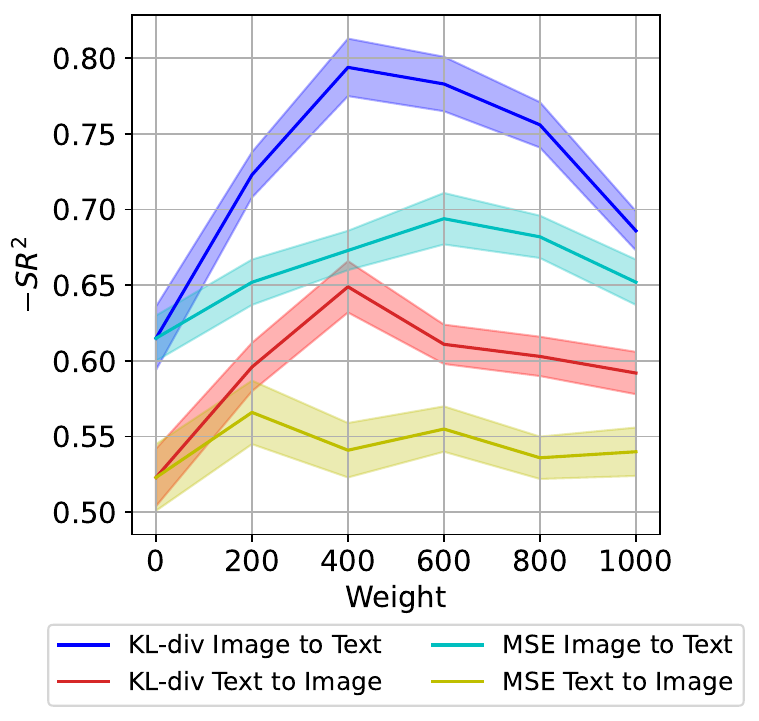} 
    \includegraphics[height=4.5cm, width=0.47\linewidth]{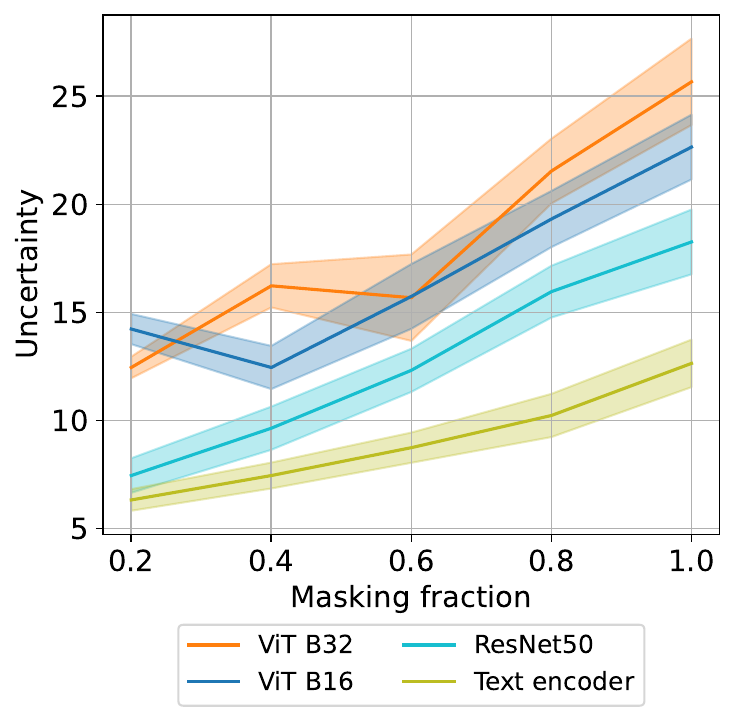}
    \caption{Ablation using MS-COCO: (i) \textbf{trade-off parameter} (left). KL-divergence improves uncertainty calibration with optimal performance at $\lambda_2=400$, with $\lambda_1=0.01$. (ii) \textbf{noisy data} (right). With increasing amount of noise in the input data, GroVE predicts higher uncertainty.}
    \label{fig:reg_masking}
\end{figure}

\textbf{Cross-modal Alignment. } We compare GroVE's KL-divergence-based alignment loss with the MSE loss-based regularization used in \cite{song2017multimodal}. The authors use GPLVM for cross-modal retrieval, regularizing the latent space with the loss function $||k_I - S||^2 + ||k_T - S||^2$, where $k_I$ and $k_T$ are the GP covariance matrices, and $S$ is the latent space similarity matrix. For comparison, we replace $\mathcal{L}_{KL}$ in our method with the MSE loss and experiment with different values of the trade-off parameter $\lambda_2$, maintaining $\lambda_1 = 0.01$. As shown in Figure~\ref{fig:reg_masking} (left), the KL-divergence alignment loss improves uncertainty calibration performance by up to 23\%, with the best performance at $\lambda_2 = 400$.
Additionally, we evaluate the cross-modal alignment KL-divergence loss against other widely-used probabilistic distance metrics: Jensen-Shannon (JS) divergence and Wasserstein-2 distance. The results in Table~\ref{tab:ablation_kl} indicate that while all metrics perform similarly, KL-divergence offers a slight edge. The distance metrics are defined in Appendix~\ref{sec:def_dist}.


\begin{table}[]
    \begin{adjustbox}{width=0.85\columnwidth, center}
    \centering
    \begin{tabular}{ccccc}\\
    \toprule
    \multirow{2}{*}{Kernel}& \multicolumn{2}{c}{Image to Text} & \multicolumn{2}{c}{Text to Image} \\ \cmidrule(lr){2-3} \cmidrule(lr){4-5}
      &  COCO & CUB & COCO & CUB\\ \midrule
     KL-Divergence & \textbf{0.79$\pm$0.02} & 0.46$\pm$0.06 & \textbf{0.65$\pm$0.02} & \textbf{0.49$\pm$0.05} \\
     JS-Divergence & 0.70$\pm$0.04 & \textbf{0.48$\pm$0.02} & 0.59$\pm$0.03 & 0.44$\pm$0.05\\
     Wasserstein-2 & 0.59$\pm$0.04 & 0.39$\pm$0.04 & 0.60$\pm$0.02 & 0.43$\pm$0.04\\
     \bottomrule
    \end{tabular}
    \end{adjustbox}
     \caption{\textbf{Ablation on probabilistic distance metric.} GroVE performs better for cross-modal alignment using KL-Divergence compared to other metrics.}
    \label{tab:ablation_kl}
\end{table}

\textbf{Noisy Data.} To evaluate the performance of GroVE against noisy inputs, we systematically introduce increasing levels of masking to both the input images and texts. This analysis employs several CLIP image encoder backbones, including ViT-B/32, ViT-B/16, and ResNet50, along with CLIP's text encoder. The results, presented in Figure~\ref{fig:reg_masking} (right), indicate that as the noise level increases, the uncertainty predicted by GroVE rises steadily as desired.

\section{Conclusion}
This paper introduces GroVE, a post-hoc approach for generating probabilistic embeddings from frozen, pre-trained VLMs to model input data ambiguities. GroVE leverages the GPLVM framework, utilizing GP models to learn a shared, low-dimensional latent space that aligns visual and textual representations. By mapping into this latent space, the GP models generate probabilistic embeddings that provide a measure of uncertainty in the predictions. GroVE demonstrates state-of-the-art performance in uncertainty calibration for cross-modal retrieval, active learning and VQA.  One limitation of GroVE is the it is computationally expensive compared to the neural network based methods (see Appendix.~\ref{sec:app_few_shot}). In latency-sensitive scenarios, such as real-time applications, neural network-based stochastic models like Neural Processes~\citep{garnelo2018neural} offer a viable alternative to GPs. Future work will focus on assessing their uncertainty calibration performance for VLMs.

\bibliography{uai2025-template}

\begin{thebibliography}{64}
\providecommand{\natexlab}[1]{#1}
\providecommand{\url}[1]{\texttt{#1}}
\expandafter\ifx\csname urlstyle\endcsname\relax
  \providecommand{\doi}[1]{doi: #1}\else
  \providecommand{\doi}{doi: \begingroup \urlstyle{rm}\Url}\fi

\bibitem[Abdar et~al.(2021)Abdar, Pourpanah, Hussain, Rezazadegan, Liu, Ghavamzadeh, Fieguth, Cao, Khosravi, Acharya, et~al.]{abdar2021review}
Moloud Abdar, Farhad Pourpanah, Sadiq Hussain, Dana Rezazadegan, Li~Liu, Mohammad Ghavamzadeh, Paul Fieguth, Xiaochun Cao, Abbas Khosravi, U~Rajendra Acharya, et~al.
\newblock A review of uncertainty quantification in deep learning: Techniques, applications and challenges.
\newblock \emph{Information fusion}, 76:\penalty0 243--297, 2021.

\bibitem[Ayhan and Berens(2018)]{ayhan2018test}
Murat~Seckin Ayhan and Philipp Berens.
\newblock Test-time data augmentation for estimation of heteroscedastic aleatoric uncertainty in deep neural networks.
\newblock In \emph{Medical Imaging with Deep Learning}, 2018.

\bibitem[Barz and Denzler(2019)]{barz2019hierarchy}
Bj{\"o}rn Barz and Joachim Denzler.
\newblock Hierarchy-based image embeddings for semantic image retrieval.
\newblock In \emph{2019 IEEE winter conference on applications of computer vision (WACV)}, pages 638--647. IEEE, 2019.

\bibitem[Blundell et~al.(2015)Blundell, Cornebise, Kavukcuoglu, and Wierstra]{blundell2015weight}
Charles Blundell, Julien Cornebise, Koray Kavukcuoglu, and Daan Wierstra.
\newblock Weight uncertainty in neural network.
\newblock In \emph{International conference on machine learning}, pages 1613--1622. PMLR, 2015.

\bibitem[Chen et~al.(2015)Chen, Fang, Lin, Vedantam, Gupta, Doll{\'a}r, and Zitnick]{lin2014microsoft}
Xinlei Chen, Hao Fang, Tsung-Yi Lin, Ramakrishna Vedantam, Saurabh Gupta, Piotr Doll{\'a}r, and C~Lawrence Zitnick.
\newblock Microsoft coco captions: Data collection and evaluation server.
\newblock \emph{arXiv preprint arXiv:1504.00325}, 2015.

\bibitem[Chun(2023)]{chun2023improved}
Sanghyuk Chun.
\newblock Improved probabilistic image-text representations.
\newblock \emph{arXiv preprint arXiv:2305.18171}, 2023.

\bibitem[Chun et~al.(2021)Chun, Oh, De~Rezende, Kalantidis, and Larlus]{chun2021probabilistic}
Sanghyuk Chun, Seong~Joon Oh, Rafael~Sampaio De~Rezende, Yannis Kalantidis, and Diane Larlus.
\newblock Probabilistic embeddings for cross-modal retrieval.
\newblock In \emph{Proceedings of the IEEE/CVF Conference on Computer Vision and Pattern Recognition}, pages 8415--8424, 2021.

\bibitem[Corbiere et~al.(2021)Corbiere, Thome, Saporta, Vu, Cord, and Perez]{corbiere2021confidence}
Charles Corbiere, Nicolas Thome, Antoine Saporta, Tuan-Hung Vu, Matthieu Cord, and Patrick Perez.
\newblock Confidence estimation via auxiliary models.
\newblock \emph{IEEE Transactions on Pattern Analysis and Machine Intelligence}, 44\penalty0 (10):\penalty0 6043--6055, 2021.

\bibitem[Duchi(2007)]{duchi2007derivations}
John Duchi.
\newblock Derivations for linear algebra and optimization.
\newblock \emph{Berkeley, California}, 3\penalty0 (1):\penalty0 2325--5870, 2007.

\bibitem[Eleftheriadis et~al.(2014)Eleftheriadis, Rudovic, and Pantic]{eleftheriadis2014discriminative}
Stefanos Eleftheriadis, Ognjen Rudovic, and Maja Pantic.
\newblock Discriminative shared gaussian processes for multiview and view-invariant facial expression recognition.
\newblock \emph{IEEE transactions on image processing}, 24\penalty0 (1):\penalty0 189--204, 2014.

\bibitem[Frome et~al.(2013)Frome, Corrado, Shlens, Bengio, Dean, Ranzato, and Mikolov]{frome2013devise}
Andrea Frome, Greg~S Corrado, Jon Shlens, Samy Bengio, Jeff Dean, Marc'Aurelio Ranzato, and Tomas Mikolov.
\newblock Devise: A deep visual-semantic embedding model.
\newblock \emph{Advances in neural information processing systems}, 26, 2013.

\bibitem[Gal and Ghahramani(2016)]{gal2016dropout}
Yarin Gal and Zoubin Ghahramani.
\newblock Dropout as a bayesian approximation: Representing model uncertainty in deep learning.
\newblock In \emph{international conference on machine learning}, pages 1050--1059. PMLR, 2016.

\bibitem[Gardner et~al.(2018)Gardner, Pleiss, Weinberger, Bindel, and Wilson]{gardner2018gpytorch}
Jacob Gardner, Geoff Pleiss, Kilian~Q Weinberger, David Bindel, and Andrew~G Wilson.
\newblock Gpytorch: Blackbox matrix-matrix gaussian process inference with gpu acceleration.
\newblock \emph{Advances in neural information processing systems}, 31, 2018.

\bibitem[Garnelo et~al.(2018)Garnelo, Schwarz, Rosenbaum, Viola, Rezende, Eslami, and Teh]{garnelo2018neural}
Marta Garnelo, Jonathan Schwarz, Dan Rosenbaum, Fabio Viola, Danilo~J Rezende, SM~Eslami, and Yee~Whye Teh.
\newblock Neural processes.
\newblock \emph{arXiv preprint arXiv:1807.01622}, 2018.

\bibitem[Goyal et~al.(2017)Goyal, Khot, Summers-Stay, Batra, and Parikh]{goyal2017making}
Yash Goyal, Tejas Khot, Douglas Summers-Stay, Dhruv Batra, and Devi Parikh.
\newblock Making the v in vqa matter: Elevating the role of image understanding in visual question answering.
\newblock In \emph{Proceedings of the IEEE conference on computer vision and pattern recognition}, pages 6904--6913, 2017.

\bibitem[Guo et~al.(2017)Guo, Pleiss, Sun, and Weinberger]{guo2017calibration}
Chuan Guo, Geoff Pleiss, Yu~Sun, and Kilian~Q Weinberger.
\newblock On calibration of modern neural networks.
\newblock In \emph{International conference on machine learning}, pages 1321--1330. PMLR, 2017.

\bibitem[Gutmann and Hyv{\"a}rinen(2010)]{gutmann2010noise}
Michael Gutmann and Aapo Hyv{\"a}rinen.
\newblock Noise-contrastive estimation: A new estimation principle for unnormalized statistical models.
\newblock In \emph{Proceedings of the thirteenth international conference on artificial intelligence and statistics}, pages 297--304. JMLR Workshop and Conference Proceedings, 2010.

\bibitem[Hensman et~al.(2015)Hensman, Matthews, and Ghahramani]{hensman2015scalable}
James Hensman, Alexander Matthews, and Zoubin Ghahramani.
\newblock Scalable variational gaussian process classification.
\newblock In \emph{Artificial Intelligence and Statistics}, pages 351--360. PMLR, 2015.

\bibitem[Hornauer et~al.(2023)Hornauer, Holzbock, and Belagiannis]{hornauer2023out}
Julia Hornauer, Adrian Holzbock, and Vasileios Belagiannis.
\newblock Out-of-distribution detection for monocular depth estimation.
\newblock In \emph{Proceedings of the IEEE/CVF International Conference on Computer Vision}, pages 1911--1921, 2023.

\bibitem[Ji et~al.(2023)Ji, Wang, Gong, Zhang, Zhu, Wang, Zhang, Sakai, and Yang]{ji2023map}
Yatai Ji, Junjie Wang, Yuan Gong, Lin Zhang, Yanru Zhu, Hongfa Wang, Jiaxing Zhang, Tetsuya Sakai, and Yujiu Yang.
\newblock Map: Multimodal uncertainty-aware vision-language pre-training model.
\newblock In \emph{Proceedings of the IEEE/CVF Conference on Computer Vision and Pattern Recognition}, pages 23262--23271, 2023.

\bibitem[Jung et~al.(2022)Jung, Zhao, Dipnall, Gabbe, and Du]{jung2022uncertainty}
Myong~Chol Jung, He~Zhao, Joanna Dipnall, Belinda Gabbe, and Lan Du.
\newblock Uncertainty estimation for multi-view data: The power of seeing the whole picture.
\newblock \emph{Advances in Neural Information Processing Systems}, 35:\penalty0 6517--6530, 2022.

\bibitem[Kenton and Toutanova(2019)]{kenton2019bert}
Jacob Devlin Ming-Wei~Chang Kenton and Lee~Kristina Toutanova.
\newblock Bert: Pre-training of deep bidirectional transformers for language understanding.
\newblock In \emph{Proceedings of naacL-HLT}, volume~1, page~2, 2019.

\bibitem[Lalchand et~al.(2022)Lalchand, Ravuri, and Lawrence]{lalchand2022generalised}
Vidhi Lalchand, Aditya Ravuri, and Neil~D Lawrence.
\newblock Generalised gaussian process latent variable models (gplvm) with stochastic variational inference.
\newblock \emph{arXiv preprint arXiv:2202.12979}, 2022.

\bibitem[Lawrence(2003)]{lawrence2003gaussian}
Neil Lawrence.
\newblock Gaussian process latent variable models for visualisation of high dimensional data.
\newblock \emph{Advances in neural information processing systems}, 16, 2003.

\bibitem[Li et~al.(2017)Li, Lee, Huang, Wang, and Yang]{li2017learning}
Dong Li, Hsin-Ying Lee, Jia-Bin Huang, Shengjin Wang, and Ming-Hsuan Yang.
\newblock Learning structured semantic embeddings for visual recognition.
\newblock \emph{arXiv preprint arXiv:1706.01237}, 2017.

\bibitem[Li et~al.(2022{\natexlab{a}})Li, Song, Gao, Zeng, Zhang, and Li]{li2022differentiable}
Hao Li, Jingkuan Song, Lianli Gao, Pengpeng Zeng, Haonan Zhang, and Gongfu Li.
\newblock A differentiable semantic metric approximation in probabilistic embedding for cross-modal retrieval.
\newblock \emph{Advances in Neural Information Processing Systems}, 35:\penalty0 11934--11946, 2022{\natexlab{a}}.

\bibitem[Li et~al.(2024)Li, Wong, Jiang, Fang, Xie, and Xu]{li2024ckdh}
Jiaxing Li, Wai~Keung Wong, Lin Jiang, Xiaozhao Fang, Shengli Xie, and Yong Xu.
\newblock Ckdh: Clip-based knowledge distillation hashing for cross-modal retrieval.
\newblock \emph{IEEE Transactions on Circuits and Systems for Video Technology}, 2024.

\bibitem[Li et~al.(2022{\natexlab{b}})Li, Li, Xiong, and Hoi]{li2022blip}
Junnan Li, Dongxu Li, Caiming Xiong, and Steven Hoi.
\newblock Blip: Bootstrapping language-image pre-training for unified vision-language understanding and generation.
\newblock In \emph{International conference on machine learning}, pages 12888--12900. PMLR, 2022{\natexlab{b}}.

\bibitem[Li et~al.(2019)Li, Yatskar, Yin, Hsieh, and Chang]{li2019visualbert}
Liunian~Harold Li, Mark Yatskar, Da~Yin, Cho-Jui Hsieh, and Kai-Wei Chang.
\newblock Visualbert: A simple and performant baseline for vision and language.
\newblock \emph{arXiv preprint arXiv:1908.03557}, 2019.

\bibitem[Li et~al.(2023)Li, Wen, Hu, and Zhou]{li2023rs}
Xiang Li, Congcong Wen, Yuan Hu, and Nan Zhou.
\newblock Rs-clip: Zero shot remote sensing scene classification via contrastive vision-language supervision.
\newblock \emph{International Journal of Applied Earth Observation and Geoinformation}, 124:\penalty0 103497, 2023.

\bibitem[Lin and Gong(2023)]{lin2023gridclip}
Jiayi Lin and Shaogang Gong.
\newblock Gridclip: One-stage object detection by grid-level clip representation learning.
\newblock \emph{arXiv preprint arXiv:2303.09252}, 2023.

\bibitem[Liu et~al.(2020)Liu, Lin, Padhy, Tran, Bedrax~Weiss, and Lakshminarayanan]{liu2020simple}
Jeremiah Liu, Zi~Lin, Shreyas Padhy, Dustin Tran, Tania Bedrax~Weiss, and Balaji Lakshminarayanan.
\newblock Simple and principled uncertainty estimation with deterministic deep learning via distance awareness.
\newblock \emph{Advances in neural information processing systems}, 33:\penalty0 7498--7512, 2020.

\bibitem[Lu et~al.(2019)Lu, Batra, Parikh, and Lee]{lu2019vilbert}
Jiasen Lu, Dhruv Batra, Devi Parikh, and Stefan Lee.
\newblock Vilbert: Pretraining task-agnostic visiolinguistic representations for vision-and-language tasks.
\newblock \emph{Advances in neural information processing systems}, 32, 2019.

\bibitem[Mukhoti et~al.(2023)Mukhoti, Kirsch, van Amersfoort, Torr, and Gal]{mukhoti2023deep}
Jishnu Mukhoti, Andreas Kirsch, Joost van Amersfoort, Philip~HS Torr, and Yarin Gal.
\newblock Deep deterministic uncertainty: A new simple baseline.
\newblock In \emph{Proceedings of the IEEE/CVF Conference on Computer Vision and Pattern Recognition}, pages 24384--24394, 2023.

\bibitem[Nilsback and Zisserman(2008)]{nilsback2008automated}
Maria-Elena Nilsback and Andrew Zisserman.
\newblock Automated flower classification over a large number of classes.
\newblock In \emph{2008 Sixth Indian conference on computer vision, graphics \& image processing}, pages 722--729. IEEE, 2008.

\bibitem[Oord et~al.(2018)Oord, Li, and Vinyals]{oord2018representation}
Aaron van~den Oord, Yazhe Li, and Oriol Vinyals.
\newblock Representation learning with contrastive predictive coding.
\newblock \emph{arXiv preprint arXiv:1807.03748}, 2018.

\bibitem[Parelli et~al.(2023)Parelli, Delitzas, Hars, Vlassis, Anagnostidis, Bachmann, and Hofmann]{parelli2023clip}
Maria Parelli, Alexandros Delitzas, Nikolas Hars, Georgios Vlassis, Sotirios Anagnostidis, Gregor Bachmann, and Thomas Hofmann.
\newblock Clip-guided vision-language pre-training for question answering in 3d scenes.
\newblock In \emph{Proceedings of the IEEE/CVF Conference on Computer Vision and Pattern Recognition}, pages 5607--5612, 2023.

\bibitem[Platt et~al.(1999)]{platt1999probabilistic}
John Platt et~al.
\newblock Probabilistic outputs for support vector machines and comparisons to regularized likelihood methods.
\newblock \emph{Advances in large margin classifiers}, 10\penalty0 (3):\penalty0 61--74, 1999.

\bibitem[Qian and Hu(2024)]{qian2024online}
Qi~Qian and Juhua Hu.
\newblock Online zero-shot classification with clip.
\newblock \emph{arXiv preprint arXiv:2408.13320}, 2024.

\bibitem[Radford et~al.(2021)Radford, Kim, Hallacy, Ramesh, Goh, Agarwal, Sastry, Askell, Mishkin, Clark, et~al.]{radford2021learning}
Alec Radford, Jong~Wook Kim, Chris Hallacy, Aditya Ramesh, Gabriel Goh, Sandhini Agarwal, Girish Sastry, Amanda Askell, Pamela Mishkin, Jack Clark, et~al.
\newblock Learning transferable visual models from natural language supervision.
\newblock In \emph{International conference on machine learning}, pages 8748--8763. PMLR, 2021.

\bibitem[Reed et~al.(2016)Reed, Akata, Lee, and Schiele]{reed2016learning}
Scott Reed, Zeynep Akata, Honglak Lee, and Bernt Schiele.
\newblock Learning deep representations of fine-grained visual descriptions.
\newblock In \emph{Proceedings of the IEEE conference on computer vision and pattern recognition}, pages 49--58, 2016.

\bibitem[Sensoy et~al.(2018)Sensoy, Kaplan, and Kandemir]{sensoy2018evidential}
Murat Sensoy, Lance Kaplan, and Melih Kandemir.
\newblock Evidential deep learning to quantify classification uncertainty.
\newblock \emph{Advances in neural information processing systems}, 31, 2018.

\bibitem[Shi and Jain(2019)]{shi2019probabilistic}
Yichun Shi and Anil~K Jain.
\newblock Probabilistic face embeddings.
\newblock In \emph{Proceedings of the IEEE/CVF International Conference on Computer Vision}, pages 6902--6911, 2019.

\bibitem[Singh et~al.(2022)Singh, Hu, Goswami, Couairon, Galuba, Rohrbach, and Kiela]{singh2022flava}
Amanpreet Singh, Ronghang Hu, Vedanuj Goswami, Guillaume Couairon, Wojciech Galuba, Marcus Rohrbach, and Douwe Kiela.
\newblock Flava: A foundational language and vision alignment model.
\newblock In \emph{Proceedings of the IEEE/CVF Conference on Computer Vision and Pattern Recognition}, pages 15638--15650, 2022.

\bibitem[Song et~al.(2017)Song, Wang, Huang, and Tian]{song2017multimodal}
Guoli Song, Shuhui Wang, Qingming Huang, and Qi~Tian.
\newblock Multimodal gaussian process latent variable models with harmonization.
\newblock In \emph{Proceedings of the IEEE international conference on computer vision}, pages 5029--5037, 2017.

\bibitem[Tan and Bansal(2019)]{tan2019lxmert}
Hao Tan and Mohit Bansal.
\newblock Lxmert: Learning cross-modality encoder representations from transformers.
\newblock \emph{arXiv preprint arXiv:1908.07490}, 2019.

\bibitem[Titsias(2009)]{titsias2009variational}
Michalis Titsias.
\newblock Variational learning of inducing variables in sparse gaussian processes.
\newblock In \emph{Artificial intelligence and statistics}, pages 567--574. PMLR, 2009.

\bibitem[Upadhyay et~al.(2023)Upadhyay, Karthik, Mancini, and Akata]{upadhyay2023probvlm}
Uddeshya Upadhyay, Shyamgopal Karthik, Massimiliano Mancini, and Zeynep Akata.
\newblock Probvlm: Probabilistic adapter for frozen vison-language models.
\newblock In \emph{Proceedings of the IEEE/CVF International Conference on Computer Vision}, pages 1899--1910, 2023.

\bibitem[Van~Amersfoort et~al.(2020)Van~Amersfoort, Smith, Teh, and Gal]{van2020uncertainty}
Joost Van~Amersfoort, Lewis Smith, Yee~Whye Teh, and Yarin Gal.
\newblock Uncertainty estimation using a single deep deterministic neural network.
\newblock In \emph{International conference on machine learning}, pages 9690--9700. PMLR, 2020.

\bibitem[Vaswani et~al.(2017)Vaswani, Shazeer, Parmar, Uszkoreit, Jones, Gomez, Kaiser, and Polosukhin]{vaswani2017attention}
Ashish Vaswani, Noam Shazeer, Niki Parmar, Jakob Uszkoreit, Llion Jones, Aidan~N Gomez, Lukasz Kaiser, and Illia Polosukhin.
\newblock Attention is all you need.
\newblock \emph{Advances in Neural Information Processing Systems}, 30, 2017.

\bibitem[Venkataramanan et~al.(2021)Venkataramanan, Laviale, Figus, Usseglio-Polatera, and Pradalier]{venkataramanan2021tackling}
Aishwarya Venkataramanan, Martin Laviale, C{\'e}cile Figus, Philippe Usseglio-Polatera, and C{\'e}dric Pradalier.
\newblock Tackling inter-class similarity and intra-class variance for microscopic image-based classification.
\newblock In \emph{International conference on computer vision systems}, pages 93--103. Springer, 2021.

\bibitem[Venkataramanan et~al.(2023{\natexlab{a}})Venkataramanan, Benbihi, Laviale, and Pradalier]{venkataramanan2023gaussian}
Aishwarya Venkataramanan, Assia Benbihi, Martin Laviale, and C{\'e}dric Pradalier.
\newblock Gaussian latent representations for uncertainty estimation using mahalanobis distance in deep classifiers.
\newblock In \emph{Proceedings of the IEEE/CVF International Conference on Computer Vision}, pages 4488--4497, 2023{\natexlab{a}}.

\bibitem[Venkataramanan et~al.(2023{\natexlab{b}})Venkataramanan, Laviale, and Pradalier]{venkataramanan2023integrating}
Aishwarya Venkataramanan, Martin Laviale, and C{\'e}dric Pradalier.
\newblock Integrating visual and semantic similarity using hierarchies for image retrieval.
\newblock In \emph{International Conference on Computer Vision Systems}, pages 422--431. Springer, 2023{\natexlab{b}}.

\bibitem[Verma et~al.(2021)Verma, Mishra, Pandey, Murthy, and Rai]{verma2021towards}
Vinay~Kumar Verma, Ashish Mishra, Anubha Pandey, Hema~A Murthy, and Piyush Rai.
\newblock Towards zero-shot learning with fewer seen class examples.
\newblock In \emph{Proceedings of the IEEE/CVF Winter Conference on Applications of Computer Vision}, pages 2241--2251, 2021.

\bibitem[Wah et~al.(2011)Wah, Branson, Welinder, Perona, and Belongie]{wah2011caltech}
Catherine Wah, Steve Branson, Peter Welinder, Pietro Perona, and Serge Belongie.
\newblock The caltech-ucsd birds-200-2011 dataset.
\newblock 2011.

\bibitem[Wang et~al.(2019)Wang, Li, Aertsen, Deprest, Ourselin, and Vercauteren]{wang2019aleatoric}
Guotai Wang, Wenqi Li, Michael Aertsen, Jan Deprest, S{\'e}bastien Ourselin, and Tom Vercauteren.
\newblock Aleatoric uncertainty estimation with test-time augmentation for medical image segmentation with convolutional neural networks.
\newblock \emph{Neurocomputing}, 338:\penalty0 34--45, 2019.

\bibitem[Williams and Rasmussen(2006)]{williams2006gaussian}
Christopher~KI Williams and Carl~Edward Rasmussen.
\newblock \emph{Gaussian processes for machine learning}, volume~2.
\newblock MIT press Cambridge, MA, 2006.

\bibitem[Wu et~al.(2023)Wu, Zhu, Zhao, and Li]{wu2023cora}
Xiaoshi Wu, Feng Zhu, Rui Zhao, and Hongsheng Li.
\newblock Cora: Adapting clip for open-vocabulary detection with region prompting and anchor pre-matching.
\newblock In \emph{Proceedings of the IEEE/CVF conference on computer vision and pattern recognition}, pages 7031--7040, 2023.

\bibitem[Xia et~al.(2023)Xia, Dong, Li, Zhu, and Ying]{xia2023clip}
Xinyu Xia, Guohua Dong, Fengling Li, Lei Zhu, and Xiaomin Ying.
\newblock When clip meets cross-modal hashing retrieval: A new strong baseline.
\newblock \emph{Information Fusion}, 100:\penalty0 101968, 2023.

\bibitem[Xing et~al.(2024)Xing, Li, Wang, Zhu, and Cao]{xing2024clipvqa}
Fengchuang Xing, Mingjie Li, Yuan-Gen Wang, Guopu Zhu, and Xiaochun Cao.
\newblock Clipvqa: Video quality assessment via clip.
\newblock \emph{arXiv preprint arXiv:2407.04928}, 2024.

\bibitem[Young et~al.(2014)Young, Lai, Hodosh, and Hockenmaier]{young2014image}
Peter Young, Alice Lai, Micah Hodosh, and Julia Hockenmaier.
\newblock From image descriptions to visual denotations: New similarity metrics for semantic inference over event descriptions.
\newblock \emph{Transactions of the Association for Computational Linguistics}, 2:\penalty0 67--78, 2014.

\bibitem[Yu et~al.(2021)Yu, Franchi, and Aldea]{yu2021slurp}
Xuanlong Yu, Gianni Franchi, and Emanuel Aldea.
\newblock Slurp: Side learning uncertainty for regression problems.
\newblock \emph{arXiv preprint arXiv:2110.11182}, 2021.

\bibitem[Zhang et~al.(2024)Zhang, Huang, Jin, and Lu]{zhang2024vision}
Jingyi Zhang, Jiaxing Huang, Sheng Jin, and Shijian Lu.
\newblock Vision-language models for vision tasks: A survey.
\newblock \emph{IEEE Transactions on Pattern Analysis and Machine Intelligence}, 2024.

\bibitem[Zhang and Saligrama(2015)]{zhang2015zero}
Ziming Zhang and Venkatesh Saligrama.
\newblock Zero-shot learning via semantic similarity embedding.
\newblock In \emph{Proceedings of the IEEE international conference on computer vision}, pages 4166--4174, 2015.

\end{thebibliography}

\newpage

\onecolumn

\title{Probabilistic Embeddings for Frozen Vision-Language Models: Uncertainty
Quantification with Gaussian Process Latent Variable Models\\(Supplementary Material)}
\maketitle

\appendix

\section{Additional Implementation Details} \label{sec:app_implementation}

This section provides details on the data processing steps for training both the baseline models and GroVE, implementation details for each baseline, and the hyper-parameter tuning procedure applied for GroVE.

\subsection{Datasets} \label{sec:app_dataset}
For the experiments, we use MS-COCO, Flickr30k, CUB-200-2011, and Oxford Flowers 102 dataset. 

\textbf{MS-COCO~\cite{lin2014microsoft}} is a widely used cross-modal retrieval dataset includes 123,287 images, each image annotated with 5 captions describing common objects. The training set comprises 113,287 images, while both the validation and test sets contain 5,000 images each. Different papers apply varying evaluation protocols on the 5,000 test images in the COCO dataset. Some cross-modal retrieval papers report results on the full 5,000 test set, while others use 1,000 unique test images, averaging results over 5 random splits. In our study, we follow the former approach, presenting results based on the entire 5,000 test set.

\textbf{Flickr30k~\cite{young2014image}} is a widely used cross-modal retrieval dataset comprising 31,783 images, each image annotated with 5 captions describing common objects. The dataset is split into 29,783 training images, with 1,000 images each in the validation and test sets.

\textbf{CUB-200-2011~\cite{wah2011caltech}} is a fine-grained bird species dataset comprising 11,788 images across 200 categories, with each image paired with 10 captions sourced from \cite{reed2016learning}. Following the split protocol in \cite{chun2021probabilistic}, the dataset includes 7,067 training images, 1,754 validation images, and 2,967 test images.

\textbf{Oxford Flowers 102~\cite{nilsback2008automated}} is a fine-grained flowers dataset comprising 8,189 images across 102 categories, with each image paired with 10 captions sourced from \cite{reed2016learning}. Following the split protocol in \cite{upadhyay2023probvlm}, the dataset includes 7,034 training images, 750 validation and 805 test images.

All images are resized to $224 \times 224$, suitable for input to VLMs. All methods are trained and evaluated using the same dataset splits for the comparison.

\subsection{Baseline methods} \label{sec:baselines}
In this section, we provide the implementation and training details for the baseline methods.

\textbf{TTDA~\cite{ayhan2018test}.} During inference, data augmentations are applied to the input, generating multiple variations to estimate prediction uncertainty. For each augmented version, a prediction is generated, and the variance across these predictions reflects the model's uncertainty. Image augmentations include random resized cropping and horizontal flipping (applied with a probability of 0.3), while text data undergoes random word masking with a 0.3 probability. The model is run for 10 passes on these augmentations, to obtain the image and text uncertainty.

\textbf{PFE~\cite{shi2019probabilistic}, PCME~\cite{chun2021probabilistic} and PCME++~\cite{chun2023improved}.} During training, we adapt these methods to process image and text embeddings derived from a frozen VLM. Following \cite{upadhyay2023probvlm}, the model architecture consists of two Multi-Layer Perceptrons (MLPs)—one for images and one for text. Each MLP has an input layer that reduces the embedding dimension to 256, a hidden layer of 256 units, and an output layer that maps from 256 back to the original embedding dimension.  We apply the respective loss functions to learn covariances for a Gaussian distribution, with mean values matching the VLM’s deterministic embeddings. Training is conducted for 200 epochs using the Adam optimizer with a learning rate of $10^{-8}$ and batch size of 64. The learning rate was fixed using a grid search over values $\{1e^{-4}, 1e^{-5}, 1e^{-6}, 1e^{-7}, 1e^{-8}\}$

\textbf{ProbVLM~\cite{upadhyay2023probvlm}.} We follow the training procedure outlined in the original paper. The model architecture consists of two MLPs—one for image embeddings and one for text embeddings—similar to previous methods. Training is conducted with the Adam optimizer for 100 epochs, using a learning rate of $10^{-4}$ and a batch size of 64.

\subsection{Hyper-parameter Tuning} \label{sec:app_hyper}

GroVE introduces the following hyper-parameters which were obtained using grid-search: latent space dimension $Q$, and the trade-off parameters $\lambda_1$ and $\lambda_2$. For $Q$, we evaluated values $Q \in \{2, 5, 10, 20, 50, 128, 256\}$. For the trade-off parameters, we used $\lambda_1 \in \{1, 0.1, 0.01, 0.001\}$, and $\lambda_2 \in \{0, 200, 400, 600, 800, 1000\}$.
based on the grid-search results, the optimal setting for $Q$ was $Q=5$ for MS-COCO and Flickr30k, and $Q=10$ for CUB-200-2011 and Oxford Flowers 102. The trade-off parameters that achieved the best performance were $\lambda_1 = 0.01$ and $\lambda_2 = 400$. The number of inducing points was selected from \{100, 150, 200, 250, 300, 350\}, with model performance plateauing beyond 250 points.
Finally, the learning rate of $1e^{-5}$ was selected based on a grid-search over values $\{ 1e^{-1}, 1e^{-2}, 1e^{-3}, 1e^{-4}, 1e^{-5}, 1e^{-6}\}$.

\section{Definitions} \label{sec:definitions}
This section presents the definitions of the GP kernels and the probabilistic distance metrics used in the ablation study.

\subsection{GP Kernels} \label{sec:def_kernel}
In Table.~\ref{tab:ablation_kernels}, we presented the results for ablation study of various kernels for GP. In this section, we provide the definitions and formulas for the kernels used.

\textbf{Radial Basis Function (RBF).} The RBF kernel, also known as the Gaussian kernel, is a popular choice in GPs. It assumes that closer data points in input space have higher similarity. The RBF kernel between two points $\mathbf{x_i}$ and $\mathbf{x_j}$ is defined as:
\begin{equation}
    k(\mathbf{x_i}, \mathbf{x_j}) = \exp \left(- \frac{\| \mathbf{x_i} - \mathbf{x_j}\|}{l^2} \right)
\end{equation}
where $l$ is the length scale parameter, controlling how quickly the similarity decreases with distance in input space. 

\textbf{Mat\'ern.} The Matérn kernel generalizes the RBF kernel, defined as
\begin{equation}
    k(\mathbf{x_i}, \mathbf{x_j}) =   \frac{1}{\Gamma(\nu)2^{\nu-1}}\Bigg(
\frac{\sqrt{2\nu}}{l} \|\mathbf{x_i} - \mathbf{x_j}\|
\Bigg)^\nu K_\nu\Bigg(
\frac{\sqrt{2\nu}}{l} \|\mathbf{x_i} - \mathbf{x_j}\|\Bigg)
\end{equation}
where $\nu$ controls the smoothness of the resulting function, $l$ is the length scale parameter, $\Gamma$ is the Gamma function and $K_\nu$ is a modified Bessel function. 

\textbf{Cosine Similarity.} This is a linear kernel with normalized inputs, and is defined as:
\begin{equation}
    k(\mathbf{x_i}, \mathbf{x_j}) = \frac{\mathbf{x_i}^T\mathbf{x_j}}{\|\mathbf{x_i}\|\|\mathbf{x_j}\|}
\end{equation}

\subsection{Probabilistic Distances} \label{sec:def_dist}
Sec.~\ref{sec:ablation} presented an ablation study on the choice of the probability distance metric for cross-modal alignment (refer Table.~\ref{tab:ablation_kl}). The definitions of the probabilistic distance metrics for two multivariate Gaussians $p = \mathcal{N}(\hat{\boldsymbol{\mu}}_\mathcal{I}, \hat{\boldsymbol{\Sigma}}_\mathcal{I})$ and $q = \mathcal{N}(\hat{\boldsymbol{\mu}}_\mathcal{T}, \hat{\boldsymbol{\Sigma}}_\mathcal{T})$, are as follows:

\textbf{Kullback-Liebler Divergence.} The KL Divergence quantifies the difference between two probability distributions, and is defined as~\cite{duchi2007derivations}:
\begin{equation} 
    D_{KL} (p\|q) =
     \frac{1}{2} \Bigg[ \text{tr}(\hat{\boldsymbol{\Sigma}}_\mathcal{T}^{-1} \hat{\boldsymbol{\Sigma}}_\mathcal{I}) + (\hat{\boldsymbol{\mu}}_\mathcal{T} - \hat{\boldsymbol{\mu}}_\mathcal{I})^{T} \hat{\boldsymbol{\Sigma}}_\mathcal{T}^{-1} (\hat{\boldsymbol{\mu}}_\mathcal{T} - \hat{\boldsymbol{\mu}}_\mathcal{I}) \\
    - D + \log \left( \frac{\det(\hat{\boldsymbol{\Sigma}}_\mathcal{T})}{\det(\hat{\boldsymbol{\Sigma}}_\mathcal{I})} \right) \Bigg],   
\end{equation}
where $\text{tr}(.)$ is the trace and $\det(.)$ is the determinant of a matrix. 
Note that the KL-Divergence is asymmetric; thus, we calculate the total cross-alignment loss as the mean of the KL divergences in both directions (refer Eq.~\ref{eq:kl}):
$\frac{1}{2}[D_{KL}(p \| q) + D_{KL}(q \| p)]$.

\textbf{Jensen-Shannon (JS) Divergence.} The JS Divergence is obtained by averaging the KL divergences between each distribution and the average distribution. The JS divergence is defined as:
\begin{equation}
    D_{JS}(p\|q) = \frac{1}{2} \left(D_{KL}(p \| m) +  D_{KL}(q \| m)\right)
\end{equation}
where $m = \frac{1}{2}(p+q)$ is the mean distribution of $p$ and $q$. 

\textbf{Wasserstein-2 distance.} The Wasserstein-2 distance quantifies the cost of transforming one distribution into another. This is defined as:
\begin{equation}
    W_2^2(p,q) = \|\hat{\boldsymbol{\mu}}_\mathcal{I} - \hat{\boldsymbol{\mu}}_\mathcal{T}\|^2 + \\\text{tr}\left(\hat{\boldsymbol{\Sigma}}_\mathcal{I} + \hat{\boldsymbol{\Sigma}}_\mathcal{T} - 2\left(\hat{\boldsymbol{\Sigma}}_\mathcal{I}^{1/2} \, \hat{\boldsymbol{\Sigma}}_\mathcal{T} \, \hat{\boldsymbol{\Sigma}}_\mathcal{I}^{1/2}\right)^{1/2}\right) 
\end{equation}

\section{Additional Results} \label{sec:add_results}
This section presents additional quantitative and qualitative results.

\subsection{Cross-modal Retrieval} \label{sec:app_cross-modal}

\subsubsection{Calibration plots}
\begin{figure*}[t]
    \centering
    \begin{tabular}{@{}c@{}c@{}c@{}c@{}c@{}}
        & Flickr30k & MS-COCO & CUB & Flowers \\
        \vspace{-5pt}
        \raisebox{1.2cm}{\rotatebox{90}{\centering Image to Text}} &
        \includegraphics[width=0.23\linewidth, height=4cm]{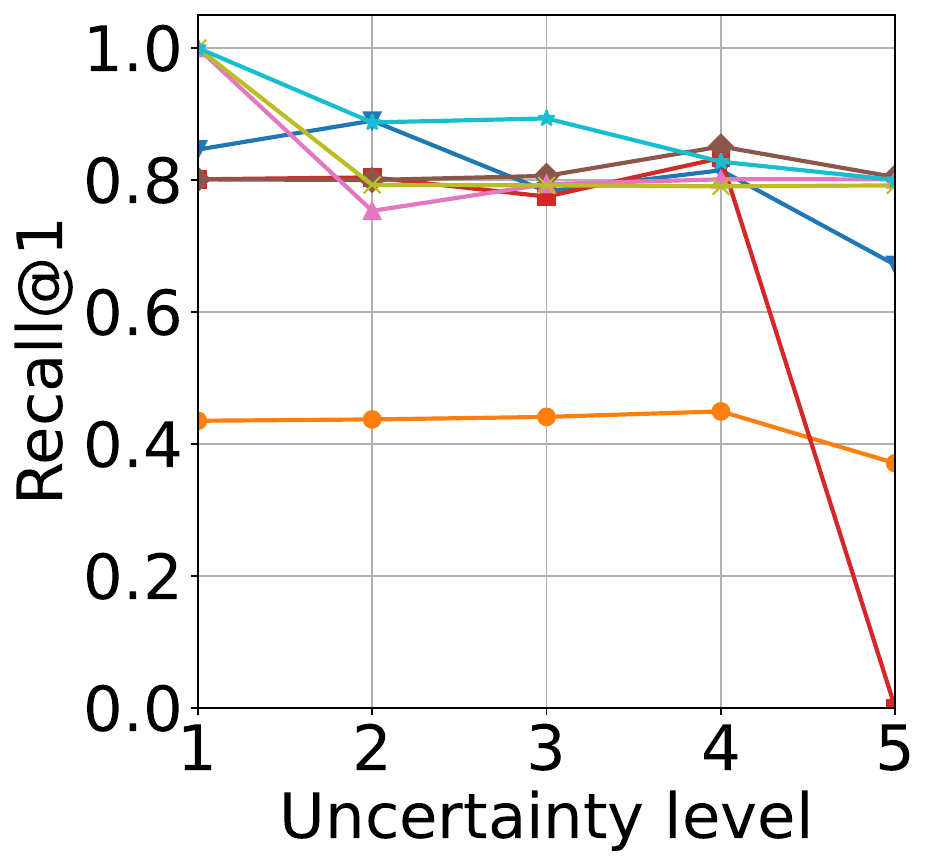} & 
        \includegraphics[width=0.23\linewidth, height=4cm]{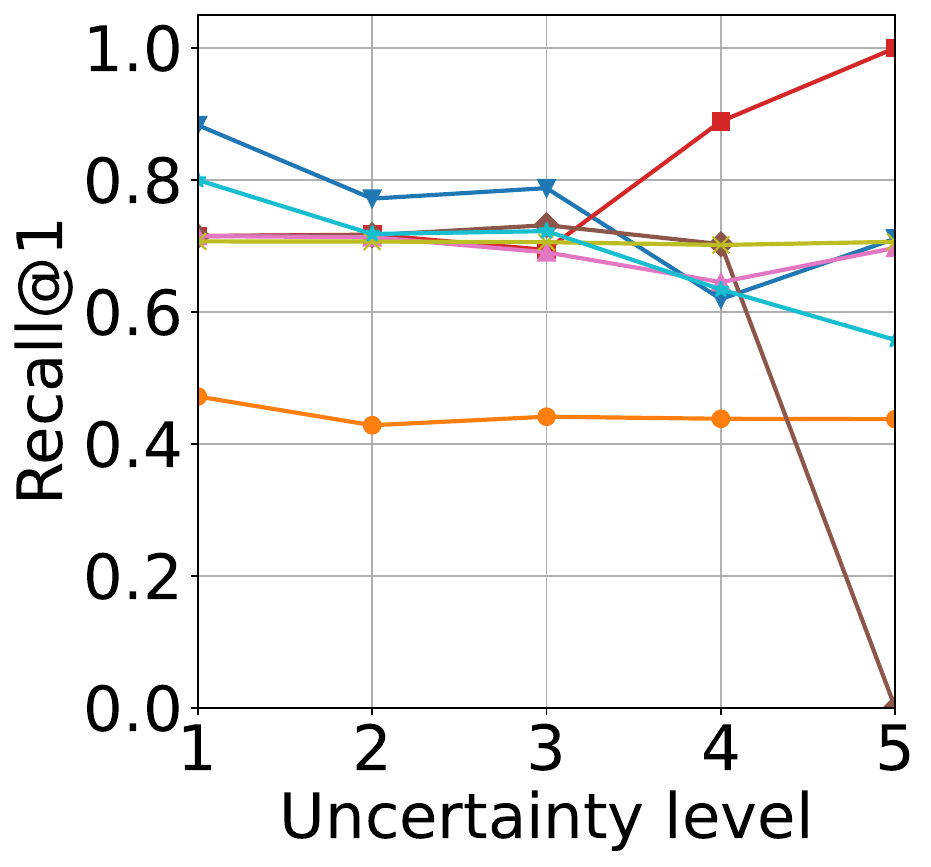} & 
        \includegraphics[width=0.23\linewidth, height=4cm]{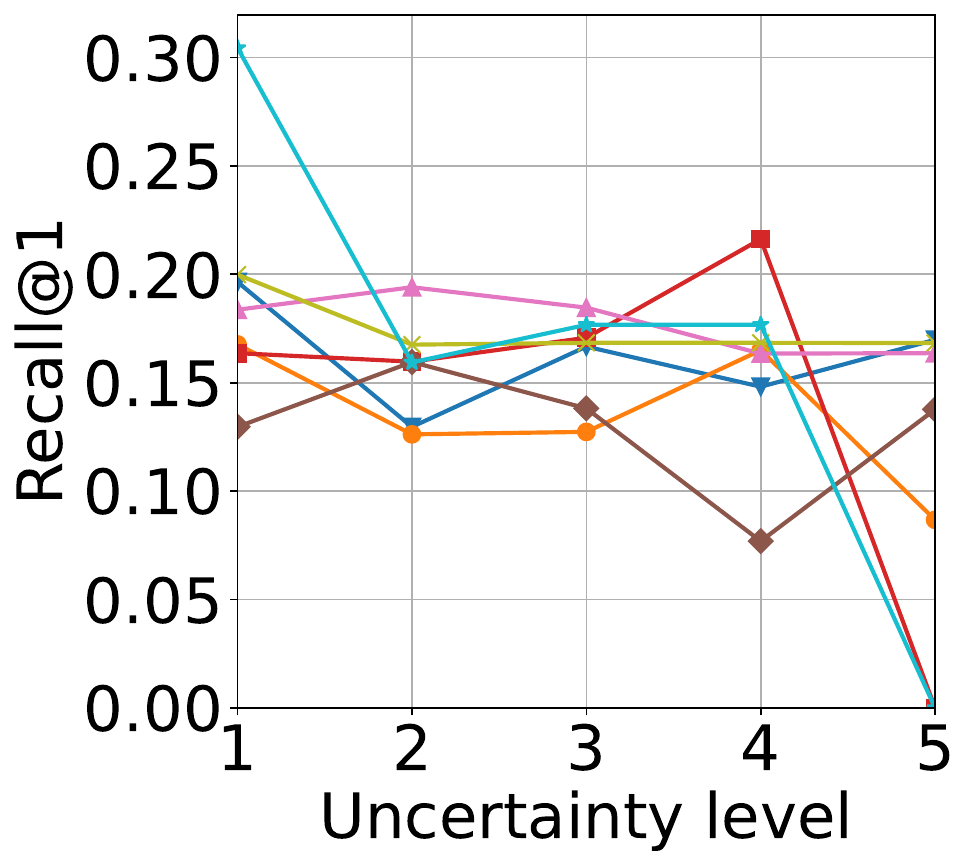} &
        \includegraphics[width=0.23\linewidth, height=4cm]{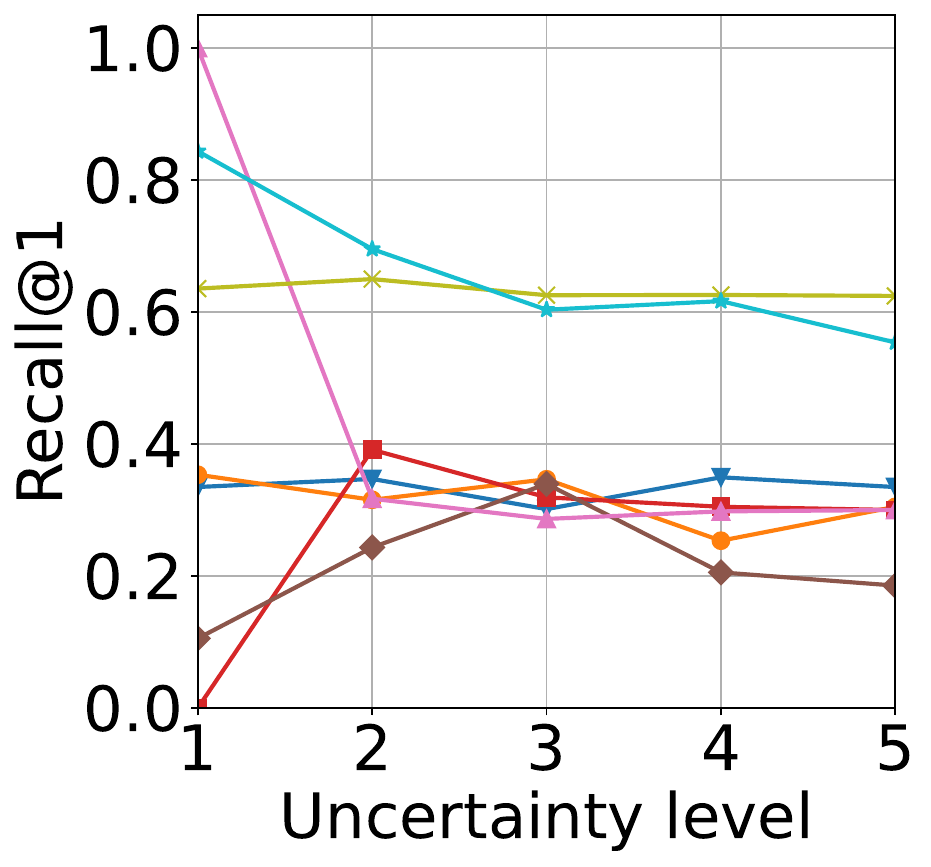}    \\    
        \raisebox{1.2cm}{\rotatebox{90}{\centering Text to Image}} &
        \includegraphics[width=0.23\linewidth, height=4cm]{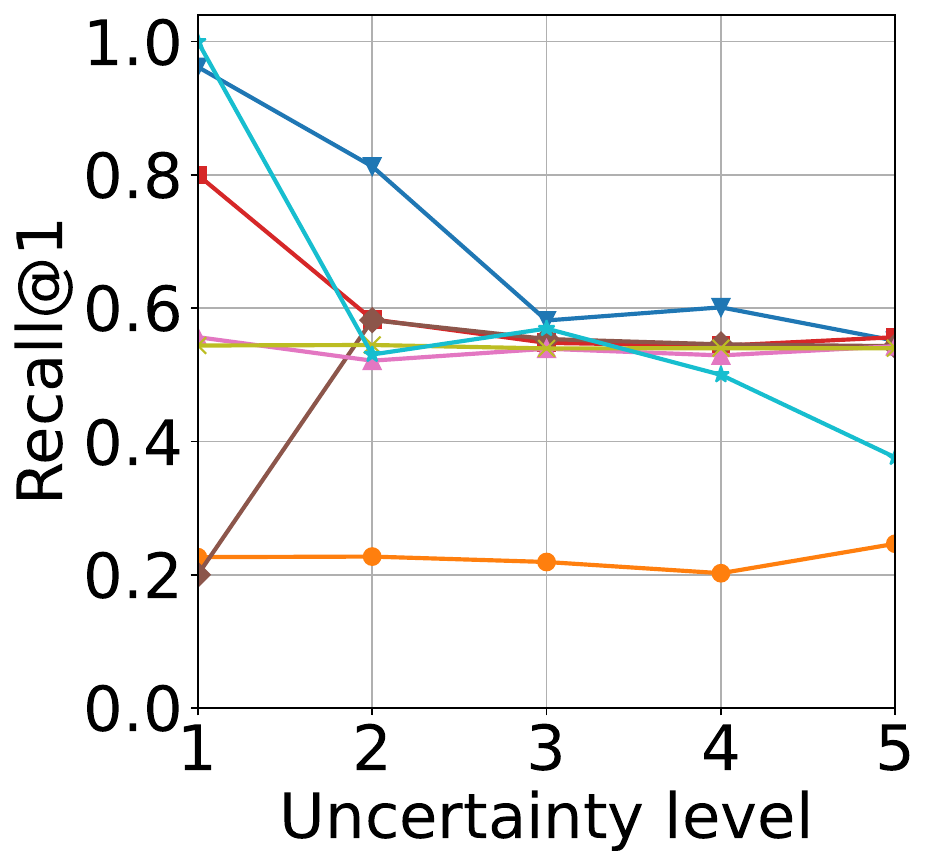} & 
        \includegraphics[width=0.23\linewidth, height=4cm]{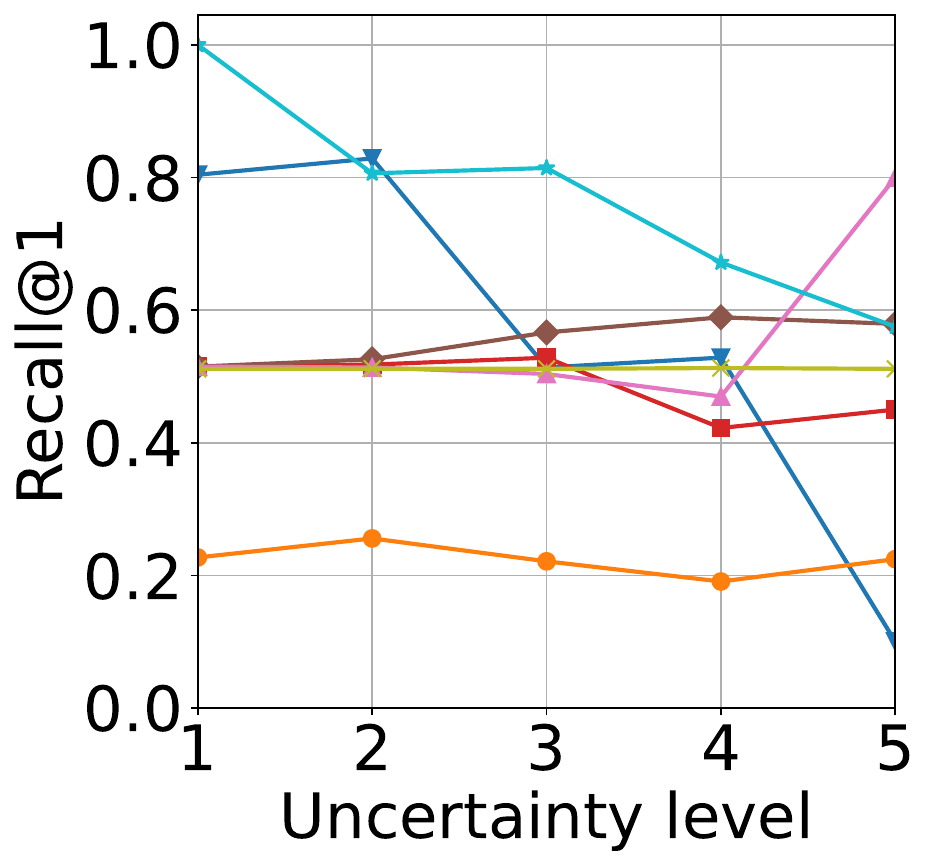} & 
        \includegraphics[width=0.23\linewidth, height=4cm]{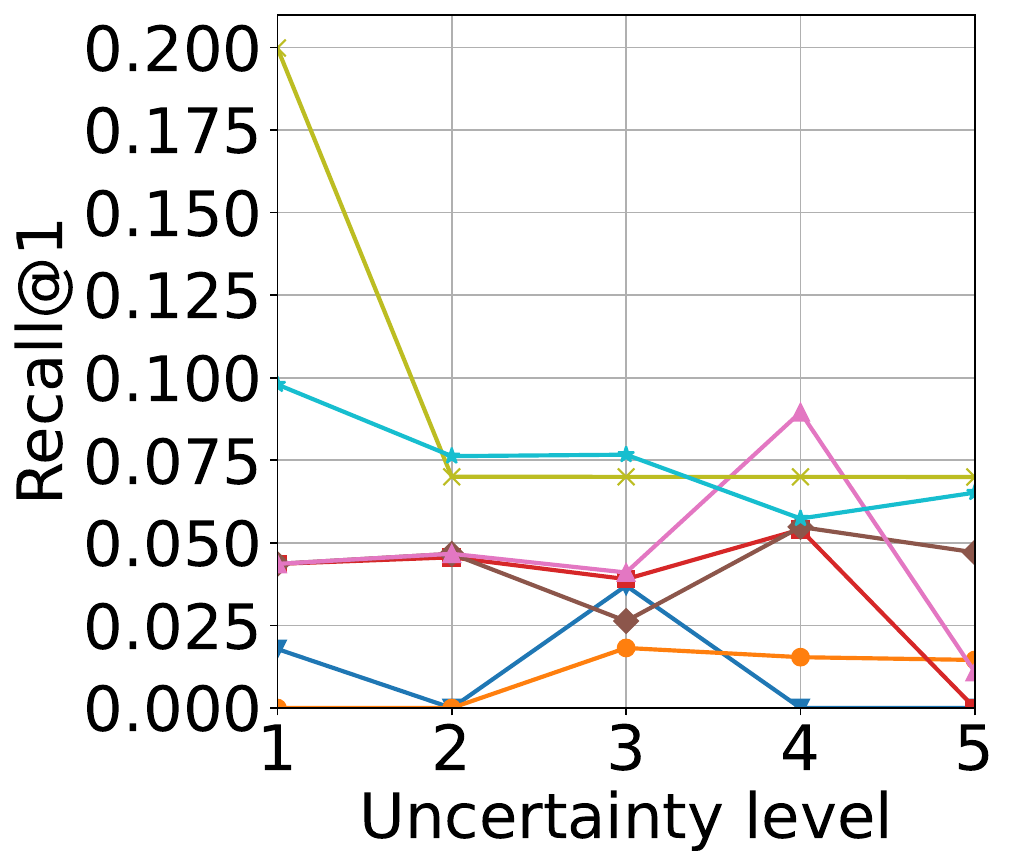} &
        \includegraphics[width=0.23\linewidth, height=4cm]{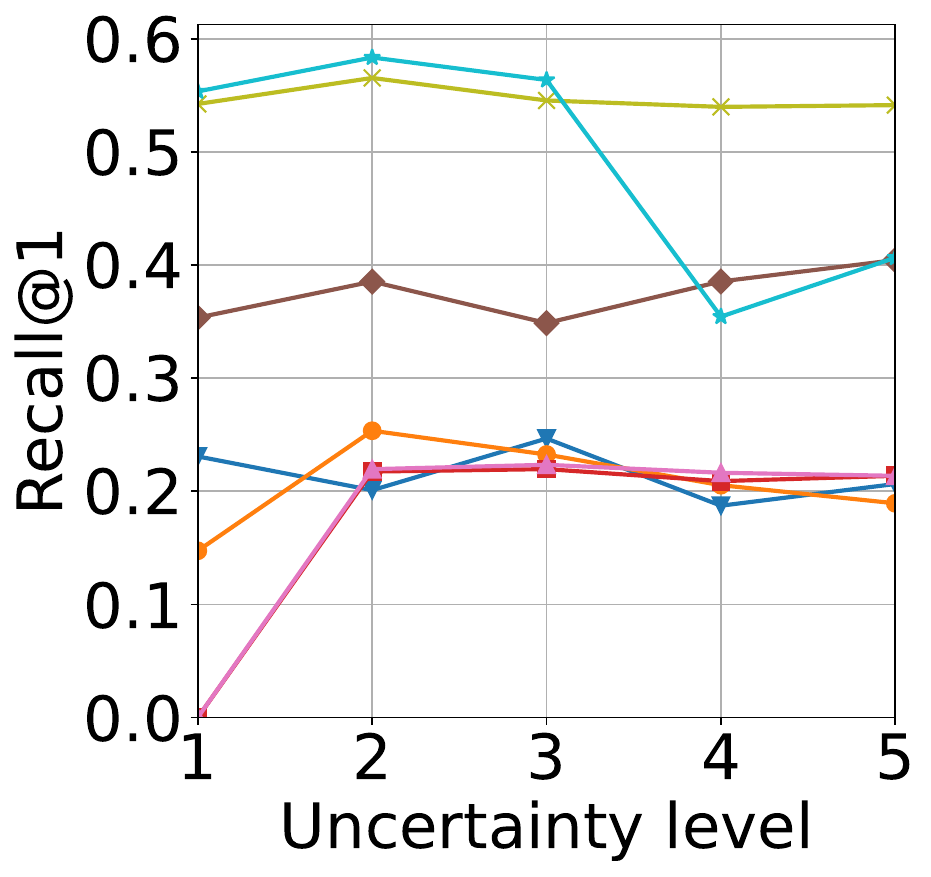} 
    \end{tabular}
    \includegraphics[width=0.8\linewidth]{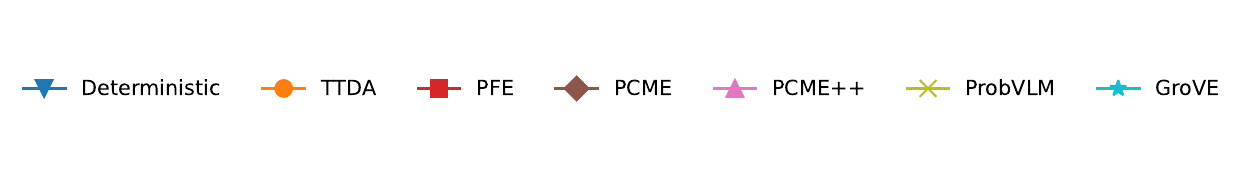} 
    \vspace{-25pt}
    \caption{\textbf{Evaluation of uncertainty calibration} for embeddings obtained from CLIP on Image-to-Text retrieval. For perfect calibration, the Recall@1 should show a monotonic decrease as uncertainty levels increase.
    GroVE exhibits a more consistent relationship between increasing uncertainty and performance degradation compared to the baseline methods.
    }
    \label{fig:clip_calibration}
\end{figure*}

\begin{figure*}[h]
    \centering
    \begin{tabular}{@{}c@{}c@{}c@{}c@{}c@{}}
        & Flickr & MS-COCO & CUB & Flowers \\
        \raisebox{1.2cm}{\rotatebox{90}{\small Image to Text}} &
        \includegraphics[width=0.23\linewidth, height=4cm]{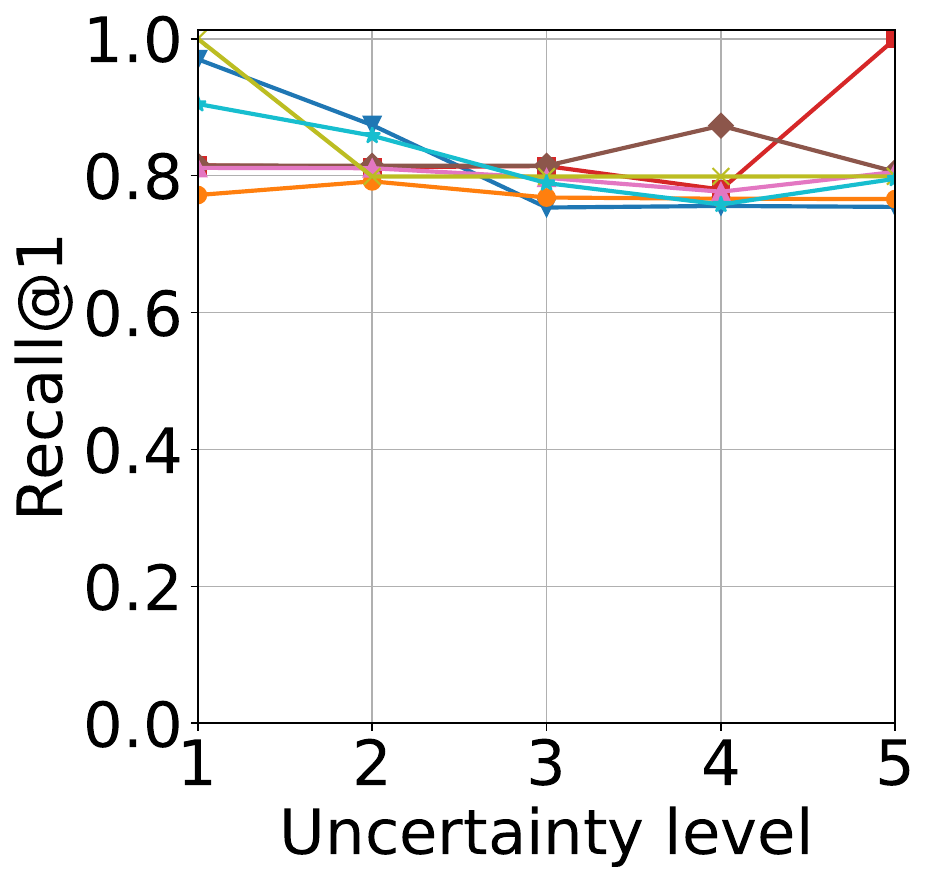} & 
        \includegraphics[width=0.23\linewidth, height=4cm]{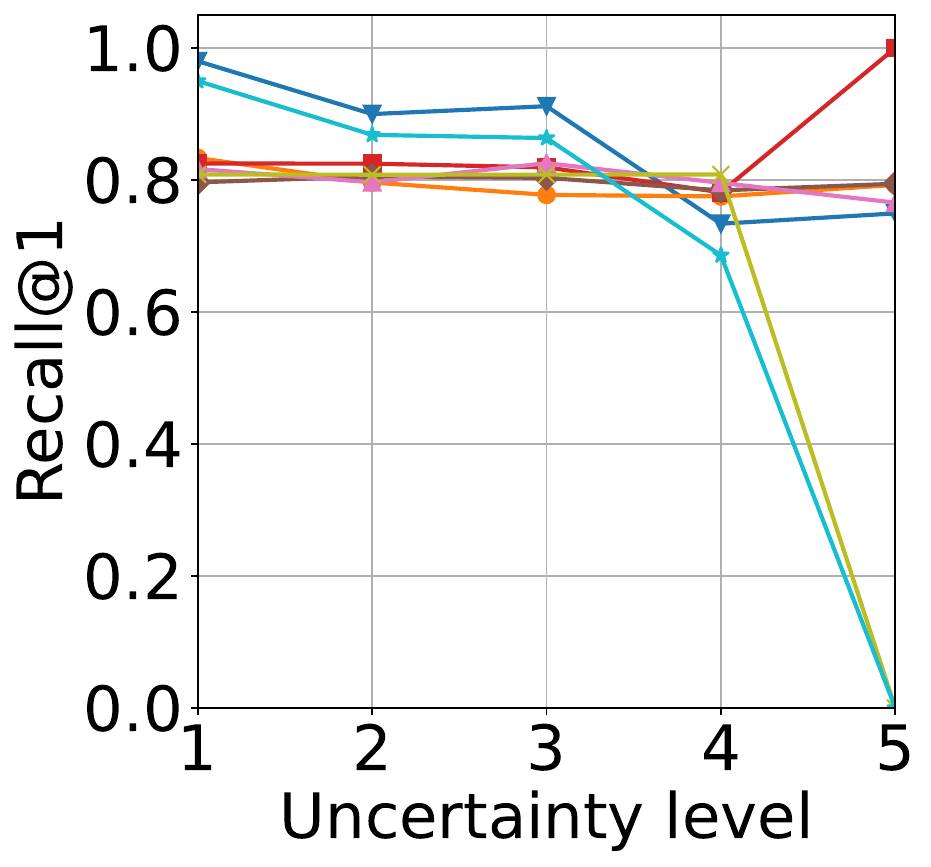} & 
        \includegraphics[width=0.23\linewidth, height=4cm]{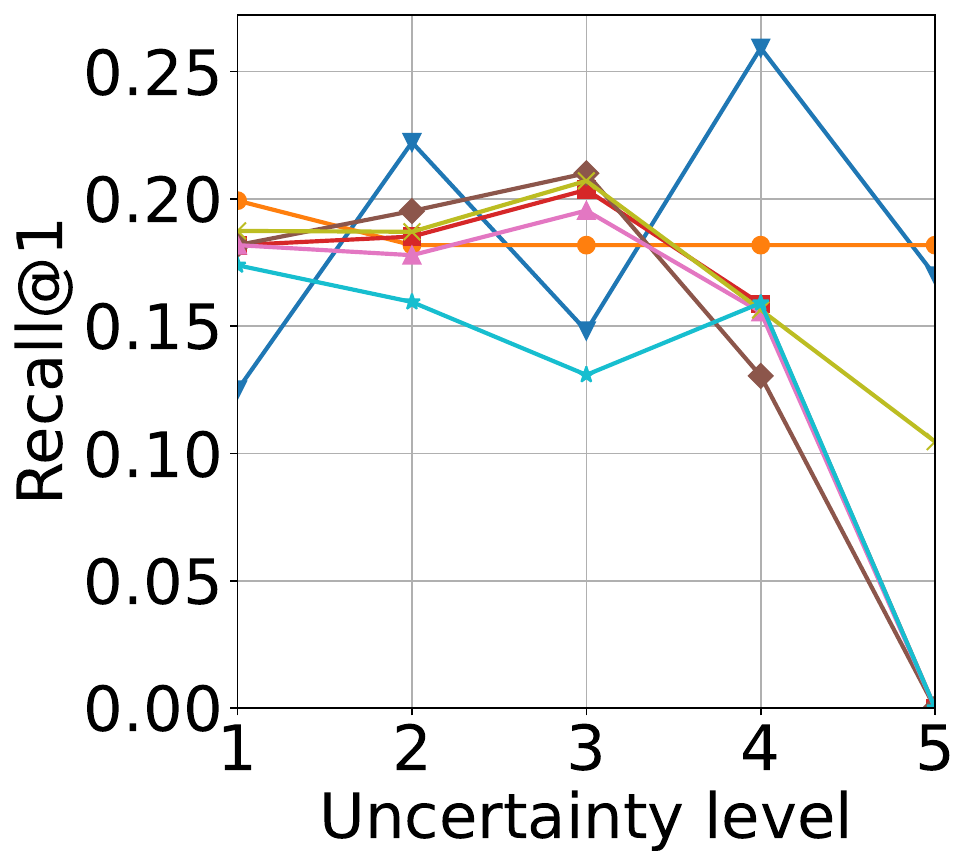} &
        \includegraphics[width=0.23\linewidth, height=4cm]{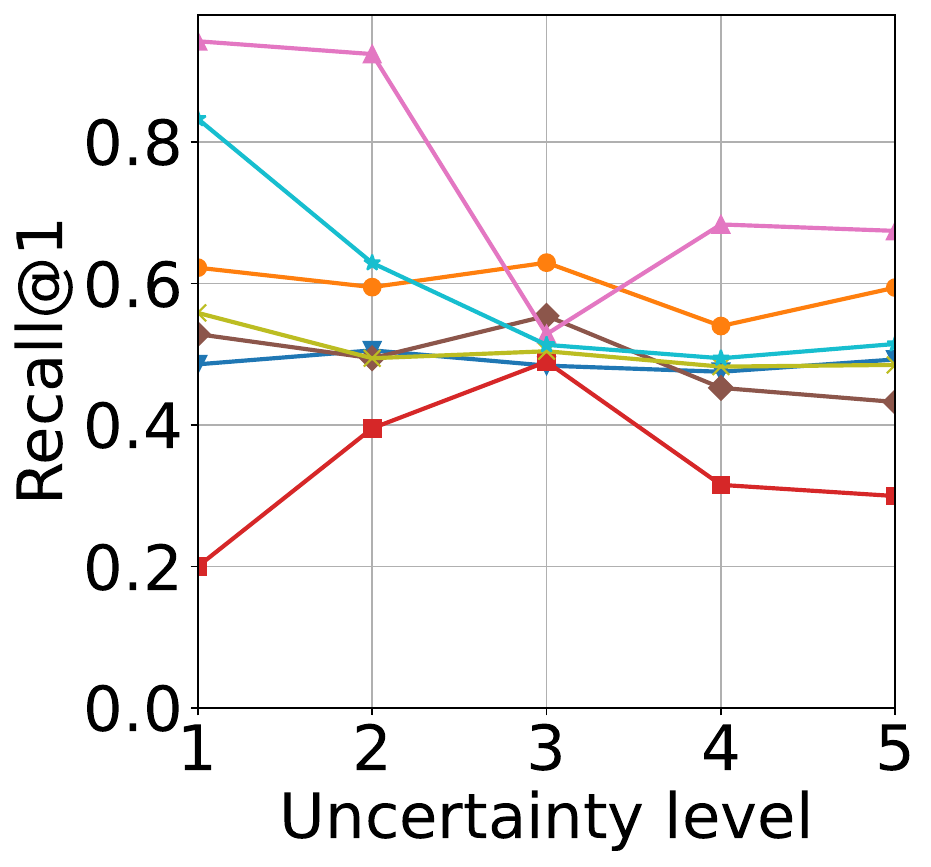}      
        \\ 
        \raisebox{1.2cm}{\rotatebox{90}{\small Text to Image}} &
        \includegraphics[width=0.23\linewidth, height=4cm]{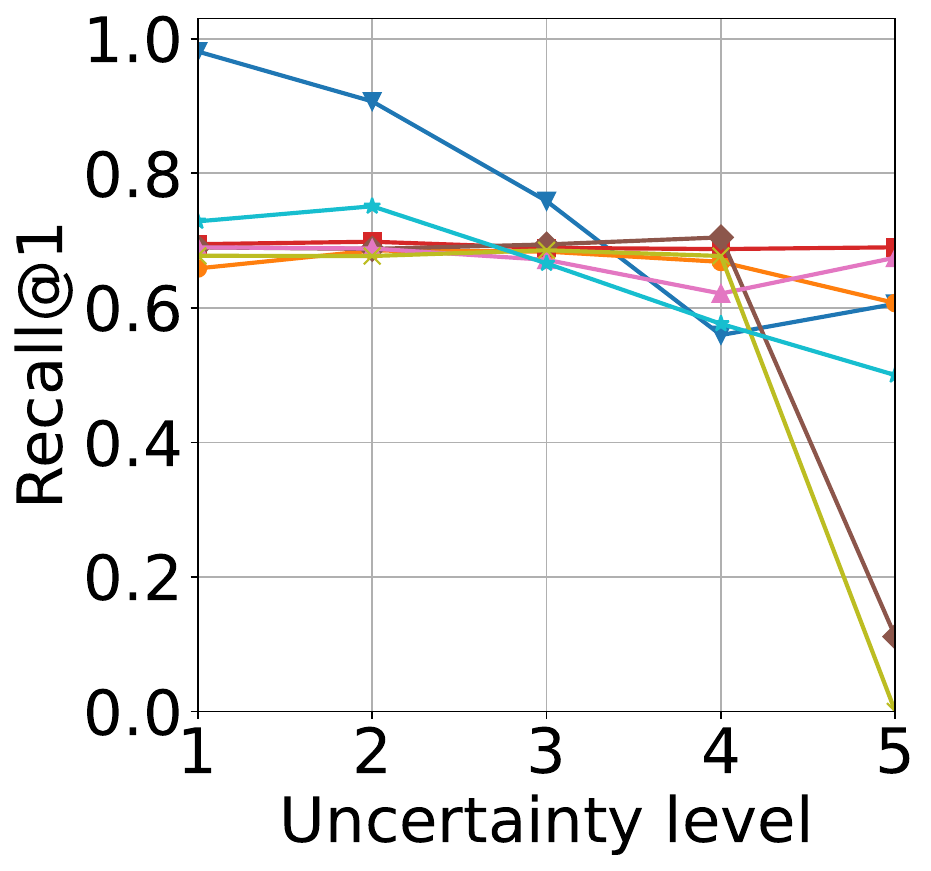} & 
        \includegraphics[width=0.23\linewidth, height=4cm]{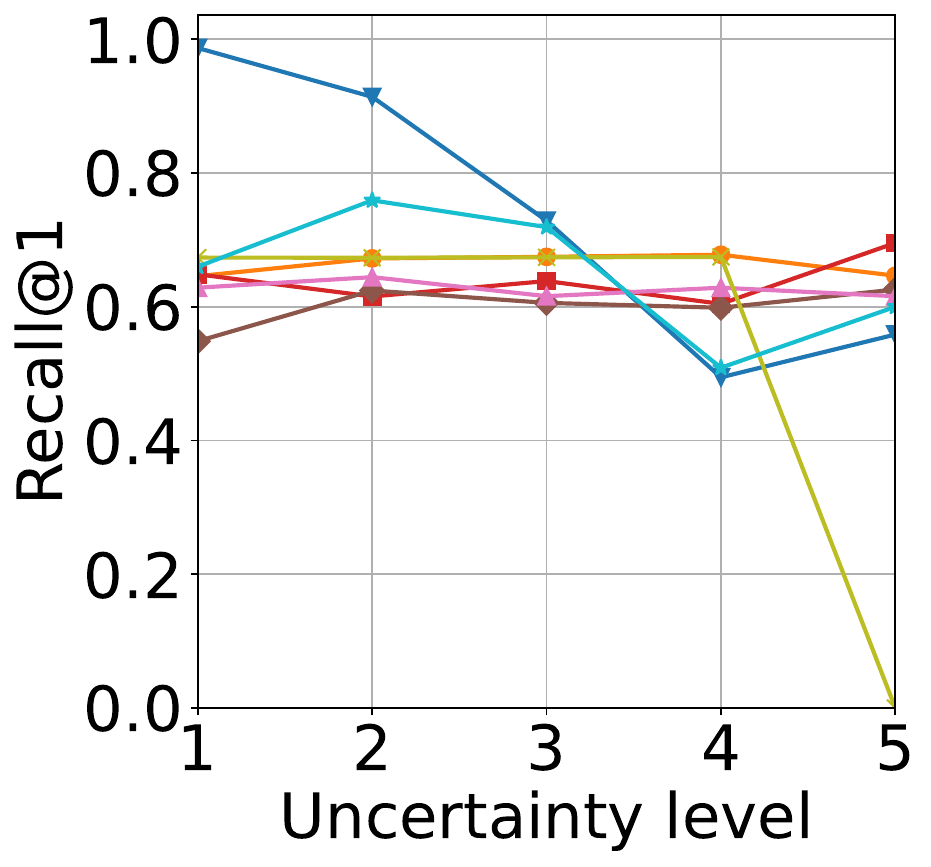} & 
        \includegraphics[width=0.23\linewidth, height=4cm]{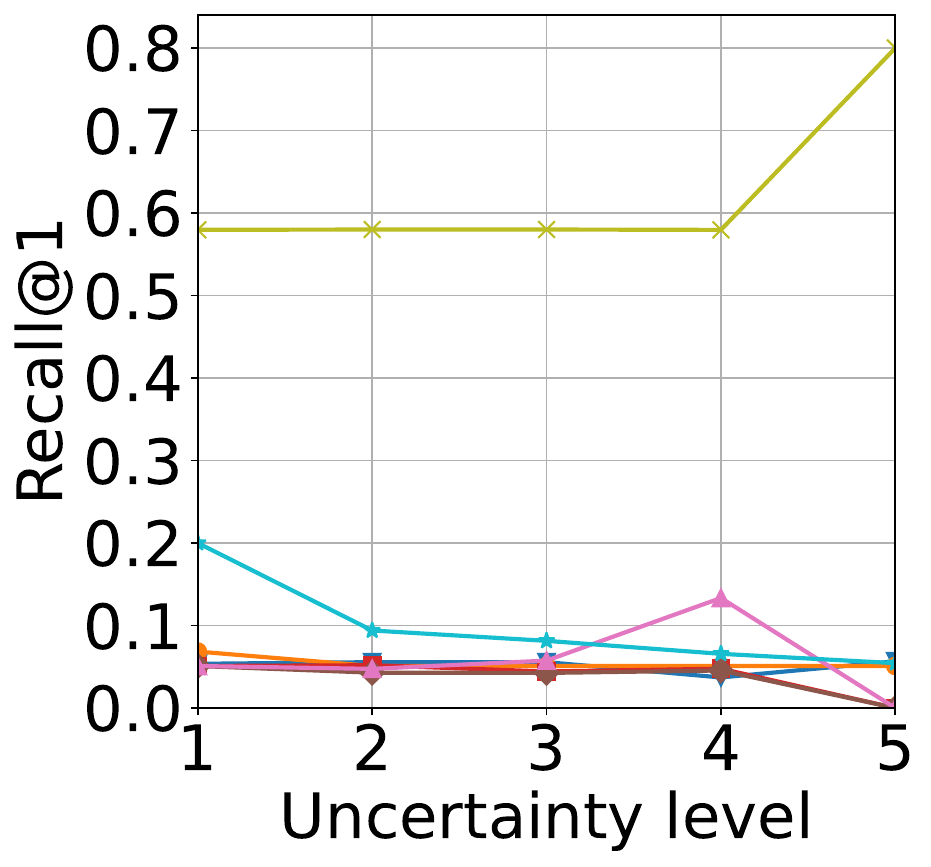} &
        \includegraphics[width=0.23\linewidth, height=4cm]{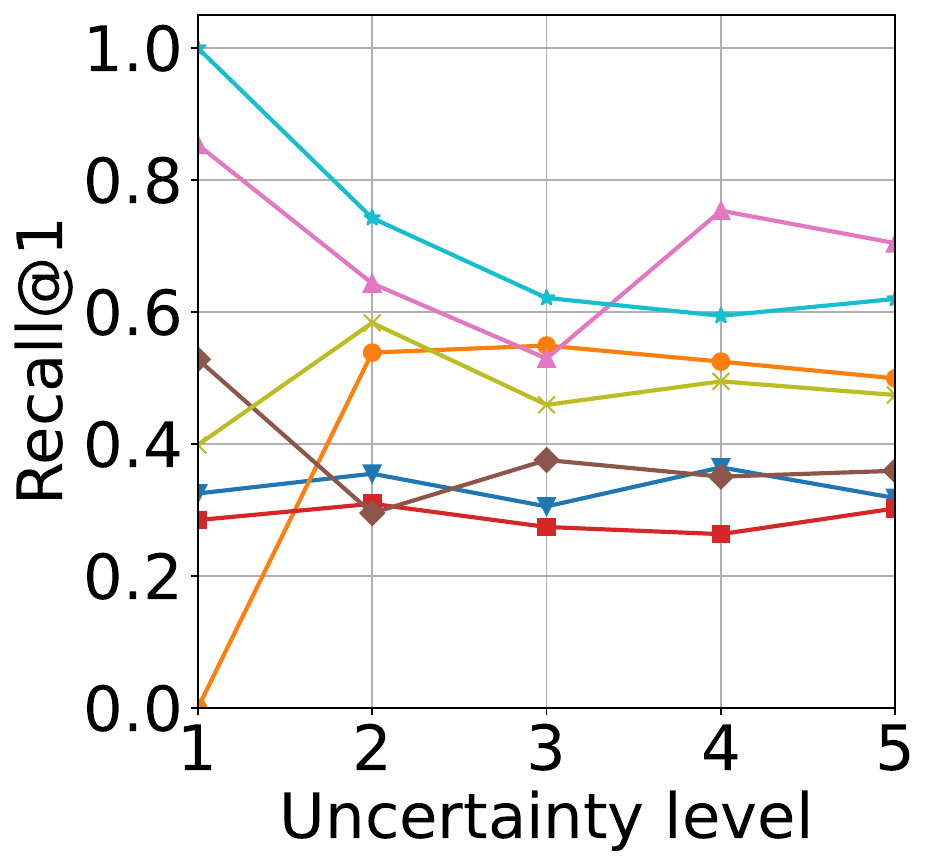}         
        \vspace{-5pt}
    \end{tabular}
    \vspace{-25pt}
    \includegraphics[width=0.8\linewidth]{legend.pdf} 
    \caption{\textbf{Evaluation of uncertainty calibration} for embeddings obtained from BLIP on Image-to-Text (top) and Text-to-Image (bottom) retrieval tasks. For perfect calibration, the Recall@1 should show a monotonic decrease as uncertainty levels increase. GroVE exhibits a more consistent relationship between increasing uncertainty and performance degradation compared to the baseline methods.}
    \label{fig:blip_calibration}
\end{figure*}

Figure~\ref{fig:clip_calibration} and Figure~\ref{fig:blip_calibration} show the calibration plots for the CLIP and BLIP models, respectively. 
Calibration plots are obtained by binning uncertainty values, referred to as uncertainty levels and computing Recall@1 for each bin. From the plots, GroVE maintains a more consistent alignment between decreasing uncertainty and increasing Recall@1.

\subsubsection{Retrieval performance}

Table~\ref{tab:retrieval} presents the Recall@1 scores for various baselines using CLIP. The score for the Deterministic baseline was computed by retrieving the nearest image/text embedding based on cosine similarity to the query text/image from the deterministic embeddings generated by the CLIP model. For the other baselines, retrieval was performed by selecting the image/text embedding with the minimum Wasserstein distance to the query, using the probabilistic image/text embeddings. Results show that GroVE achieves a good performance on the fine-grained CUB and Flowers dataset, whereas deterministic achieves the best scores in MS-COCO and Flickr30k dataset. 

\begin{table}[h]
    \centering
    \begin{tabular}{p{0.2cm}lcccc}
    \toprule
    &{Method} & Flickr & COCO & CUB & Flowers \\
    \midrule
        \multirow{7}{*}{\rotatebox{90}{\textbf{Image to Text}}} 
        &Deterministic & \textbf{0.801} & \textbf{0.715} & \textbf{0.532} & \underline{0.357}\\ 
        &TTDA & 0.423 & 0.326 & 0.133 & 0.289\\
        &PFE & 0.238 & 0.213 & 0.101 & 0.102\\
        &PCME & 0.392 & 0.246 & 0.129 & 0.134\\
        &PCME++ & 0.423 &0.397 & 0.124 & 0.111\\
        &ProbVLM & 0.491 & 0.303 & 0.136 & 0.245\\
        &GroVE & \underline{0.569} & \underline{0.512} & \underline{0.307} & \textbf{0.402}\\
    
     \midrule
     \multirow{7}{*}{\rotatebox{90}{\textbf{Text to Image}}} 
        &Deterministic & \textbf{0.543} & \textbf{0.515} & \underline{0.141} & \underline{0.109}\\ 
        &TTDA & 0.202 & 0.139 & 0.046 & 0.057\\
        &PFE & \underline{0.298} & 0.219 & 0.023 & 0.024\\
        &PCME & 0.092 & 0.102 & 0.099 & 0.029\\
        &PCME++ & 0.133 & 0.125 & 0.087 & 0.058\\
        &ProbVLM & 0.104 & 0.156 & 0.005 & 0.102\\
        &GroVE & 0.241 & \underline{0.288} & \textbf{0.343} & \textbf{0.379}\\
    
     \bottomrule
        
    \end{tabular}
    \caption{\textbf{Retrieval performance using CLIP.} Table shows the Recall@1 scores obtained using the different baselines. GroVE achieves the best scores for the fine-grained datasets.
    }
    \label{tab:retrieval}
\end{table}

\subsubsection{Qualitative Analysis}
\begin{figure}[h]
    \centering
    \includegraphics[angle=-90,width=0.8\linewidth]{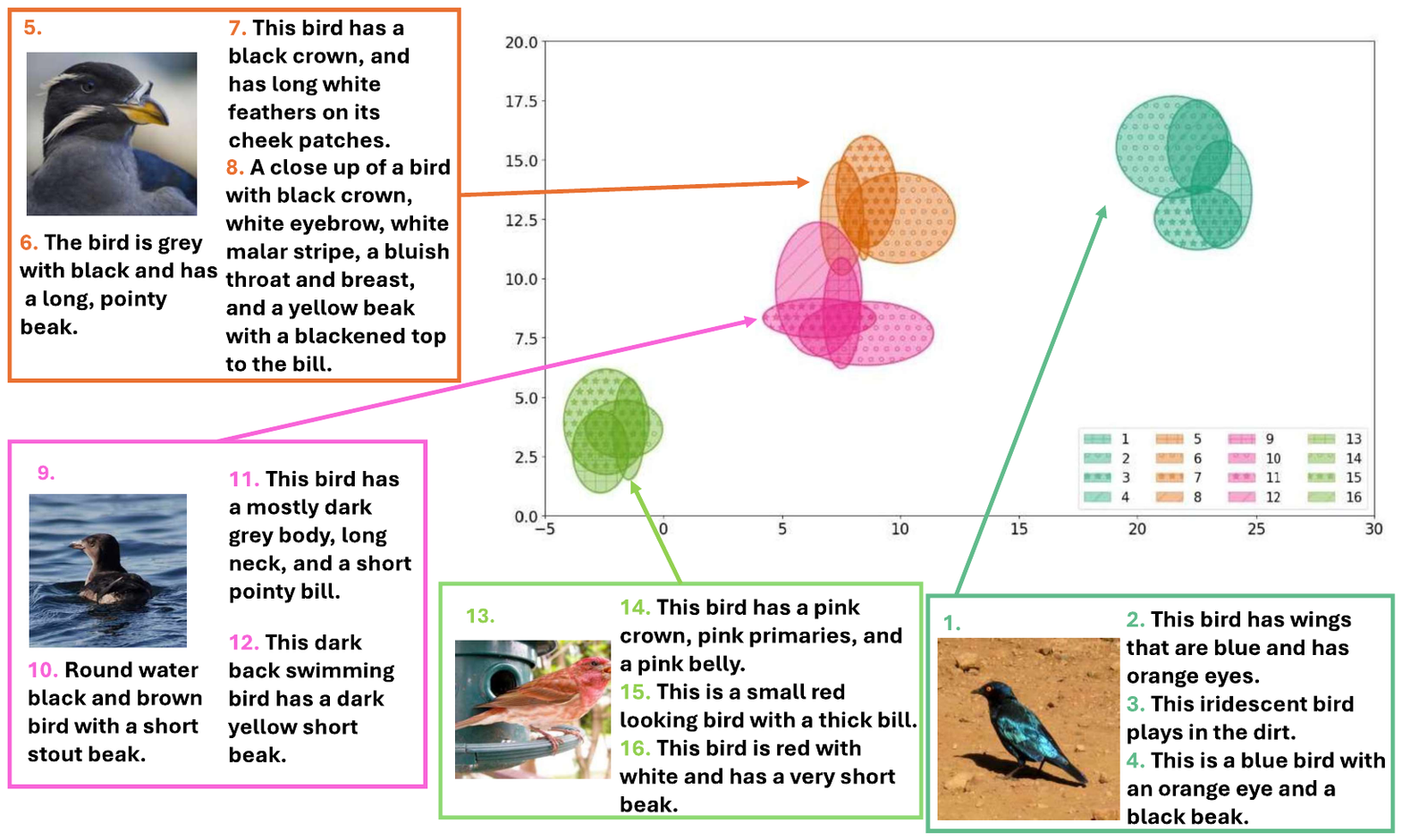}
    \vspace{-25pt}
    \caption{\textbf{t-SNE visualization of the probabilistic representations} generated by GroVE on a subset of the CUB-200-2011 dataset. Starting from deterministic embeddings provided by frozen VLMs, GroVE produces corresponding probabilistic representations that capture input ambiguities. }

    \vspace{-10pt}
    \label{fig:latent}
\end{figure}

\begin{figure}
    \centering
    \includegraphics[width=0.8\linewidth]{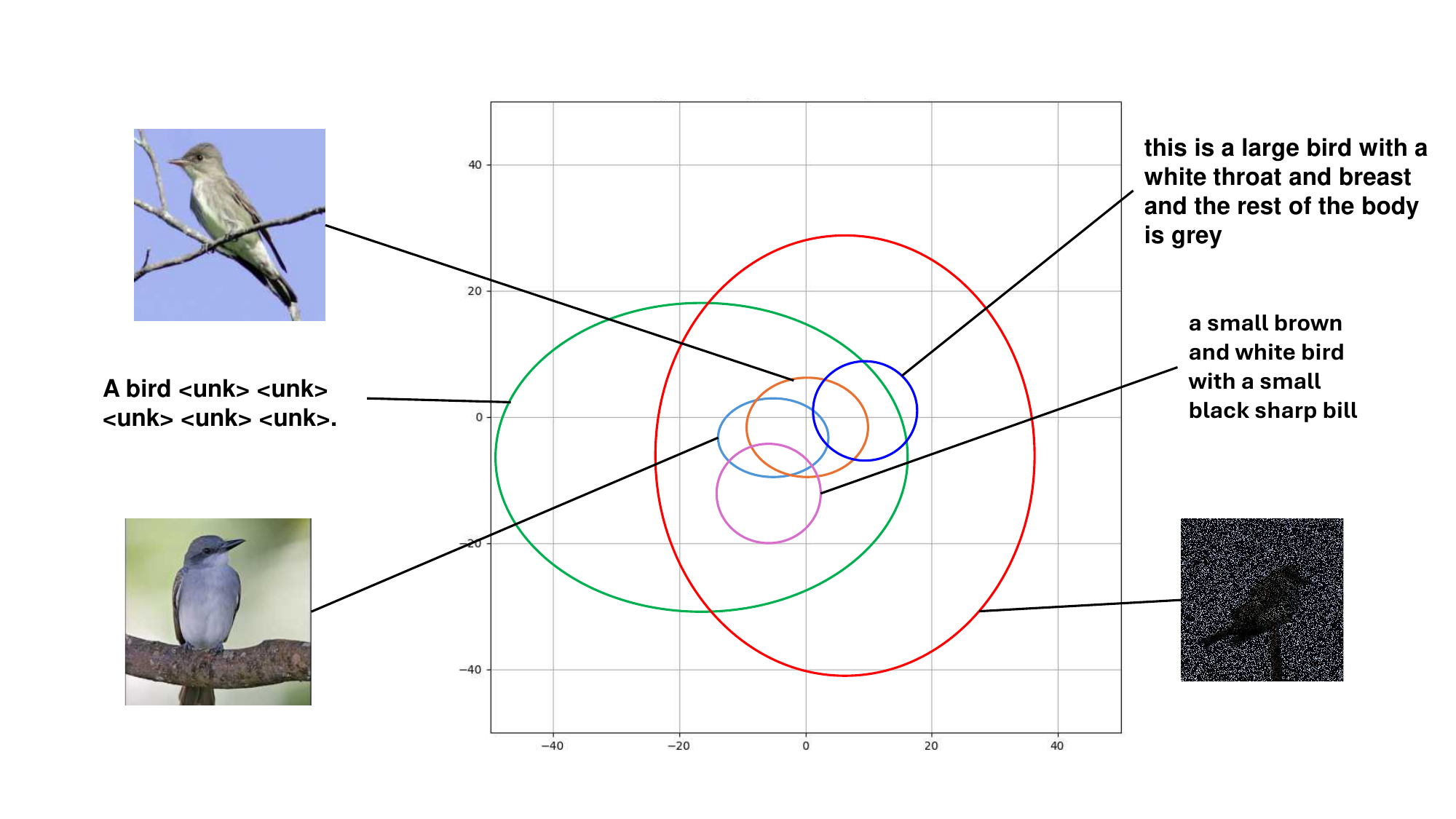}
    \caption{Illustration of failure case of GroVE where the model makes incorrect predictions of the CUB-200-2011 dataset due to the high inter-class similarity.}
    \label{fig:failure}
\end{figure}

A t-SNE visualization of the probabilistic embeddings from GroVE on a subset of the CUB dataset is provided in Figure~\ref{fig:latent}, where the uncertainty corresponds to the area of the embedding. 
The plot shows that images and texts with similar semantic content are clustered together, and the probabilistic embeddings are able to capture the uncertainty arising from the data ambiguities. Figure~\ref{fig:failure} illustrates a scenario from the CUB-200-2011 dataset where incorrect predictions sometimes occur due to high inter-class similarity~\cite{venkataramanan2021tackling} between the image and text descriptions of two distinct bird species. We also include a scenario where either the image or text is masked, introducing ambiguity. In such cases, GroVE assigns a distribution with higher variance, reflecting increased uncertainty.

\clearpage

\subsection{Zero-shot Uncertainty Calibration} \label{sec:app_zs}

We evaluate the generalization of uncertainty calibration across methods that use auxiliary models for probabilistic embeddings on out-of-distribution datasets. Two CLIP experiments are conducted by training on MS-COCO and CUB, then evaluating on their respective unseen datasets. 
Table~\ref{tab:coco_zs} and \ref{tab:cub_zs} presents the $-SR^2$ scores with models trained on MS-COCO and CUB respectively. The models trained on MS-COCO show a
strong performance on Flickr30k due to its similarity to MS-COCO, thereby exhibiting better generalization. There is a drop in performance on the more fine-grained Flowers and CUB datasets, particularly for text-to-image retrieval. GroVE, however, demonstrates better generalization than the baseline methods for both the experiments. 

\begin{table}[tb]
    \centering
    \begin{tabular}{p{0.2cm}lccc}
    \toprule
    &{Method} & Flickr & Flowers & CUB \\
    \midrule
        \multirow{5}{*}{\rotatebox{90}{\textbf{Image to Text}}} 
        &PFE & 0.01$\pm$0.03 & 0.38$\pm$0.04 & 0.02$\pm$0.02\\
        &PCME & 0.04$\pm$0.02 & 0.13$\pm$0.04 &0.09$\pm$0.06\\
        &PCME++ & 0.01$\pm$0.02 & \underline{0.48$\pm$0.03} & 0.03$\pm$0.02\\
        &ProbVLM & \underline{0.55$\pm$0.03}  & 0.19$\pm$0.04 & \underline{0.15$\pm$0.04}\\
        &GroVE & \textbf{0.74$\pm$0.03} & \textbf{0.69$\pm$0.02} & \textbf{0.41$\pm$0.03}\\
    
     \hline
        \multirow{5}{*}{\rotatebox{90}{\textbf{Text to Image}}} 
        &PFE & \underline{0.41$\pm$0.03} & 0.02$\pm$0.03   & 0.04$\pm$0.01\\
        &PCME &  0.24$\pm$0.03 & -0.01$\pm$0.02 & \underline{0.02$\pm$0.03}\\
        &PCME++ & -0.43$\pm$0.03 &  \underline{0.05$\pm$0.03} & 0.03$\pm$0.02\\
        &ProbVLM & 0.14$\pm$0.05 & 0.01$\pm$0.03 & 0.00$\pm$0.01\\
        &GroVE & \textbf{0.42$\pm$0.02} & \textbf{0.09$\pm$0.03} & \textbf{0.04$\pm$0.02}\\
    
     \bottomrule
        
    \end{tabular}
    \caption{\textbf{Zero-shot uncertainty calibration - MS-COCO.} GroVE outperforms other baselines in most cases, achieving superior uncertainty calibration in zero-shot settings. The best scores are highlighted in bold and the second-best scores are underlined.
    }
    \label{tab:coco_zs}
\end{table}

\begin{table}[tb]
    \centering
    \begin{tabular}{p{0.2cm}lccc}
    \toprule
    &{Method} & Flickr & COCO & Flowers \\
    \midrule
        \multirow{5}{*}{\rotatebox{90}{\textbf{Image to Text}}} 
        &PFE & 0.00$\pm$0.04 & 0.02$\pm$0.03 & -0.13$\pm$0.03\\
        &PCME & \underline{0.46$\pm$0.02} & 0.01$\pm$0.05 &0.02$\pm$0.04\\
        &PCME++ & 0.40$\pm$0.03 & 0.10$\pm$0.02 & \underline{0.44$\pm$0.03}\\
        &ProbVLM & 0.15$\pm$0.02  & \underline{0.38$\pm$0.03} & 0.18$\pm$0.03\\
        &GroVE & \textbf{0.59$\pm$0.03} & \textbf{0.45$\pm$0.03} & \textbf{0.50$\pm$0.04}\\
    
     \hline
        \multirow{5}{*}{\rotatebox{90}{\textbf{Text to Image}}} 
        &PFE & -0.01$\pm$0.02 & 0.31$\pm$0.04   & -0.12$\pm$0.03\\
        &PCME &  \underline{0.14$\pm$0.02} & -0.19$\pm$0.03 & 0.15$\pm$0.04\\
        &PCME++ & 0.13$\pm$0.02 &  \textbf{0.52$\pm$0.03} & \underline{0.36$\pm$0.03}\\
        &ProbVLM & 0.01$\pm$0.03 & 0.01$\pm$0.02 & 0.02$\pm$0.03\\
        &GroVE & \textbf{0.76$\pm$0.03} & \underline{0.42$\pm$0.02} & \textbf{0.37$\pm$0.03}\\
    
     \bottomrule
        
    \end{tabular}
    \caption{\textbf{Zero-shot uncertainty calibration - CUB-200-2011.} GroVE outperforms other baselines in most cases, achieving superior uncertainty calibration in zero-shot settings. The best scores are highlighted in bold and the second-best scores are underlined.
    }
    \label{tab:cub_zs}
\end{table}

\begin{figure*}[t]
    \centering
    \begin{tabular}{@{}c@{}c@{}c@{}c@{}}
        & Flickr & Flowers & CUB  \\
        \raisebox{1.2cm}{\rotatebox{90}{\small Image to Text}} &
        \includegraphics[width=0.25\linewidth, height=4cm]{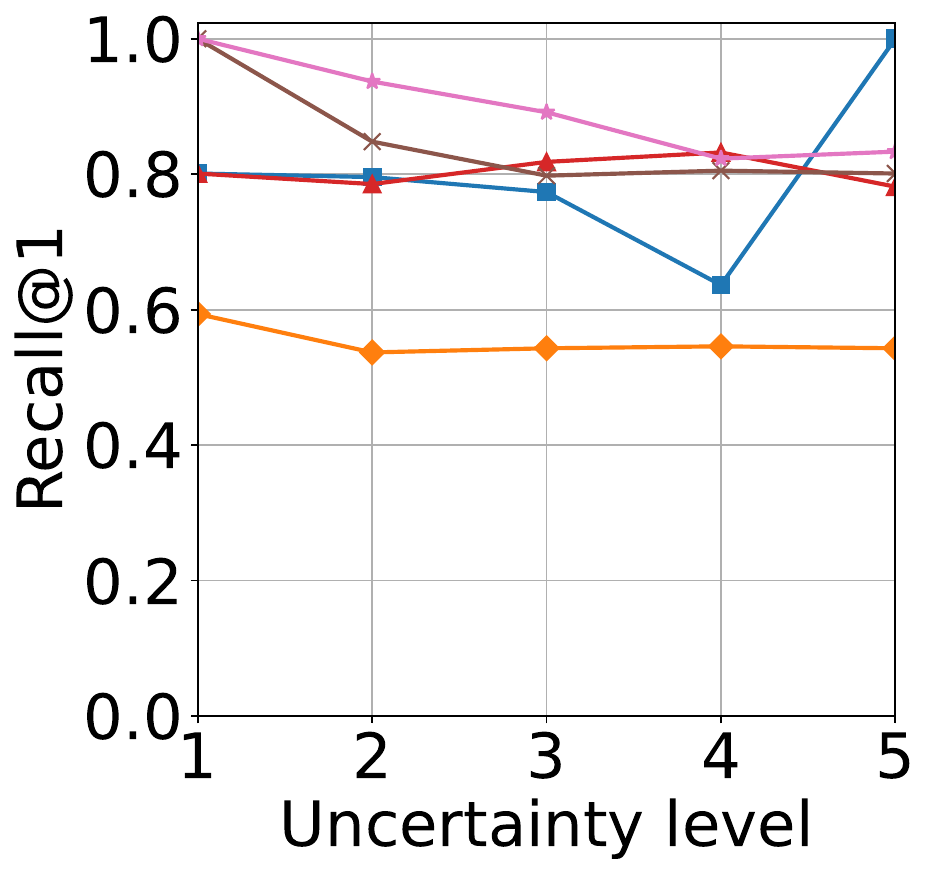} & 
        \includegraphics[width=0.25\linewidth, height=4cm]{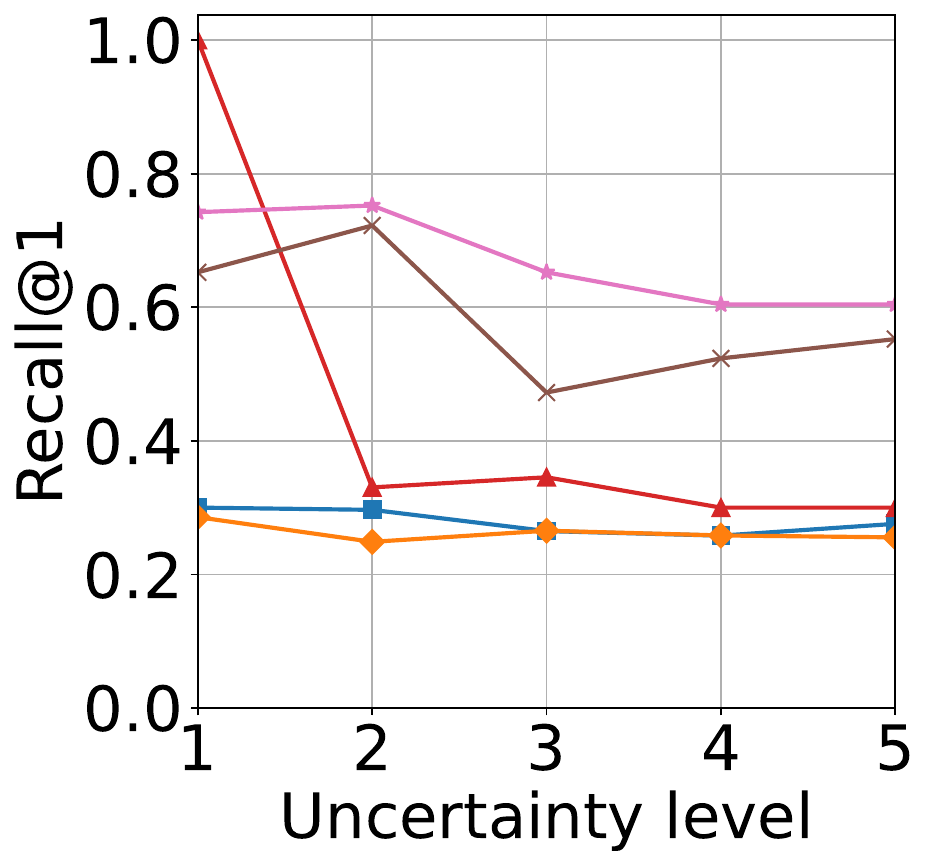} & 
        \includegraphics[width=0.25\linewidth, height=4cm]{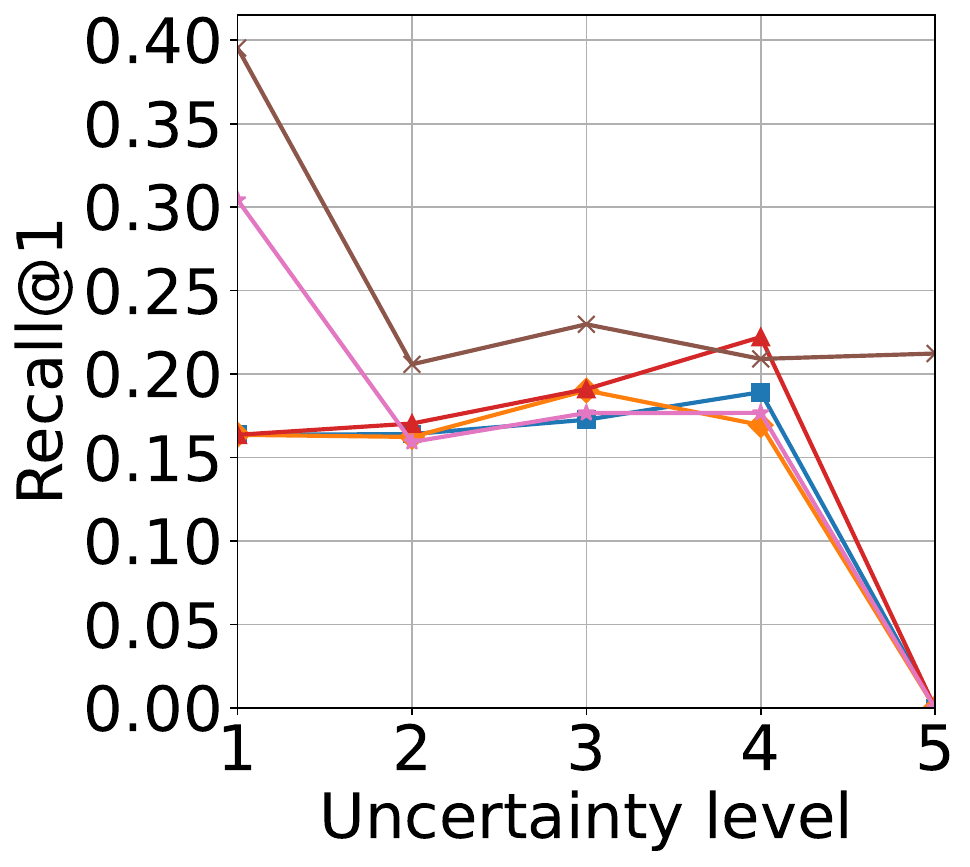}     
        \\
        
        \raisebox{1.2cm}{\rotatebox{90}{\small Text to Image}} &
        \includegraphics[width=0.25\linewidth, height=4cm]{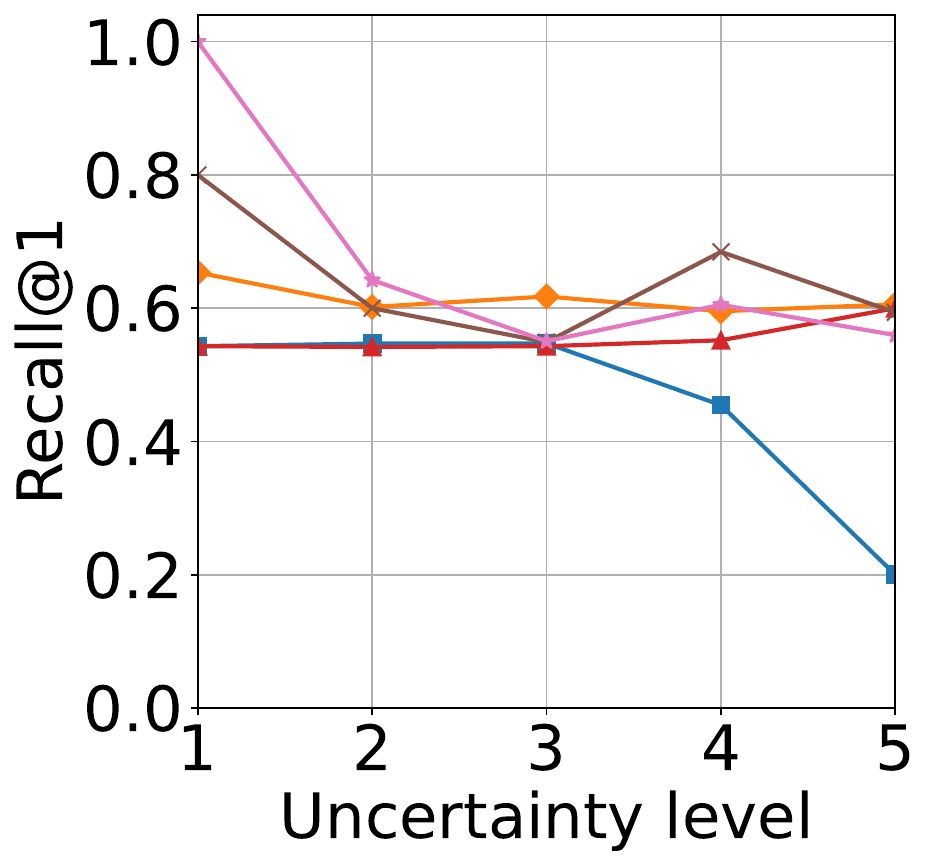} & 
        \includegraphics[width=0.25\linewidth, height=4cm]{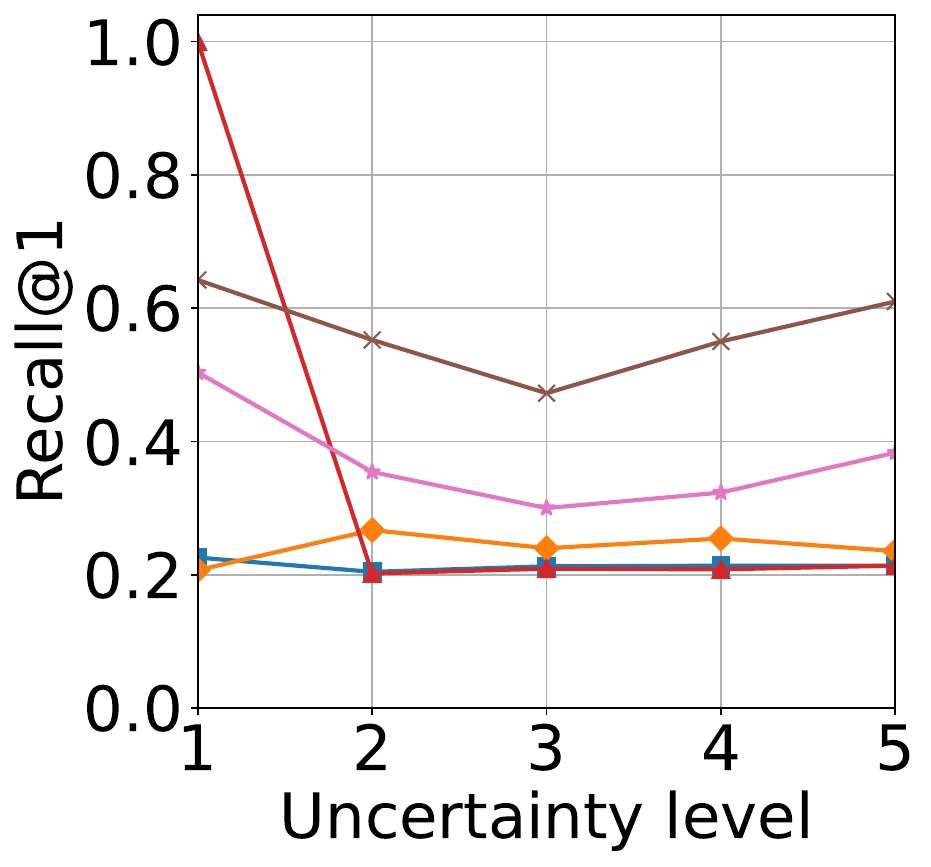} & 
        \includegraphics[width=0.25\linewidth, height=4cm]{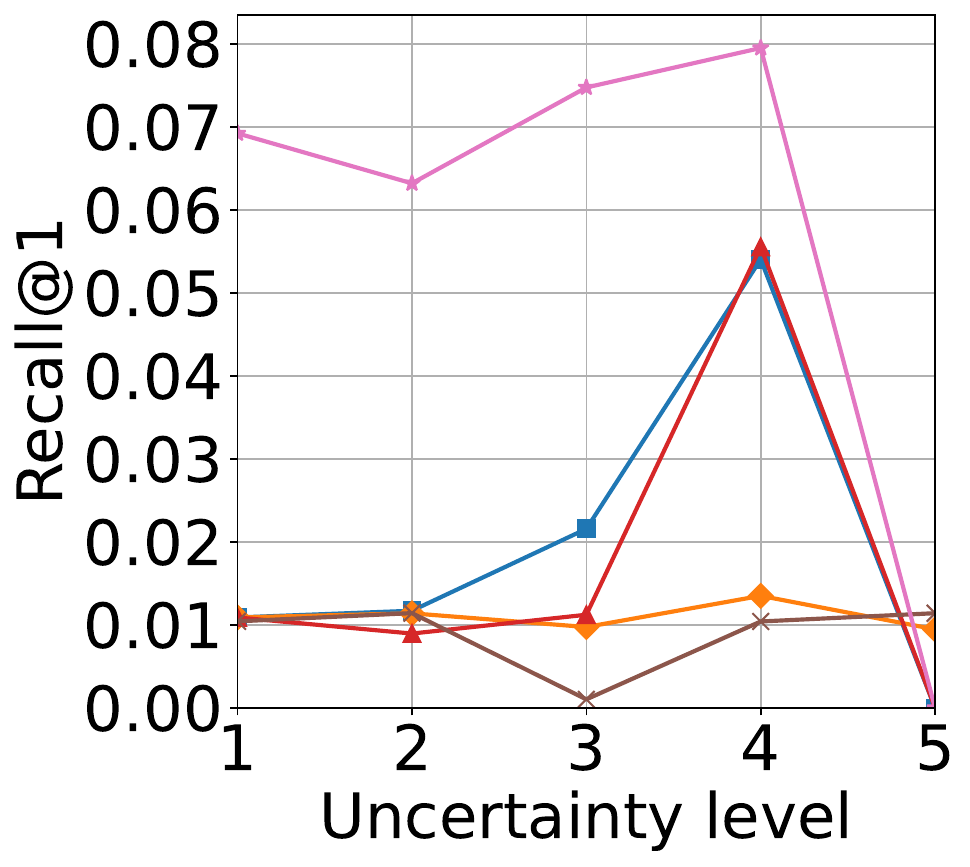}     
    \end{tabular}
    \vspace{-5pt}
    \includegraphics[width=0.7\linewidth]{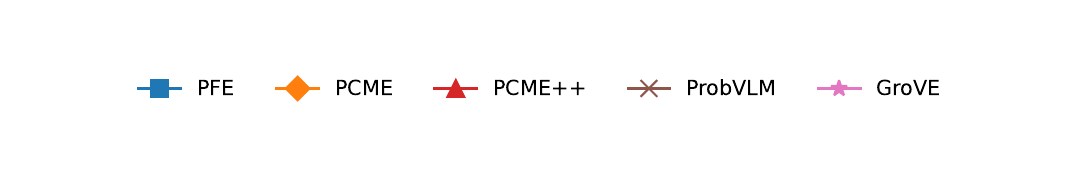} 
    \vspace{-25pt}
    \caption{\textbf{Evaluation of zero-shot uncertainty calibration using MS-COCO} for embeddings obtained from CLIP on Image-to-Text (top) and Text-to-Image (bottom) retrieval tasks. For perfect calibration, the Recall@1 should show a monotonic decrease as uncertainty levels increase. GroVE exhibits a more consistent relationship between increasing uncertainty and performance degradation compared to the baseline methods.}
    \label{fig:coco_zs_calibration}
\end{figure*}

\begin{figure*}[h]
    \centering
    \begin{tabular}{@{}c@{}c@{}c@{}c@{}}
        & Flickr & MS-COCO & Flowers  \\
        \raisebox{1.2cm}{\rotatebox{90}{\small Image to Text}} &
        \includegraphics[width=0.25\linewidth, height=4cm]{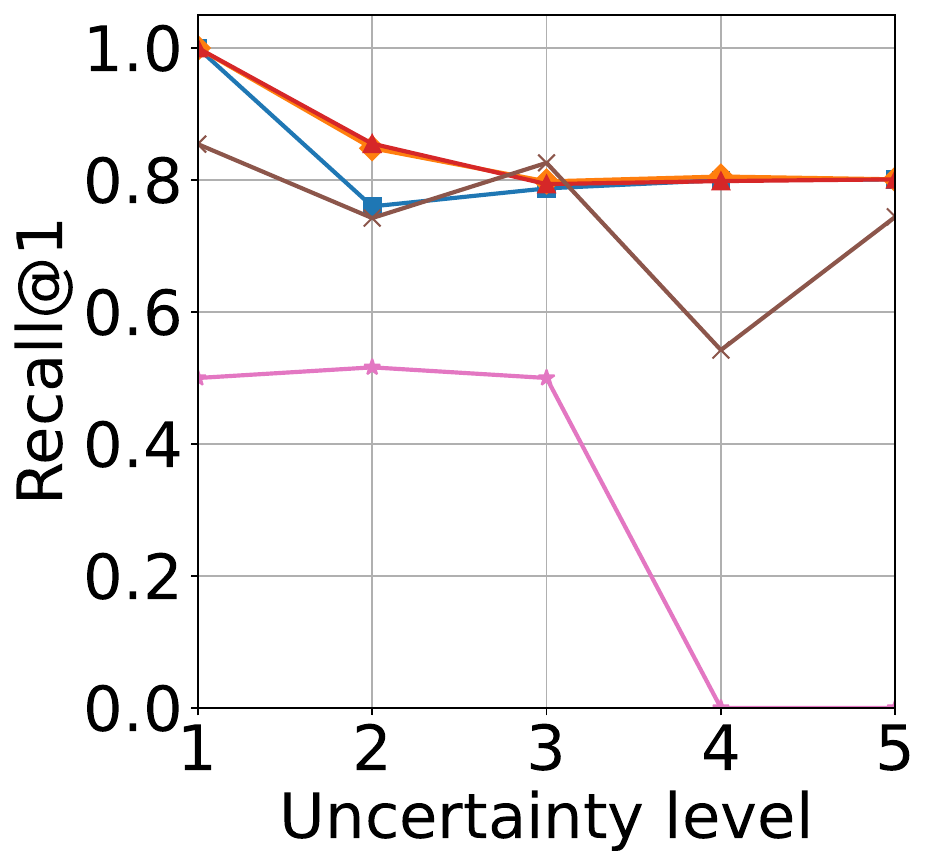} & 
        \includegraphics[width=0.25\linewidth, height=4cm]{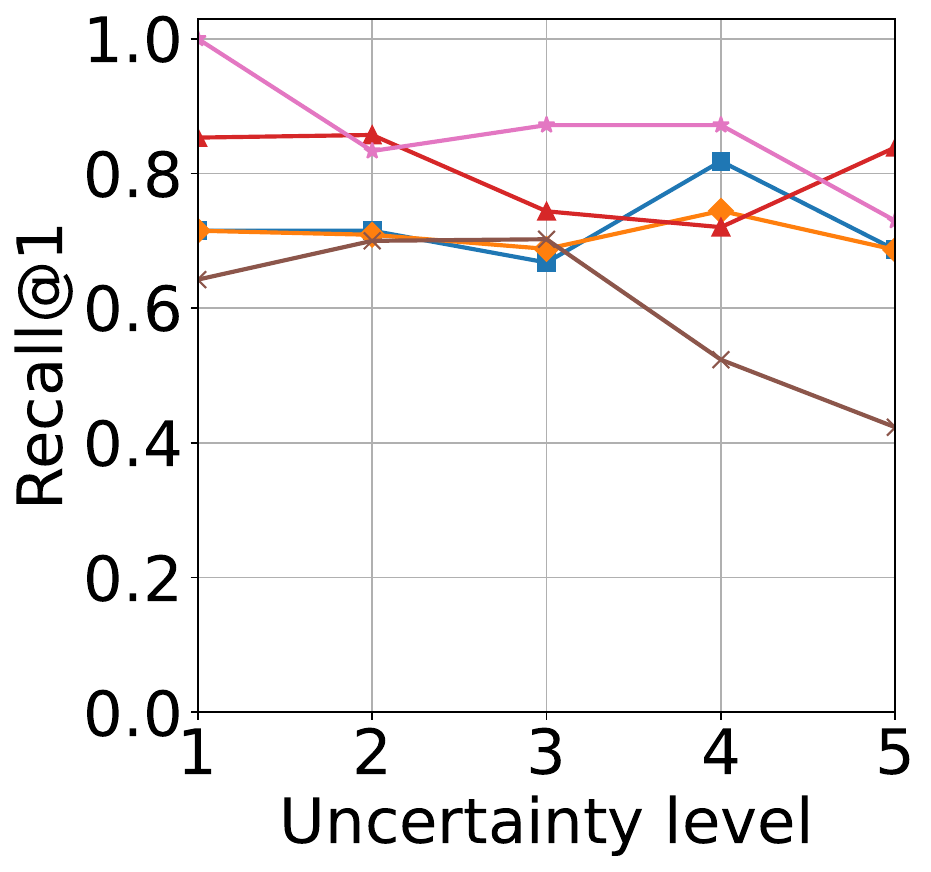} & 
        \includegraphics[width=0.25\linewidth, height=4cm]{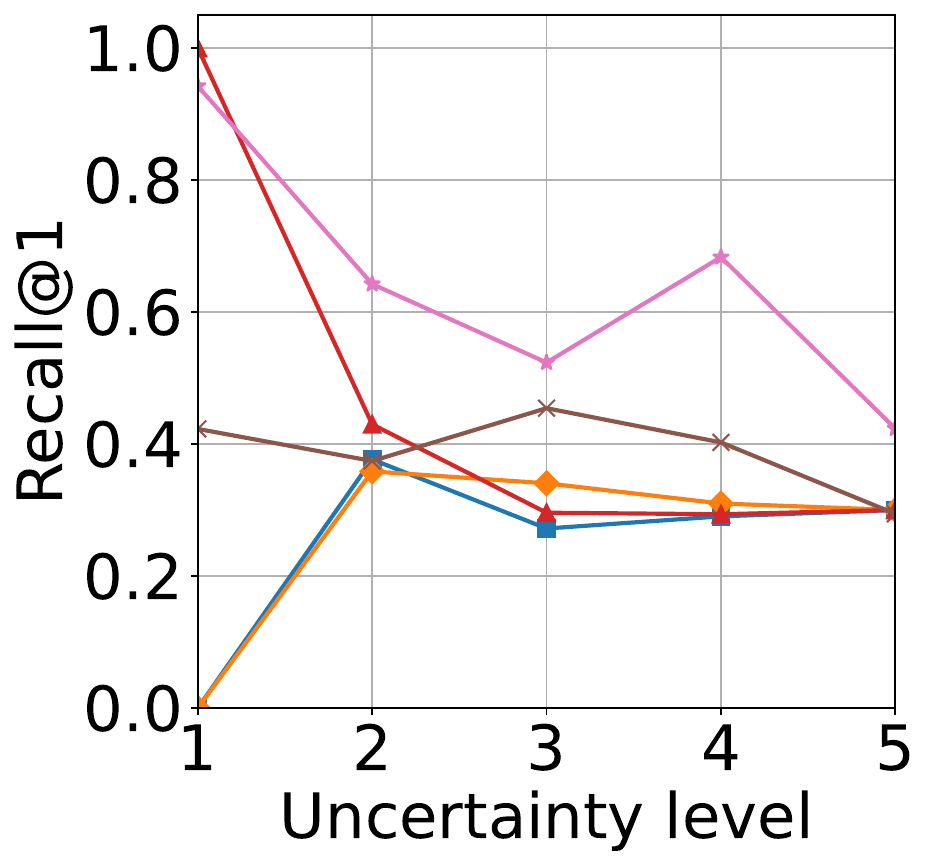}     
        \\
        
        \raisebox{1.2cm}{\rotatebox{90}{\small Text to Image}} &
        \includegraphics[width=0.25\linewidth, height=4cm]{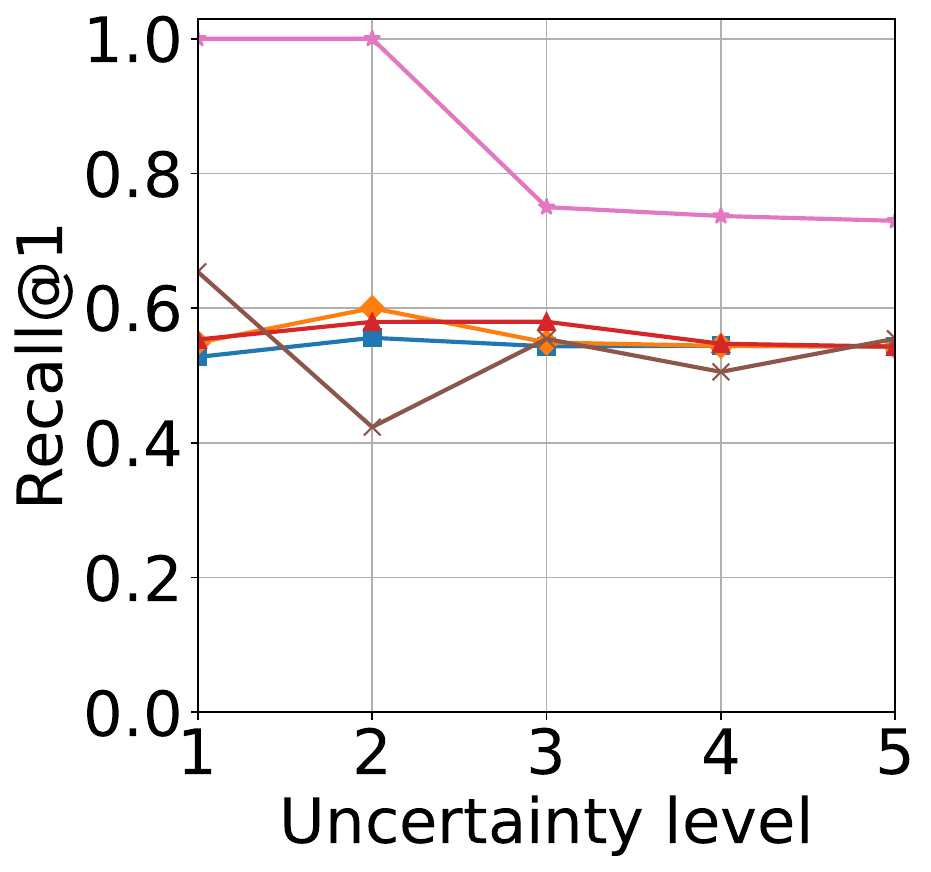} & 
        \includegraphics[width=0.25\linewidth, height=4cm]{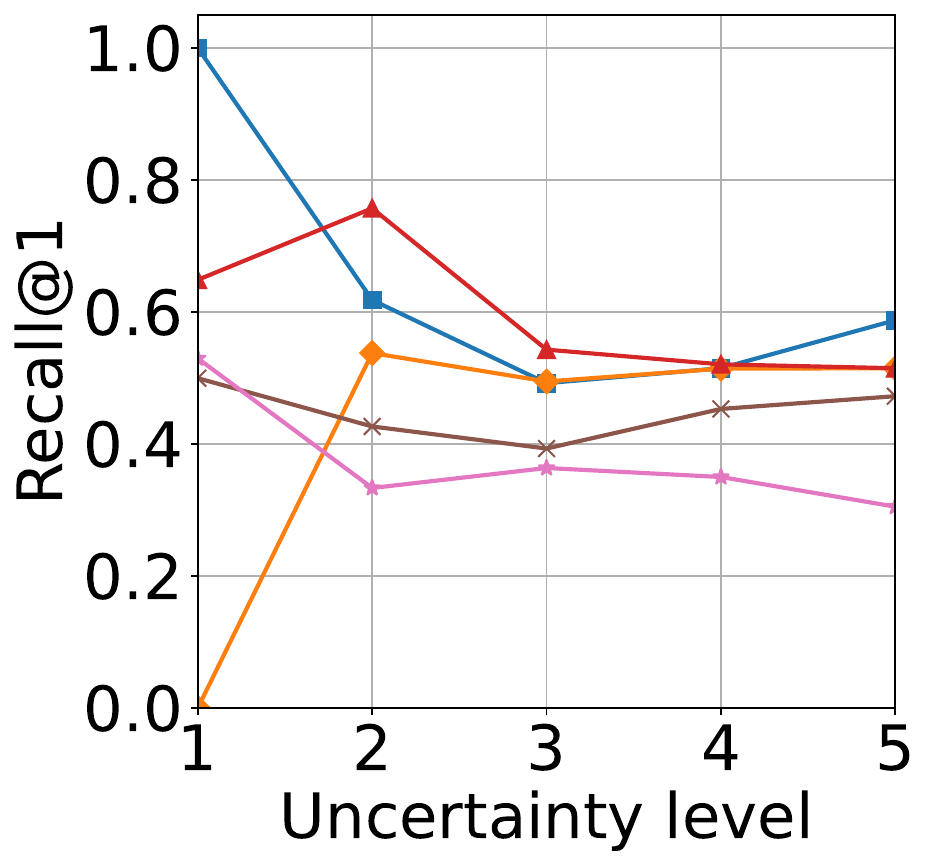} & 
        \includegraphics[width=0.25\linewidth, height=4cm]{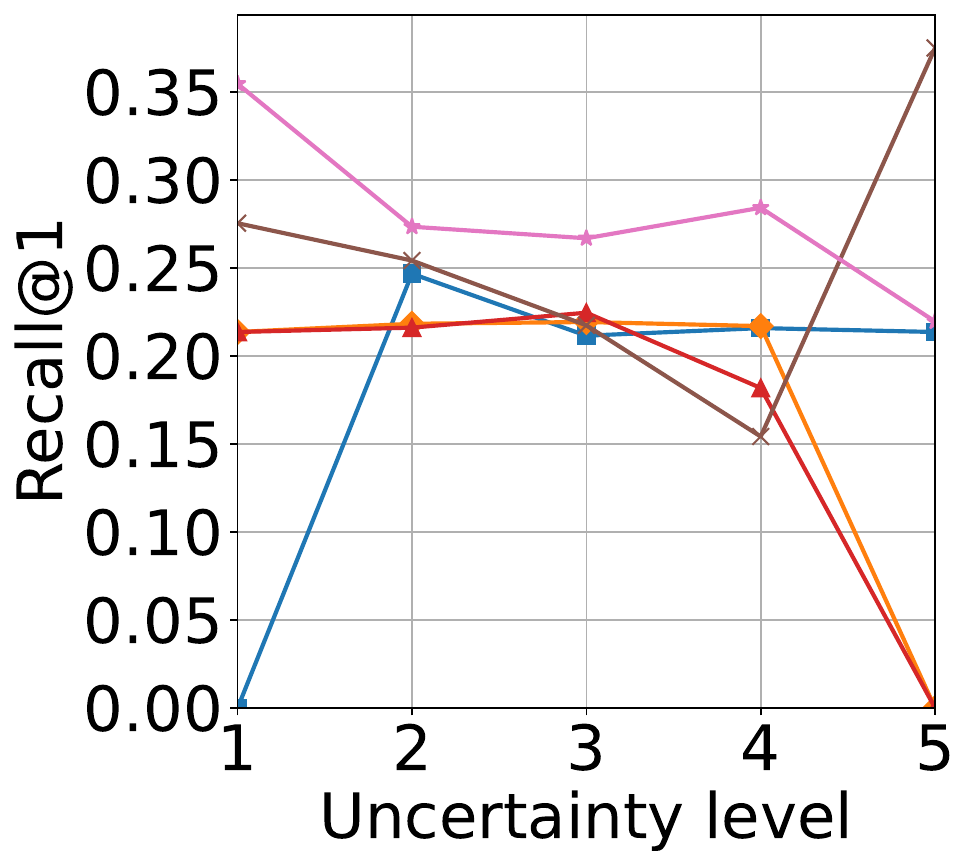}      
    \end{tabular}
    \vspace{-5pt}
    \includegraphics[width=0.7\linewidth]{legend_zs.pdf} 
    \vspace{-25pt}
    \caption{\textbf{Evaluation of zero-shot uncertainty calibration using CUB-200-2011} for embeddings obtained from CLIP on Image-to-Text (top) and Text-to-Image (bottom) retrieval tasks. For perfect calibration, the Recall@1 should show a monotonic decrease as uncertainty levels increase. GroVE exhibits a more consistent relationship between increasing uncertainty and performance degradation compared to the baseline methods.}
    \label{fig:cub_zs_calibration}
\end{figure*}

The calibration results for the experiments are presented in Figure~\ref{fig:coco_zs_calibration} and Figure~\ref{fig:cub_zs_calibration}, respectively, where 
GroVE maintains a more consistent alignment between decreasing uncertainty and increasing Recall@1.

\clearpage

\subsection{Few-shot Uncertainty Calibration} \label{sec:app_few_shot}

Table~\ref{tab:few-shot-retrieval} shows the Recall@1 scores for the cross-modal retrieval task for the auxiliary models trained using limited data from the synthetic CUB dataset.  The performance of the neural network based methods drop, which is expected given the insufficient number of data points for the model to generalize. Note that deterministic and TTDA are agnostic to the few shot setting since they work directly on the VLM embeddings for the prediction. Among the methods using auxiliary models, GroVE achieves a higher score, leveraging the ability of GPs to generalize well even with limited data because of their distance awareness property by capturing structure through kernel functions. Moreover, as the number of inducing points increases, GroVE's performance improves, with the best results achieved when performing exact GP. 
However, GroVE is computationally expensive compared to the neural network based approaches, with longer inference time as the number of inducing points increases.

\begin{table}[tbh]
    \centering
    \begin{tabular}{cccc}
    \toprule
     Method  & Image to Text & Text to Image & Time (ms/example) ($\downarrow$)\\ \midrule
     Determinsitic  & \textbf{0.532} & \underline{0.141} & \textbf{29.98}\\
     TTDA (10 passes) &  \underline{0.133$\pm$0.003} & 0.046$\pm$0.011 & 288.51 \\
     PFE & 0.062$\pm$0.001 & 0.026$\pm$0.010 & 31.59\\
     PCME & 0.074$\pm$0.002 & 0.031$\pm$0.005 & 31.60\\
     PCME++ & 0.063$\pm$0.003 & 0.031$\pm$0.003 & \underline{31.55}\\
     ProbVLM & 0.081$\pm$0.001 & 0.034$\pm$0.005 & 32.80\\
     \midrule
     GroVE (M=50) & 0.062$\pm$0.002 & 0.035$\pm$0.009 & 47.62\\
     GroVE (M=150) & 0.084$\pm$0.004 & 0.049$\pm$0.004 & 142.85\\
     GroVE (M=250) & 0.086$\pm$0.003 & 0.056$\pm$0.004 & 392.16\\
     GroVE (exact GP) & 0.103$\pm$0.002 & \textbf{0.182$\pm$0.002} & 1130.09\\ \bottomrule
    \end{tabular}
    \caption{Retrieval performance using Recall@1 scores and inference speed per instance for few-shot experiment using CUB-200-2011. The best results are highlighted in bold and the second best are underlined.}
    \label{tab:few-shot-retrieval}
\end{table}

\end{document}